\documentclass[journal]{IEEEtran}

\usepackage{amsmath,amssymb,amsfonts}
\usepackage{algorithmic}
\usepackage{graphicx}
\usepackage{textcomp}

\def\ARXIVVERSION{}
\usepackage[utf8]{inputenc}

\usepackage{hyperref}
\usepackage[style=ieee,sorting=none]{biblatex}
\usepackage{array}
\usepackage{tikz}
\usepackage[export]{adjustbox}

\newcommand{\Sectionref}[1]{\hyperref[#1]{Section \ref*{#1}}}
\newcommand{\sectionref}[1]{\hyperref[#1]{section \ref*{#1}}}

\newcommand{\Ie}{I.e\@.}
\newcommand{\ie}{i.e\@.}

\newcommand{\Eg}{E.g\@.}
\newcommand{\eg}{e.g\@.}

\newcommand{\ia}{i.a\@.}

\newcommand{\vs}{vs\@.}
\newcommand{\etc}{etc\@.}
\newcommand{\etal}{\textit{et al\@.}}

\newcommand{\markAuthor}[1]{#1}

\newcommand{\figureWidthModLarge}{1}
\newcommand{\figureWidthModSmall}{1}
\newcommand{\figureWidthModSpecificPoseAngle}{0.8}
\newcommand{\tableFontSize}{}

\ifdefined\ARXIVVERSION
\newcommand{\Description}[1]{}
\fi

\graphicspath{{./img/}}

\begin{document}
\title{Face Image Quality Assessment:\\ A Literature Survey}
\author{
Torsten Schlett,
Christian Rathgeb,
Olaf Henniger,
Javier Galbally,
Julian Fierrez,
and
Christoph Busch%
\thanks{T. Schlett, C. Rathgeb and C. Busch are with the  da/sec - Biometrics and Internet Security Research Group, Hochschule Darmstadt, Germany, \{torsten.schlett, christian.rathgeb, christoph.busch\}@h-da.de}%
\thanks{O. Henniger is with the Fraunhofer Institute for Computer Graphics Research IGD, Darmstadt, Germany, olaf.henniger@igd.fraunhofer.de}%
\thanks{J. Galbally is with the European Commission, Joint Research Center, Ispra, Italy, javier.galbally@ec.europa.eu}%
\thanks{J. Fierrez is with the Universidad Autonoma de Madrid, Madrid, Spain, julian.fierrez@uam.es}%
}

\maketitle

\begin{abstract}
The performance of face analysis and recognition systems depends on the quality of the acquired face data,
which is influenced by numerous factors.
Automatically assessing the quality of face data in terms of biometric utility can thus be useful to
detect low-quality data and make decisions accordingly.
This survey provides an overview of the face image quality assessment literature,
which predominantly focuses on visible wavelength face image input.
A trend towards deep learning based methods is observed,
including notable conceptual differences among the recent approaches,
such as the integration of quality assessment into face recognition models.
Besides image selection,
face image quality assessment can also be used in a variety of other application scenarios, which are discussed herein.
Open issues and challenges are pointed out,
\ia{} highlighting the importance of comparability for algorithm evaluations,
and the challenge for future work to create deep learning approaches that are interpretable in addition to providing accurate utility predictions.
\end{abstract}

\begin{IEEEkeywords}
Biometrics,
biometric sample quality,
face quality assessment,
face recognition.
\end{IEEEkeywords}

\IEEEpeerreviewmaketitle

\nocite{Henniger-FQA-HandcraftedFeatures-BIOSIG-2020}
\nocite{Rose-FQA-FacialAttributes-Springer-2020}
\nocite{Lijun-FQA-MultibranchCNN-ICCT-2019}
\nocite{Rose-FQA-FacialAttributesDeepLearning-ASONAM-2019}
\nocite{Khodabakhsh-FQA-SubjectiveVsObjectiveISO297945Quality-ICBEA-2019}
\nocite{Yu-FQA-LightCNNwithMFM-PRLE-2018}
\nocite{Wang-FQA-SubjectiveRandomForestHybrid-ICCC-2017}
\nocite{Wasnik-FQA-SmartphoneISO297945-IWBF-2017}
\nocite{Zhang-FQA-SubjectiveIlluminationResNet50-ICONIP-2017}
\nocite{Kim-FQA-FaceImageAssessment-ICIP-2015}
\nocite{Damer-FRwithFQA-PersonalizedFaceReferenceVideo-FFER-2015}
\nocite{Abaza-FQA-PhotometricIQA-IET-2014}
\nocite{Kim-FQA-CascadedVideoFrame-ISM-2014}
\nocite{Raghavendra-FQA-ABCVideoPoseGLCM-ICPR-2014}
\nocite{Nikitin-FQA-InVideo-GraphiCon-2014}
\nocite{Phillips-FQA-ExistenceOfFaceQuality-BTAS-2013}
\nocite{Ferrara-FQA-BioLabICAO-TIFS-2012}
\nocite{Hua-FQA-BlurMTF-ICB-2012}
\nocite{Abaza-FQA-QualityMetricsPractical-ICPR-2012}
\nocite{Liao-FQA-GaborCascadeSVM-ICBEB-2012}
\nocite{Nasrollahi-FQA-LowResolutionVideoSequence-TCSVT-2011}
\nocite{Demarsico-FQA-LandmarkPoseLightSymmetry-MiFor-2011}
\nocite{Rizorodriguez-FQA-IlluminationQualityMeasure-ICPR-2010}
\nocite{Beveridge-FQA-LightingAndFocus-CVPRW-2010}
\nocite{Beveridge-FQA-QuoVadisFaceQualityFRVT-IMAVIS-2010}
\nocite{Sang-FQA-StandardGaborIDCT-ICB-2009}
\nocite{Zhang-FQA-AsymmetrySIFT-ISVC-2009}
\nocite{Beveridge-FQA-PredictingFRVTPerformance-FG-2008}
\nocite{Rua-FQAwithFR-VideoFrameSelectionAndScoreNormalization-BioID-2008}
\nocite{Nasrollahi-FQA-InVideoSequences-BioID-2008}
\nocite{Fourney-FQA-VideoFaceImageLogs-CRV-2007}
\nocite{Gao-FQA-StandardizationSampleQualityISO297945-ICB-2007}
\nocite{Abdelmottaleb-FQA-BlurLightPoseExpression-CIM-2007}
\nocite{Hsu-FQA-QualityAssessmentISO197945-BCC-2006}
\nocite{Kryszczuk-FQA-ScoreAndSignalLevelGMM-EUSIPCO-2006}
\nocite{Kryszczuk-FQA-OnFaceImageQualityMeasures-MMUA-2006}
\nocite{Subasic-FQA-ValidationICAO-ISPA-2005}
\nocite{Yang-FQA-PoseVideoFrame-ICPR-2004}
\nocite{Luo-FQA-TrainingbasedNoreferenceIQAA-ICIP-2004}
\nocite{Fu-FQA-DeepInsightMeasuring-WACV-2022}
\nocite{Fu-FQA-FaceMask-FGR-2021}
\nocite{Fu-FQA-RelativeContributionsOfFacialParts-BIOSIG-2021}
\nocite{Chen-FQA-LightQNet-SPL-2021}
\nocite{Meng-FRwithFQA-MagFace-arXiv-2021}
\nocite{Ou-FQA-SimilarityDistributionDistance-arXiv-2021}
\nocite{Chen-FRwithFQA-ProbFace-arXiv-2021}
\nocite{Xie-FQA-PredictiveUncertaintyEstimation-BMVC-2020}
\nocite{Hernandezortega-FQA-FaceQnetV1-2020}
\nocite{Chang-FRwithFQA-UncertaintyLearning-CVPR-2020}
\nocite{Terhorst-FQA-SERFIQ-CVPR-2020}
\nocite{Zhao-FQA-SemiSupervisedCNN-ICCPR-2019}
\nocite{Shi-FRwithFQA-ProbabilisticFaceEmbeddings-ICCV-2019}
\nocite{Hernandezortega-FQA-FaceQnetV0-ICB-2019}
\nocite{Yang-FQA-DFQA-ICIG-2019}
\nocite{Wasnik-FQA-EvaluationSmartphoneCNN-BTAS-2018}
\nocite{Qi-FQA-VideoFrameCNN-ICB-2018}
\nocite{Bestrowden-FQA-FromHumanAssessments-arXiv-2017}
\nocite{Hu-FQA-IlluminationKPLSR-PIC-2016}
\nocite{Vignesh-FQA-VideoCNN-GlobalSIP-2015}
\nocite{Chen-FQA-LearningToRank-SPL-2015}
\nocite{Bharadwaj-FQA-HolisticRepresentations-ICIP-2013}
\nocite{Qu-FQA-GaussianLowPassIllumination-CCIS-2012}
\nocite{Klare-FQA-ImpostorbasedUniqueness-BTAS-2012}
\nocite{Wong-FQA-PatchbasedProbabilistic-CVPRW-2011}
\nocite{Sellahewa-FQA-LuminanceDistortion-TIM-2010}

\section{Introduction}
\label{sec:introduction}

\begin{figure}[t]
    \centering
    \includegraphics[width=\figureWidthModLarge\linewidth]{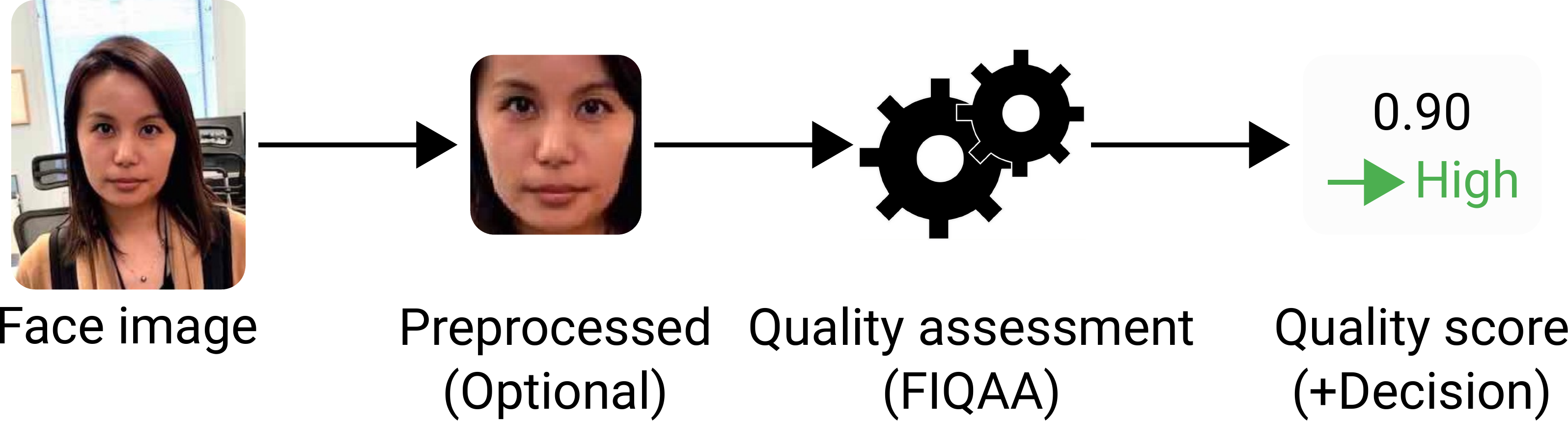}
    \caption{Typical FIQA (Face Image Quality Assessment) process: A face image is preprocessed and a FIQAA (FIQA Algorithm) is applied,
    resulting in a scalar quality score output, based on which a decision can be made.
    Face image taken from \cite{Grother-FQA-4thDraftOngoingFRVT-2021}.}
    \label{fig:fqa-concept}
    \Description{Fully described in the text.}
\end{figure}

\begin{figure}[b]
    \centering
    \includegraphics[width=\figureWidthModSmall\linewidth]{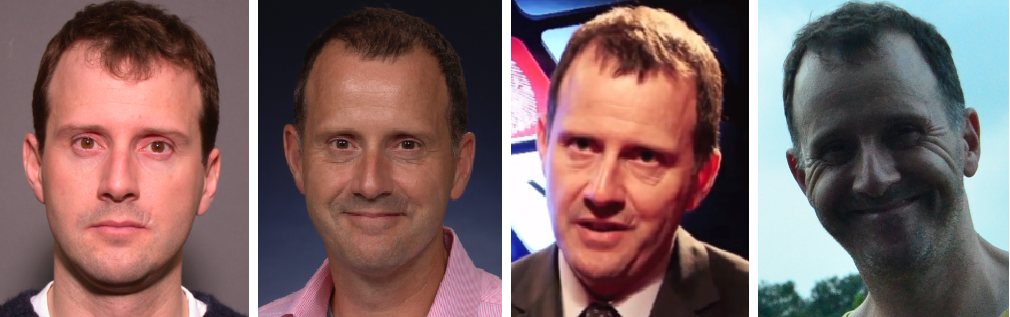}
    \caption{Face images of a single subject with various qualities. Face image quality degrades from left to right as quality degradation factors such as facial expression, pose, and illumination are introduced. Face images taken from \cite{Grother-FQA-4thDraftOngoingFRVT-2021}.}
    \label{fig:quality-degradation-types}
    \Description{Fully described in the text.}
\end{figure}

Face Image Quality Assessment (FIQA) refers to the process of taking a face image as input to produce some form of ``quality'' estimate as output,
as illustrated in \autoref{fig:fqa-concept}.
A FIQA algorithm (FIQAA) is an automated FIQA approach.
See \autoref{fig:quality-degradation-types} for some example images with varying quality.
While FIQA and general Image Quality Assessment (IQA) are overlapping research areas,
there are important distinctions, which we discuss in \autoref{sec:fiqa-vs-iqa}.
Most of the published FIQA literature focuses on single face image input in the visible spectrum.
Therefore, unless otherwise specified in this survey, FIQA(A) refers to single-image Face Image Quality Assessment (Algorithms) in the visible spectrum, with a Quality Score (QS \cite{ISO-IEC-29794-5-TR-FaceQuality-100312}) output that can be
represented by: A) a single scalar value, or B) a vector of quality values measuring different quality-related features.
For a discussion of (F)IQA that instead compares two image variants, \ie{} full/reduced-reference methods, see \autoref{sec:frn-reference-qa}.
Regarding FIQA outside the visible spectrum, see \autoref{sec:unexplored}.

\begin{table}[t]
	\caption{\label{tab:reader-roadmap} Most relevant survey parts for readers with different intent and knowledge background.}
	\centering
	\tableFontSize
\ifdefined\ARXIVVERSION
	\setlength{\tabcolsep}{2pt}
	\begin{tabular}{>{\raggedright\arraybackslash}p{0.55\linewidth}>{\raggedright\arraybackslash}p{0.18\linewidth}>{\raggedright\arraybackslash}p{0.22\linewidth}}
	\hline
	\textbf{Intent of knowledge acquisition} & \textbf{Knowledge background} & \textbf{Relevant parts} \\
	\hline
	Basics (definition, goal, etc.) & Non-expert & Section \ref{sec:introduction} \\
	\hline
	Concepts and categorization \newline (input data, training data, etc.) & Expert & Sections \ref{sec:controlled-unconstrained} to \ref{sec:quality-paradox} and \ref{sec:categorization} \\
	\hline
	Applications & Non-expert  & Section \ref{sec:application-areas} \\
	(use-cases in automated systems) & & \\
	\hline
	Overview of published works (coarse) & Expert & Sections \ref{sec:fiqaa-factor}, \ref{sec:fiqaa-monolithic}, and \ref{sec:summary};
		Tables \ref{tab:datasets}, \ref{tab:fiqaa-factor}, and \ref{tab:fiqaa-monolithic}
		\\
	\hline
	Survey of published works (detailed) & Expert & Sections \ref{sec:fiqaa-literature-factor} and \ref{sec:fiqaa-literature-monolithic} \\
	\hline
	Comparison and evaluation\newline (selective comparison, metrics, etc.) & Expert & Section \ref{sec:evaluation} \\
	\hline
	Open issues and challenges\newline (research directions, problems, etc.) & Non-expert & Sections \ref{sec:challenges} and \ref{sec:summary} \\
	\hline& 
	\end{tabular}
\else
	\setlength{\tabcolsep}{2pt}
	\begin{tabular}{lll}
	\hline
	\textbf{Intent of knowledge acquisition} & \textbf{Knowledge background} & \textbf{Relevant parts} \\
	\hline
	Basics (definition, goal, etc.) & Non-expert & Section \ref{sec:introduction} \\
	\hline
	Concepts and categorization (input data, training data, etc.) & Expert & Sections \ref{sec:controlled-unconstrained} to \ref{sec:quality-paradox} and \ref{sec:categorization} \\
	\hline
	Applications (use-cases in automated systems) & Non-expert  & Section \ref{sec:application-areas} \\
	\hline
	Overview of published works (coarse) & Expert & Sections \ref{sec:fiqaa-factor}, \ref{sec:fiqaa-monolithic}, and \ref{sec:summary};
	Tables \ref{tab:datasets}, \ref{tab:fiqaa-factor}, and \ref{tab:fiqaa-monolithic}
	\\
	\hline
	Survey of published works (detailed) & Expert & Sections \ref{sec:fiqaa-literature-factor} and \ref{sec:fiqaa-literature-monolithic} \\
	\hline
	Comparison and evaluation (selective comparison, metrics, etc.) & Expert & Section \ref{sec:evaluation} \\
	\hline
	Open issues and challenges (research directions, problems, etc.) & Non-expert & Sections \ref{sec:challenges} and \ref{sec:summary} \\
	\hline& 
	\end{tabular}
\fi
\end{table}

The term ``quality'' is an intrinsically subjective concept that can be defined in different ways,
with ISO/IEC 29794-1 \cite{ISO-IEC-29794-1-QualityFramework-160915}
differentiating between three aspects referred to as character, fidelity, and utility.
In the context of facial biometrics these can be described as follows
\cite{Alonsofernandez-QualityMeasures-SecPri-2012}:
\begin{itemize}
\item \textbf{Character}:
Attributes inherent to the source biometric characteristic being acquired (\eg{} the face topography or skin texture)
that cannot be controlled during the biometric acquisition process (\eg{} scars) \cite{ISO-IEC-2382-37-170206}.

\item \textbf{Fidelity}:
For a biometric sample \cite{ISO-IEC-2382-37-170206},
\eg{} a face image,
fidelity reflects the degree of
similarity to its source biometric characteristic \cite{ISO-IEC-29794-1-QualityFramework-160915}.
For instance, a blurred image of a face omits detail and has low fidelity \cite{Grother-FQA-4thDraftOngoingFRVT-2021}.

\item \textbf{Utility}:
The fitness of a sample to accomplish or fulfill the biometric
function (\eg{} face recognition comparison),
which is influenced \ia{} by the character and fidelity \cite{ISO-IEC-2382-37-170206}.
Thus, the term utility is used to indicate the value of an image to a receiving algorithm \cite{Grother-FQA-4thDraftOngoingFRVT-2021}.
\end{itemize}

This survey considers ``utility'' as the primary definition of what a quality score should convey,
which is in accordance to
the quality score definition of ISO/IEC 2382-37 \cite{ISO-IEC-2382-37-170206} and the definition in the ongoing Face Recognition Vendor Test (FRVT) for face image quality assessment \cite{Grother-FQA-4thDraftOngoingFRVT-2021}.
Thus, a QS should be indicative of the Face Recognition (FR) performance.
Note that this entails that the output of a specific FIQAA may be more accurate for a specific FR system,
so the FIQA utility prediction effectivity ultimately depends on the combination of both, the FIQAA and the FR system.
To facilitate interoperability,
it is however desirable that the FIQAA is predictive of recognition performance in general for a range of relevant systems,
instead of being dependent on a single
FR technology.

In short, under this survey's definitions, a FIQAA is typically meant to output a scalar quality score to predict the FR performance from a single face input image.
Being able to predict FR performance without necessarily running an FR algorithm makes FIQA useful for a variety of scenarios,
which are described further in \autoref{sec:application-areas}.
FIQA as a predictor for FR performance has attracted the predominant interest of researchers so far and is thus the main focus in the present survey.
FIQA for other tasks in the field of face biometrics, such as
emotion analysis \cite{Pena-LearningEmotionalBlindedFaceRepresentations-ICPR-2021},
attention level estimation \cite{Daza-mEBAL-MultimodalDatabaseEyeBlinkAttentionLevel-arXiv-2020},
gender or other soft biometrics recognition \cite{Gonzalezsosa-Face-SoftRecognitionWild-TIFS-2018},
\etc{} may open interesting research lines in the future
and can take advantage of current developments that employ FIQA for FR performance prediction.

The contributions of this survey are:
\begin{itemize}
\item
An introduction to FIQA (\autoref{sec:fundamentals}), \ia{} including
the distinction against general IQA (\autoref{sec:fiqa-vs-iqa}),
the conceptual problem with single-image utility assessment (\autoref{sec:quality-paradox}),
and an overview of both common and uncommon FIQA application areas (\autoref{sec:application-areas}).
\item
A categorization of the surveyed FIQA approaches (\autoref{sec:categorization})
with a taxonomy that differentiates between factor-specific and monolithic approaches,
in addition to various other aspects (\autoref{fig:taxonomy}).
\item
A survey of more than 60 FIQAA publications from 2004 to 2021 (\autoref{sec:fiqaa}),
including condensed overview tables for the publications (\autoref{tab:fiqaa-factor}, \autoref{tab:fiqaa-monolithic})
and their used datasets (\autoref{tab:datasets}).
This part is meant for literature overview purposes and does not have to be read in sequence.
\\
Prior work listed varying publication numbers,
with \markAuthor{Hernandez-Ortega \etal{}} \cite{Hernandezortega-FQA-FaceQnetV1-2020} being a recent example that contained a summary for some prior publications ranging from 2006 to 2020.
A fingerprint\slash iris\slash face quality assessment survey by \markAuthor{Bharadwaj \etal{}} \cite{Bharadwaj-Survey-FingerprintIrisFaceQualityAssessment-JIVP-2014} considered less than ten FIQAA publications from 2005 to 2011.
The European JRC-34751 report \cite{Galbally-Face-JRC34751SchengenInformationSystem-EuropeanUnion-2019} also listed some FIQAAs from 2007 to 2018.
To our knowledge this FIQA survey is the most comprehensive one to date.
\item
An introduction for the Error-versus-Reject-Characteristic (ERC) evaluation methodology (\autoref{sec:evaluation-erc}),
which is a standardization candidate in addition to being commonly used in recent FIQA literature,
and a subsequent concrete evaluation that includes a variety FIQA approaches (\autoref{sec:evaluation-concrete}).
The ERC introduction mentions details not considered in recent FIQA literature,
and the evaluation discusses its weaknesses to note opportunities and challenges for future work.
\item
A detailed discussion of various FIQA issues and challenges (\autoref{sec:challenges}),
including avenues for future work.
\end{itemize}

\autoref{tab:reader-roadmap} should allow readers with different intent and background knowledge to quickly identify the most relevant parts of this survey.

\section{Quality Assessment in Face Recognition}
\label{sec:fundamentals}
During enrolment,
a classical face recognition system acquires a reference face image from an individual,
proceeds to pre-process it, including the step of face detection,
and finally extracts a set of features which are stored as reference template.
At the time of authentication a probe face image is captured and processed in the same way
and compared against a reference template of a claimed identity (verification)
or up to all stored reference templates (identification).
Refer to ISO/IEC 2382-37 \cite{ISO-IEC-2382-37-170206} for the standardized vocabulary definitions of terms such as enrolment, templates or references.

\begin{figure}
    \centering
    \includegraphics[width=\figureWidthModSpecificPoseAngle\linewidth]{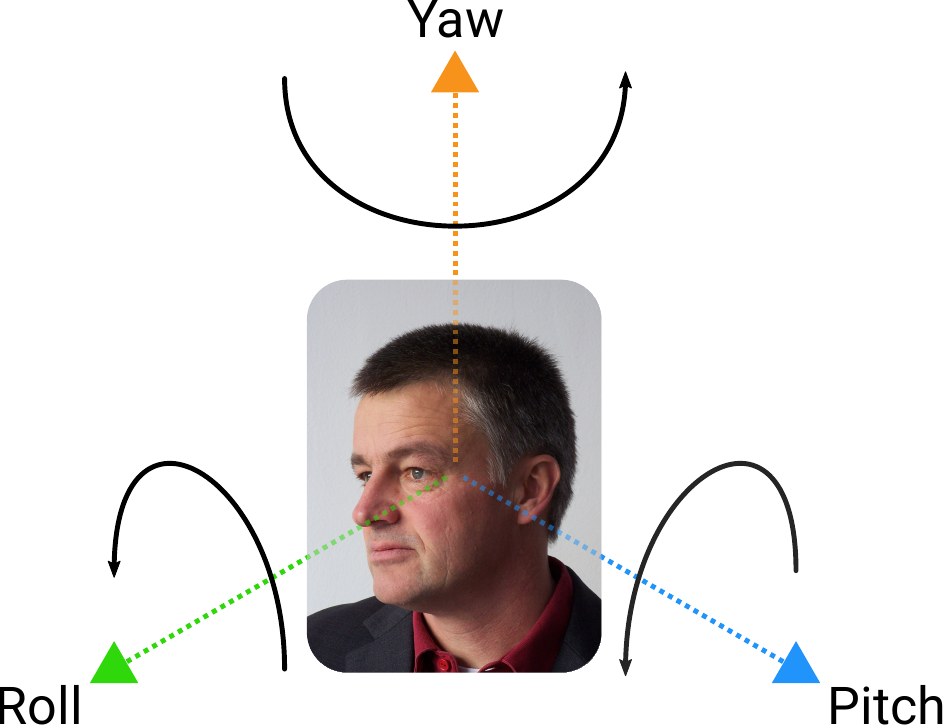}
    \caption{
    Facial pose is usually represented by the pitch, yaw, and roll angles
    defined by ISO/IEC 39794-5 \cite{ISO-IEC-39794-5-FaceInterchangeFormats-191220}.
    Pitch and yaw are also known as tilt and pan.
    A frontal face has $0^{\circ}$ for all three angles.}
    \label{fig:face-pose-angle-pitch-yaw-roll}
    \Description{When viewing a frontal face, the axis vector for the roll angle can be seen as the direction from the head center towards the view position, parallel to some depth axis for the image.
    For the pitch angle, the vector points to the right of the face from the center,
    and the vector for the yaw angle points to the top.
    All angles are measured counter-clockwise for these vectors.}
    \vspace{-1em}
\end{figure}

\subsection{Controlled and Unconstrained Acquisition}
\label{sec:controlled-unconstrained}

Regarding the face image acquisition \cite{ISO-IEC-2382-37-170206},
two different scenarios can be distinguished
\cite{Galbally-Face-JRC34751SchengenInformationSystem-EuropeanUnion-2019}:
\begin{itemize}
\item \textbf{Controlled}: In a controlled scenario, the biometric capture subject is cooperative \cite{ISO-IEC-2382-37-170206}, so that \eg{} the head pose (see \autoref{fig:face-pose-angle-pitch-yaw-roll}) is adjusted to  frontally face the camera with a neutral expression,
and the environmental conditions such as lighting can be controlled.
This is typically the case when face images are acquired for government-issued ID documents.
\item \textbf{Unconstrained}: Here the capture subject is not cooperative, \ie{} the subject is either indifferent \cite{ISO-IEC-2382-37-170206} or intentionally uncooperative \cite{ISO-IEC-2382-37-170206}, and there is no control over the environmental conditions.
Surveillance video FR is an example for this scenario \cite{Proenca-FaceSurveillance-TrendsAndControversies-IntellSyst-2018}.
\end{itemize}
There are other scenarios in between those two extremes,
\eg{} smartphone FR with a cooperative capture subject but incomplete control over the environment \cite{Galbally-Face-JRC34751SchengenInformationSystem-EuropeanUnion-2019},
and the literature usually refers to close-to-optimal capture conditions as ``controlled'', with anything else falling under the ``unconstrained'' category \cite{Galbally-Face-JRC34751SchengenInformationSystem-EuropeanUnion-2019}.
FIQA can be used during controlled acquisition to ensure a certain level of quality by providing immediate feedback.
For unconstrained acquisition, \eg{} via video cameras, FIQA can be used to filter out images below a certain quality level.
While the same FIQAA type and configuration could be used for both,
stricter requirements that are desirable for a controlled government ID image acquisition scenario
may be too strict for unconstrained scenarios.
To facilitate helpful feedback,
FIQA for the controlled scenario preferably should also be able to provide an explanation in terms of multiple separate human-understandable factors,
such as the pose angles (see \autoref{fig:face-pose-angle-pitch-yaw-roll}) or the illumination direction.
In contrast, FIQA for the fully unconstrained scenario by definition cannot benefit from explainability during the acquisition process since there is no control,
\eg{} when automatically deciding whether a video frame is processed further
or not.
However, explainable FIQA can also be beneficial when images are analysed after the acquisition process is complete.
Hence, using FIQA for actionable feedback during a controlled acquisition is just one important application scenario,
while other use cases are independent of the acquisition type.

\subsection{FIQA versus IQA}
\label{sec:fiqa-vs-iqa}

FIQA can be seen as a specific application within the wider field of Image Quality Assessment (IQA), which is a very active research area of image processing.
Even though related to IQA, FIQA has been mainly developed within the biometric context and focuses on distinctive face features.
Consequentially, general IQA algorithms (IQAA) have shown poor performance when directly applied to FIQA,
and, conversely, the very specific FIQA algorithms usually do not generalize to the broader application field of IQA.

General non-biometric IQA typically aims to assess images in terms of subjective (human) perceptual quality,
meaning that technically objective quality scores generated by such IQAAs usually intent to predict or model subjective perceptual quality \cite{Zhai-Survey-PerceptualIQA-2020}.

Biometric FIQA on the other hand is usually concerned with the assessment of the biometric utility for facial biometrics, which can be objectively defined in the context of specific FR systems.
FIQA works may also test or train FIQAAs using ground truth data stemming from human quality assessments,
but for biometric purposes the intent still differs from general perceptual quality assessment,
insofar that the question is how well the images can be used for facial biometrics,
versus how good/undistorted the images look overall for a human.

It can be expected that perceptual quality and biometric utility coincide to some degree,
thus general IQA can be utilized for FIQA as well.
The reverse is less likely, since FIQA algorithms may be specifically developed for face images, so that results for non-face images are not expected to be useful.
This also means that FIQA can perform better for the purpose of biometric utility prediction than a general IQA that has not been developed with facial biometrics in mind.
Some of the surveyed FIQA literature tested known IQA algorithms together with specialized FIQA algorithms.
For instance, \markAuthor{Terhörst \etal{}} \cite{Terhorst-FQA-SERFIQ-CVPR-2020} tested the general IQAAs
BRISQUE \cite{mittalNoReferenceImageQuality2012}, NIQE \cite{mittalMakingCompletelyBlind2013}, and PIQE \cite{venkatanathnBlindImageQuality2015})
together with their fully FR-specialized SER-FIQ FIQAA and three other FIQAAs.

\subsection{Full/Reduced/No-reference Quality Assessment}
\label{sec:frn-reference-qa}

IQA literature draws a distinction between
approaches that require a ``reference'' version of the input and those that do not \cite{Bharadwaj-Survey-FingerprintIrisFaceQualityAssessment-JIVP-2014}\cite{Yang-FQA-DFQA-ICIG-2019}\cite{Hernandezortega-FQA-FaceQnetV1-2020}
(not to be confused with biometric references \cite{ISO-IEC-2382-37-170206}, \eg{} in a FR database):
\begin{itemize}
\item \textbf{Full-reference}: IQA that compares the input image against a known reference version thereof, \ie{} a version that is known to be of higher or equal quality.
Conversely, the input image can be seen as a potentially degraded (\eg{} blurred) version of the reference image.
\item \textbf{Reduced-reference\slash Partial-reference}: Similar to full-reference IQA, a reference version of the input image has to exist first,
but only incomplete information of the reference is known and used for the IQA, \eg{} some statistics of the image.
The distinction between full-reference and reduced-reference approaches is not necessarily clear,
since full-reference approaches may also ``reduce'' their input to a different representation, with information loss, before the comparison step.
\item \textbf{No-reference}: No reference version of the input image is required for the IQA.
Note that such an IQAA can still use other forms of internal data:
An IQAA could \eg{} utilize some fixed set of images unrelated to the input image and still be categorized as no-reference IQA.
Likewise, machine learning IQA models are not automatically classified as reduced-reference IQA just because they incorporate information from training images.
\end{itemize}

\begin{figure}
    \centering
    \includegraphics[width=\figureWidthModSmall\linewidth]{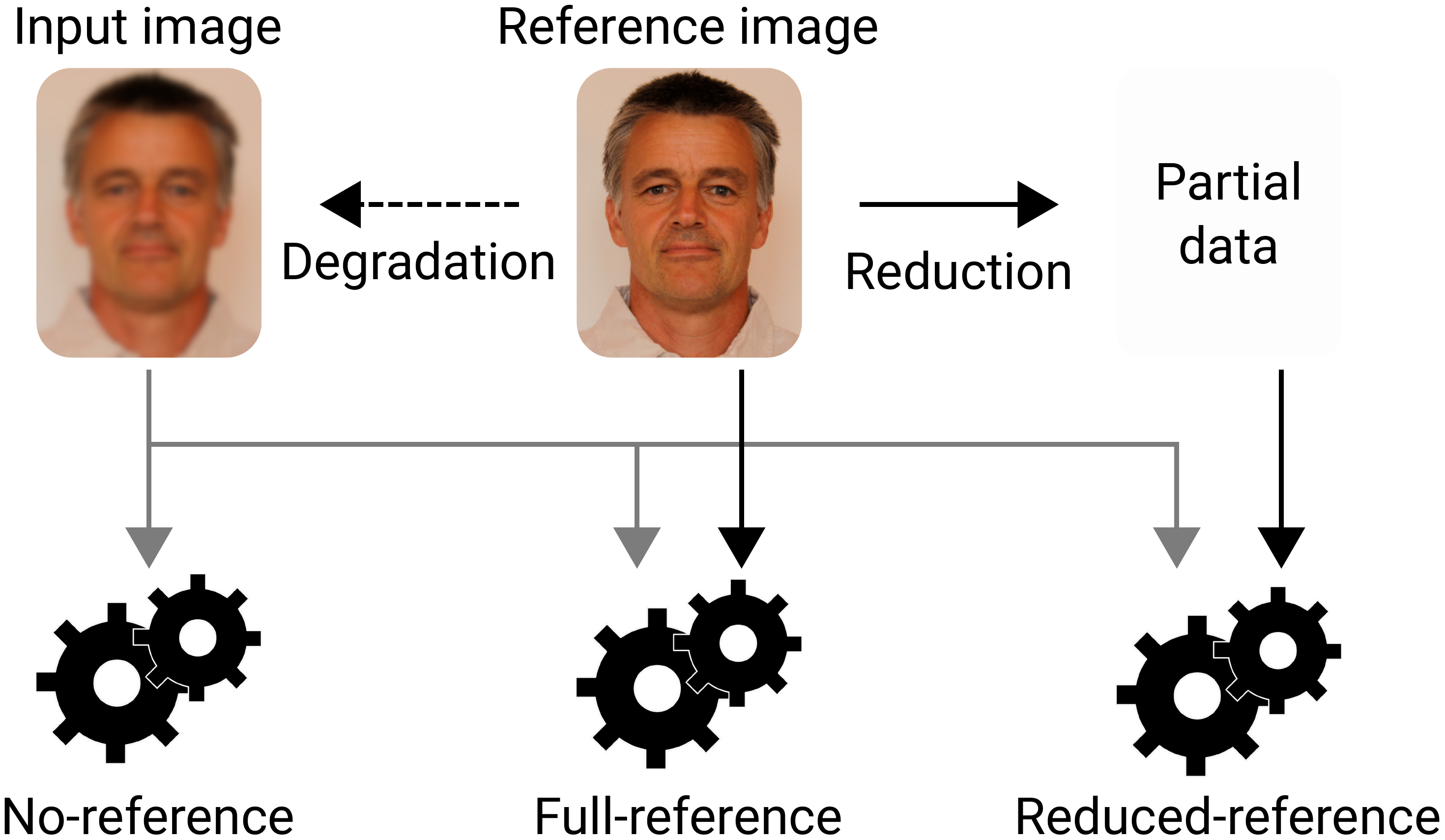}
    \caption{Full-reference, reduced\slash partial-reference, and no-reference quality assessment approaches differ in the used input data, as described in \autoref{sec:fundamentals}.}
    \label{fig:full-reduced-no-reference}
    \Description{Fully described in the text.}
\end{figure}

See \autoref{fig:full-reduced-no-reference} for an illustration of the three concepts.
Full- or reduced-reference approaches are more common and viable for IQA than for FIQA, since both an original and a degraded image exists, \eg{} for an image or video compression scenario \cite{Zhai-Survey-PerceptualIQA-2020} (a use case neglected by FIQA literature so far).
Almost all of the published FIQA literature more specifically considered single-image input FIQA approaches,
which implies no-reference FIQA, and means that no other data specific to the corresponding person (or biometric capture subject \cite{ISO-IEC-2382-37-170206}) is required to facilitate the FIQA.
An outlier is the recent work from \markAuthor{Dihin \etal{}} \cite{Dihin-FullReferenceFIQAandIdentification-JSJU-2020},
which does consider multiple full-reference IQAAs for face images, for both FIQA and for FR.
Note that any FR comparison method can technically fall under the definition of full/reduced-reference (F)IQA
if the comparison scores are repurposed as quality scores.
Furthermore, any full/reduced-reference (F)IQA method can technically be used as a no-reference method
if an image degradation function is added,
such that the single input image serves as the unmodified ``reference'' as well as the degraded input.
Obviously this has less potential for FIQA than specialized approaches.
Nonetheless, this idea has in fact been applied to utilize full-reference IQA
for single-image face presentation attack detection (PAD).
An prominent example for this is the work by \markAuthor{Galbally and Marcel} \cite{Galbally-PAD-BasedOnGeneralIQA-ICPR-2014},
which incorporated various full-reference IQAAs and applied Gaussian filtering as the degradation function,
using the IQAA output to classify the input image as either genuine or as a presentation attack.
Many of these PAD works which are utilizing full-reference IQA appear to use similar IQAA configurations,
and neither FIQA nor FR is their primary concern,
so we do not reference more herein.

\subsection{The Quality Paradox}
\label{sec:quality-paradox}

Usually FIQA algorithms are intended to predict biometric utility for a single biometric sample,
meaning that a single quality score is produced for a single image.
Predicting biometric utility in the context of face recognition implies that the quality score has to indicate the ``accuracy'' or ``certainty'' of comparison scores generated for a sample pair that includes the assessed sample.
Thus, a FIQAA only receives a single sample $S$, which is also part of one or more comparisons with other samples unknown to the FIQAA during the assessment of sample $S$.
This conceptual problem is referred to as the ``quality paradox''.
How FIQA approaches are affected by this quality paradox differs with the concepts:
\begin{itemize}
\item FIQA approaches that only repurpose general IQA methods are already inherently not conceptually linked to FR utility,
\ie{} independently of the quality paradox.
\item FIQA approaches trained on ground truth QSs do have to consider the quality paradox when the ground truth QSs are generated:
\begin{itemize}
\item Relying on human-defined ground truth QSs will generally depend on the subjective assessments,
again technically independent of the quality paradox,
except for human quality assessments that are guided by some protocol (\eg{} collective human FIQA via pairwise comparisons in \cite{Bestrowden-FQA-FromHumanAssessments-arXiv-2017}).
\item For FR-derived ground truth QSs the quality paradox becomes fully relevant,
since the FR comparison pairs have to be selected and the pairwise FR comparison scores have to be transformed into QSs per sample.
Thus, the task of deriving the ground truth QSs itself becomes important to the FIQA design.
Some recent examples of differing ground truth generation approaches are:
\begin{itemize}
\item FaceQnet v0 \cite{Hernandezortega-FQA-FaceQnetV0-ICB-2019}: Normalized comparison score between a target sample and a mated ICAO-compliant (\ie{} assumed high quality) sample as the target sample ground truth QS.
\item FaceQnet v1 \cite{Hernandezortega-FQA-FaceQnetV1-2020}: Extended the FaceQnet v0 \cite{Hernandezortega-FQA-FaceQnetV0-ICB-2019} approach by score fusion for multiple FR systems.
\item PCNet \cite{Xie-FQA-PredictiveUncertaintyEstimation-BMVC-2020}: FIQA model training with loss as the squared difference between the minimum of the predicted per-sample QS for a mated pair of samples and a corresponding FR comparison score.
\item SDD-FIQA \cite{Ou-FQA-SimilarityDistributionDistance-arXiv-2021}: Computed the ground truth QS per sample as the Wasserstein distance between FR comparison score sets for randomly selected mated and non-mated pairs (that each include the sample).
\end{itemize}
\end{itemize}
\item There also exist FIQA approaches that directly use FR models during training/inference without ground truth QS generation,
and approaches that unify FR/FIQA in one model.
While these approaches still technically have to contend with the limits imposed by the quality paradox for single-sample FIQA,
they can more directly estimate the quality (or ``certainty'') of the feature embeddings that the FR model generates.
\end{itemize}
The data aspect categorization described in \autoref{sec:aspect-data} is especially relevant with respect to these considerations.

\subsection{Application Areas of FIQA}
\label{sec:application-areas}

There are various use cases for FIQA:

\begin{itemize}
\item \textbf{Acquisition process threshold}: Face images that result in a quality score below a set threshold can be rejected during the acquisition process \cite{ISO-IEC-2382-37-170206}.
Besides assessing image data stemming directly from cameras,
FIQA could also be applied to measure the impact of printing and scanning,
but among the surveyed literature this was only evaluated indirectly in one work by \markAuthor{Liao \etal{}} \cite{Liao-FQA-GaborCascadeSVM-ICBEB-2012}.

\item \textbf{Acquisition process feedback}: One or multiple FIQAAs may not only be used for image rejection, but also to provide feedback to assist the FR system operator.
\Eg{} individual requirements from
ISO/IEC 39794-5 \cite{ISO-IEC-39794-5-FaceInterchangeFormats-191220},
ICAO \cite{ICAO-PortraitQuality-TR-2018}\cite{ICAO-9303-p9-2015},
or ISO/IEC 19794-5 \cite{ISO-IEC-19794-5-G2-FaceImage-110304}
can be checked and reported automatically when an image is acquired
for FR system enrolment \cite{ISO-IEC-2382-37-170206},
or for passports and other government-issued ID documents.
Capture subjects \cite{ISO-IEC-2382-37-170206}
themselves can also receive immediate feedback for possibly less rigid requirements,
\eg{} during ABC (Automatic Border Control) at airports.

\item \textbf{Quality summarization} \cite{Tabassi-QualitySummarization-NISTIR7422-2007}: Quality can also be monitored by summarizing it over time,
for different capture devices \cite{ISO-IEC-2382-37-170206} or locations \cite{ISO-IEC-29794-1-QualityFramework-160915},
or per user.
This, for instance, enables the identification of defective or underperforming capture devices,
problematic locations, times of day, or seasonal variations,
as well as users that consistently yield low quality samples \cite{Tabassi-QualitySummarization-NISTIR7422-2007}.

\item \textbf{Video frame selection}: Images in a video sequence can be ranked and selected by their assigned quality scores.
This can be used \eg{} to improve both computational performance and recognition performance for identification  via video-surveillance.

\item \textbf{Conditional enhancement}: Optional image enhancement could be applied to images within a certain quality range: Images of sufficiently high quality may not require enhancement, images with very low quality may not be salvageable by enhancement, and images within a medium quality range may be adequate for enhancement.
In addition, multiple enhancement steps could be applied depending on the quality variation after each application,
and different enhancement configurations may be selected for different quality aspects.
While image enhancement could be applied to every image unconditionally,
this could technically degrade\slash falsify otherwise high quality images,
and introduce a significant computational overhead that could make additional hardware necessary (\eg{} GPUs).
The former drawback was shown \eg{} for illumination FIQA by \markAuthor{Rizo-Rodriguez \etal{}} \cite{Rizorodriguez-FQA-IlluminationQualityMeasure-ICPR-2010}.
Likewise, the FIQA application list of
\markAuthor{Hernandez-Ortega \etal{}} \cite{Hernandezortega-FQA-FaceQnetV1-2020} noted \cite{Song-FaceEnhancement-JointHallucinationDeblurring-IJCV-2019} and \cite{Grm-FaceEnhancement-CascadedSuperResolutionIdentityPriors-TIP-2020} as examples for the latter drawback,
with \cite{Song-FaceEnhancement-JointHallucinationDeblurring-IJCV-2019} listing multiple methods taking seconds to minutes,
while \cite{Grm-FaceEnhancement-CascadedSuperResolutionIdentityPriors-TIP-2020} states a requirement of 30ms per single image using a GPU.
Furthermore, multiple images can be selected by quality as a collective basis to construct an improved image - this was done in an enhancement approach stage of the video-focused method by \markAuthor{Nasrollahi and Moeslund} \cite{Nasrollahi-FQA-LowResolutionVideoSequence-TCSVT-2011}.
Lastly, it is also possible to enhance image regions individually depending on region-specific quality scores, which was done in one approach of \markAuthor{Sellahewa and Jassim} \cite{Sellahewa-FQA-LuminanceDistortion-TIM-2010}.

\item \textbf{Compression control}: The change in quality can be measured when an image is compressed in a lossy fashion.
Analogous to conditional enhancement, this measurement can further be used to control the compression, \eg{} by iteratively adjusting the overall compression factor.
Besides the FIQAA literature listed in this survey,
it is also possible to employ full\slash reduced-reference FIQA\slash IQA for this use case,
since a reference is available in the form of the compression input image.

\item \textbf{Database maintenance}:
Existing images in a database can be ranked and filtered by quality. This means that the image with the highest quality can be selected per subject, and that a FR system operator can be notified automatically if a subject has no image of sufficient quality. In systems that do not store images to preserve privacy or storage space, any FIQAA of course needs to be applied beforehand to obtain a quality score (QS).
Furthermore, images or templates \cite{ISO-IEC-2382-37-170206} in the database can be updated in a controlled manner, by comparing the associated QS to the QS of a new image\slash template. This could be done automatically \eg{} after a successful verification.
\markAuthor{Hernandez-Ortega \etal{}} \cite{Hernandezortega-FQA-FaceQnetV1-2020} noted that such updates may also consist of incremental improvements \cite{Asthana-IncrementalFaceAlignmentWild-CVPR-2014}\cite{Didaci-UnsupervisedTemplateUpdate-PRLE-2014}, instead of replacements.
Besides subject-specific incremental improvements,
new quality-controlled data can also be employed to improve biometric models via online learning \cite{Bhatt-FaceClassifierOnlineCotraining-IJCB-2011}\cite{Bharadwaj-Survey-FingerprintIrisFaceQualityAssessment-JIVP-2014}.
Database maintenance, in conjunction with quality summarization/monitoring, is especially relevant in large systems with multiple contributors to a single central database,
such as the European Schengen Information System (SIS), the VISA Information System (VIS), the Entry Exit System (EES), or the US ESTA (Electronic System for Travel Authorization).

\item \textbf{Context switching} \cite{Bharadwaj-Survey-FingerprintIrisFaceQualityAssessment-JIVP-2014}\cite{Hernandezortega-FQA-FaceQnetV1-2020}: A recognition system can adapt to different quality contexts by switching between multiple recognition algorithm configurations (or modes \cite{ISO-IEC-2382-37-170206}), using quality assessment for the
switch activation \cite{Alonsofernandez-MultiBiometrics-QualityBasedConditionalProcessing-SMC-2010}.
Such a strategy does not necessarily have to be applied to a pure FR system - it could also be devised for a multi-modal biometric system \cite{ISO-IEC-2382-37-170206}.

\item \textbf{Quality-weighted fusion} \cite{Bharadwaj-Survey-FingerprintIrisFaceQualityAssessment-JIVP-2014}\cite{Hernandezortega-FQA-FaceQnetV1-2020}:
Similar to full context switching,
a biometric system can fuse scores or decisions
in a weighted fashion based on quality assessments \cite{Fierrez-Fusion-MultipleClassifiersQualityBased-INFFUS-2018}\cite{Singh-Fusion-ComprehensiveOverview-INFFUS-2019}.
Quality-based feature-level fusion for face video frames is considered \eg{} in the surveyed literature \cite{Damer-FRwithFQA-PersonalizedFaceReferenceVideo-FFER-2015} and \cite{Shi-FRwithFQA-ProbabilisticFaceEmbeddings-ICCV-2019}.

\item \textbf{Comparison improvement}: Quality can be used directly as part of FR comparisons \cite{ISO-IEC-2382-37-170206}.
For example, \markAuthor{Shi and Jain} \cite{Shi-FRwithFQA-ProbabilisticFaceEmbeddings-ICCV-2019} computed quality in terms of uncertainty for each FR feature dimension and incorporated it in their comparison algorithm.

\item \textbf{Face detection filter}: In more general terms than video frame selection, FIQA could inherently be used to increase the robustness of face detection by ignoring candidate areas in an image with especially low quality.
This kind of application is however only indirectly examined through the video frame selection works among the surveyed literature.
Conversely, the confidence of face detectors themselves can be
utilized as a type of FIQA, which was used by \markAuthor{Damer \etal{}} \cite{Damer-FRwithFQA-PersonalizedFaceReferenceVideo-FFER-2015}.

\item \textbf{Partial presentation attack avoidance}: Although the surveyed literature does not focus on this application, rejecting or weighing images based on their assessed quality can also reduce the opportunities for presentation attacks \cite{ISO-IEC-2382-37-170206}\cite{Hadid-SystemsUnderSpoofingAttack-SPM-2015},
since accepting images for enrolment or as probe irrespective of their quality could be a potential vulnerability.
FIQA or IQA can also be employed specifically for the purpose of PAD (Presentation Attack Detection) \cite{Galbally-ImageQualityPAD-Iris-Face-Fingerprint-IEEEIP-2014}.
Pure FIQA is however not meant for comprehensive PAD,
because such attacks can consist of data with high biometric utility too.

\item \textbf{Progressive identification}: 
An identification system could conduct searches going progressively from the highest quality reference templates to the lowest quality ones.
Assuming that these templates vary noticeably in quality
and that the search requires an extensive amount of time,
such a strategy can help by showing results with higher confidence (due to higher qualities) early on in the search process.
This could also be used to stop a search early, \ie{} once a number of matches with acceptable certainty has been found.
However, a sufficiently fast identification over the entire database makes such considerations irrelevant,
and this approach is presumably not as useful as general computational workload reduction strategies surveyed by \markAuthor{Drozdowski \etal{}} \cite{Drozdowski-WorkloadSurvey-IET-2019},
since it relies on the existence of exploitable quality variation in the database.
While the listed FIQA literature does not explore this approach,
it does consider FIQA-based computational workload reduction in terms of video frame selection.
Instead of progressing from highest to lowest quality,
\markAuthor{Hernandez-Ortega \etal{}} \cite{Hernandezortega-FQA-FaceQnetV1-2020} noted that the system could use the quality of the probe image to start with comparisons to templates of similar quality,
which may also imply similar acquisition conditions,
and thus could improve the accuracy.
\end{itemize}

\section{Categorization}
\label{sec:categorization}

\newcommand{\attrNondl}{}
\newcommand{\attrDl}{\textbf{\textcolor[HTML]{E53935}{Dl}}}

\newcommand{\attrV}{\textbf{\textcolor[HTML]{d335e5}{V}}}

\newcommand{\attrFc}{\textbf{\textcolor[HTML]{35e561}{Fc}}}
\newcommand{\attrFe}{\textbf{\textcolor[HTML]{35e561}{Fe}}}
\newcommand{\attrFt}{\textbf{\textcolor[HTML]{35e561}{Ft}}}

\newcommand{\attrDhc}{\textbf{\textcolor[HTML]{35a2e5}{Dhc}}}
\newcommand{\attrDuat}{\textbf{\textcolor[HTML]{35a2e5}{Duat}}}
\newcommand{\attrDhgt}{\textbf{\textcolor[HTML]{35a2e5}{Dhgt}}}
\newcommand{\attrDfrt}{\textbf{\textcolor[HTML]{35a2e5}{Dfrt}}}
\newcommand{\attrDfri}{\textbf{\textcolor[HTML]{35a2e5}{Dfri}}}
\newcommand{\attrDint}{\textbf{\textcolor[HTML]{35a2e5}{Dint}}}

\begin{figure*}
    \centering
    \includegraphics[width=\linewidth]{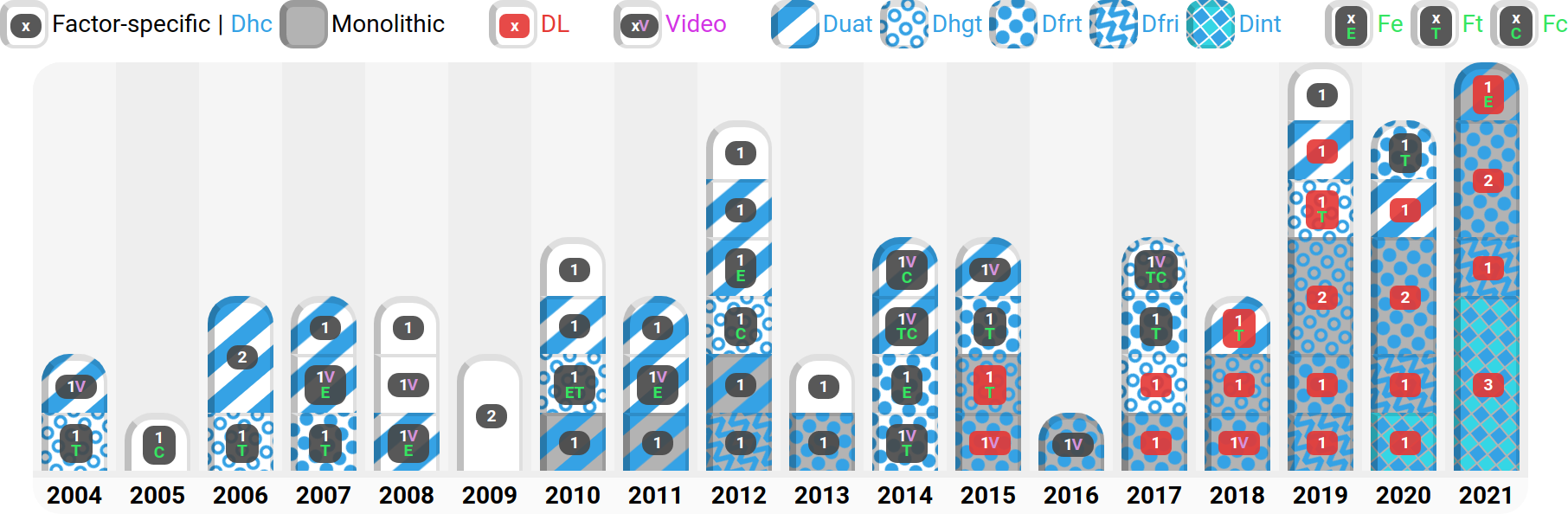}
    \caption{Timeline of the surveyed FIQA literature with categories as depicted by \autoref{fig:taxonomy}. Numbers in the bars denote literature counts.}
    \label{fig:fqa-timeline}
    \Description{Fully described in the text and tables.}
    \vspace{-1em}
\end{figure*}

\begin{figure*}
\centering
\begin{tabular}{cc}
\includegraphics[width=0.6\textwidth,valign=c]{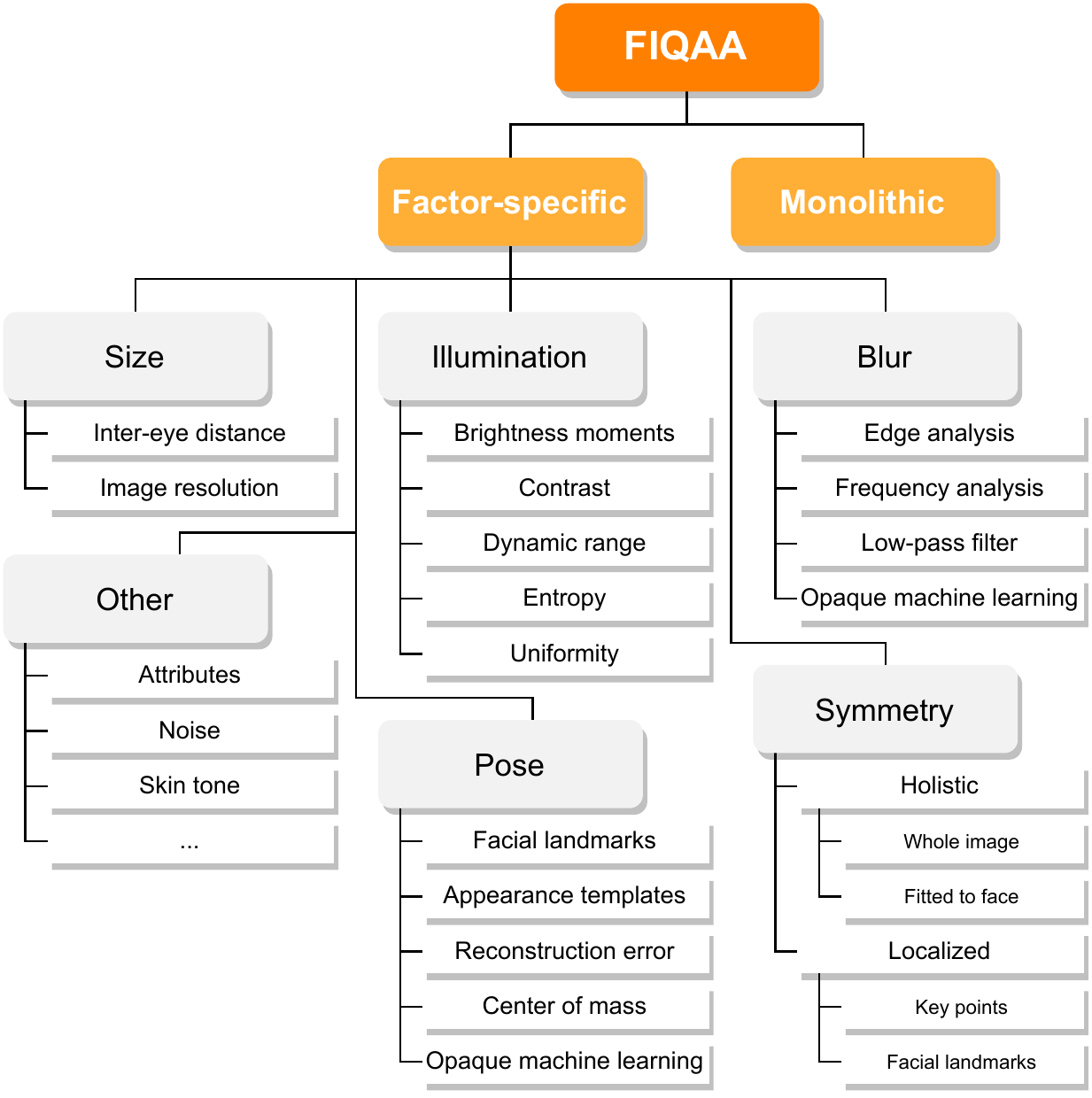} &
\begin{tabular}{ll}
\multicolumn{2}{c}{Data} \\
\attrDhc{} & Hand-crafted \\
\attrDuat{} & Utility-agnostic training \\
\attrDhgt{} & Human ground truth training \\
\attrDfrt{} & FR-based ground truth training \\
\attrDfri{} & FR-based inference \\
\attrDint{} & FR-integration \\
\hline
\multicolumn{2}{c}{Fusion} \\
\attrFe{} & Explicit \\
\attrFt{} & Trained \\
\attrFc{} & Cascade \\
          & None \\
\hline
\multicolumn{2}{c}{Deep Learning} \\
\attrDl{} & Used \\
\attrNondl{}  & Not used \\
\hline
\multicolumn{2}{c}{Video} \\
\attrV{} & Video-frame context \\
         & Single image context \\
\end{tabular}
\end{tabular}
\caption{\label{fig:taxonomy}A taxonomy of the FIQA approaches in the surveyed literature (left), with additional separate aspect-specific categories (right).}
\end{figure*}

The surveyed works are categorized using a taxonomy and several additional aspects.
At the highest level our taxonomy differentiates between
factor-specific FIQA approaches
and monolithic FIQA approaches.
The factor-specific taxonomy branch subdivides methods into categories for interpretable (and typically actionable) factors,
such as blur, which could help an operator to avoid face image deficiencies in a re-capture attempt.
The monolithic approaches produce comparatively opaque assessments/quality scores,
which cannot be immediately interpreted with respect to some concrete separable factor by themselves,
but can indicate overall FR utility.
As described in \autoref{sec:aspect-capture-subject-related},
some of the factor-specific branches can be seen as predominantly capture-related or subject-related.
The subsequent \autoref{sec:aspect-data}, \autoref{sec:aspect-fusion}, \autoref{sec:aspect-dl}, and \autoref{sec:aspect-video}
describe aspects that are assigned per literature
in \autoref{tab:fiqaa-factor} and \autoref{tab:fiqaa-monolithic}.
\autoref{fig:taxonomy} shows an overview of both the taxonomy and the per-literature aspect abbreviations.
The primary approach commonalities are described together with the corresponding literature references
in \autoref{sec:fiqaa-factor} and \autoref{sec:fiqaa-monolithic}.

Note that the taxonomy is meant to group common FIQA approaches in the surveyed literature,
it is not meant to enumerate all feasible FIQA concepts.
Also note that many of the surveyed works described multiple approaches that belong to different categories of the taxonomy.
Some of the surveyed works considered certain quality measure types,
but did not specify a concrete approach,
and are consequently not present in the method-specific reference lists of the taxonomy-describing text passages (\eg{} pose in \cite{Hsu-FQA-QualityAssessmentISO197945-BCC-2006} or \cite{Phillips-FQA-ExistenceOfFaceQuality-BTAS-2013}).

\subsection{Aspect: Capture- and Subject-related FIQA}
\label{sec:aspect-capture-subject-related}

ISO/IEC TR 29794-5:2010 \cite{ISO-IEC-29794-5-TR-FaceQuality-100312} includes an informative facial quality classification scheme that distinguishes between static/dynamic subject characteristics/acquisition process properties.
At the time of writing a standard ISO/IEC 29794-5 is under development,
which will replace the former Technical Report (TR),
and it is intended to further categorize its included factor-specific measures as either capture-related or subject-related.

Capture-related FIQA is influenced by circumstances external to the capture subject, such as the used sensor (\eg{} camera focus, resolution) or the illumination setup.
Subject-related FIQA conversely is influenced by the subject, \eg{} pose, expression, or movement.
While some methods or factors can be predominantly seen as either capture- or subject-related,
others are more obviously influenced by a mixture of capture- and subject-related properties.
This can be mapped directly to the factor-specific categories used in this survey,
instead of individual methods or papers:
\begin{itemize}
\item Size - Inter-eye distance: This is subject-related (distance to camera, facial structure). It is technically capture-related as well, since the camera/image resolution is involved, but that typically is a static acquisition property. \Ie{} it is usually assumed that the camera and its resolution cannot be improved during acquisition, meaning that only the distance to the subject can be adjusted in a re-capture attempt.
\item Size - Image resolution: If the considered image was cropped to the face, then the measure is subject-related similar to inter-eye distance. Otherwise, if the camera's full image resolution is assessed, this factor is fully capture-related.
\item Illumination: Illumination is generally seen as a dynamic acquisition process property \cite{ISO-IEC-29794-5-TR-FaceQuality-100312}, \ie{} capture-related.
But measures may be influenced by subject-related properties too - \eg{} facial hair and skin tone (lighter/darker hair/skin), or possibly pose.
Conversely, it is of course also possible that illumination conditions happen to be sufficiently extreme to disrupt any primarily subject-related measure.
\item Pose: This is predominantly subject-related.
\item Blur: Blur is both capture-related and subject-related, since it can be caused by subject/camera motion, or improper camera configuration.
\item Symmetry: Measures for symmetry depend on symmetric illumination, and most of the surveyed variants implicitly measured frontal pose deviation as well (landmark-based approaches being the exception, although they naturally still rely on a pose that allows landmark detection).
Thus these measures are both capture-related and subject-related.
\end{itemize}
Monolithic approaches can by definition generally be considered as both capture-related and subject-related.

\subsection{Aspect: Data}
\label{sec:aspect-data}

The following data aspect categories are ordered to reflect the degree of FR(-data)-integration or -utilization,
ranging from hand-crafted designs to full FR model integration:
\begin{enumerate}
\item \attrDhc{} - Hand-crafted:
Methods that do not require any training data, except for the optional tuning of parameters such as thresholds.
All of the surveyed approaches belonging to this category are factor-specific,
such as for example the symmetry and blur measures in \cite{Sang-FQA-StandardGaborIDCT-ICB-2009}.

\item \attrDuat{} - Utility-agnostic training:
Methods that require some kind of training data, but do not train to predict ground truth QSs.
Pose angle estimation for FIQA is one example where training may be required,
but where the training does not intend to directly predict utility.
This category also includes approaches that compare the input image against information (\eg{} some image statistics) derived from a training set,
as long as this comparison does not use a FR system.
In this category,
a concrete example for a factor-specific approach is the landmark-based pose estimation in \cite{Demarsico-FQA-LandmarkPoseLightSymmetry-MiFor-2011},
and a concrete example for a monolithic approach is \cite{Qu-FQA-GaussianLowPassIllumination-CCIS-2012}, which compares the input against a fixed averaged image.

\item Ground truth QS training:
Approaches that are trained using ground truth QSs to predict utility or subjective estimates thereof.
\begin{enumerate}
\item \attrDhgt{} - Human ground truth: Works using human assessments for training.
The multi-branch deep learning model in \cite{Lijun-FQA-MultibranchCNN-ICCT-2019} is a factor-specific example in this category,
and the deep learning model trained on human-derived binary quality labels in \cite{Zhao-FQA-SemiSupervisedCNN-ICCPR-2019} is a monolithic example.
\item \attrDfrt{} - FR-based ground truth: Ground truth QSs were derived either via one or multiple FR systems.
A recent factor-specific example for this category is the random forest fusion in \cite{Henniger-FQA-HandcraftedFeatures-BIOSIG-2020},
and a prominent monolithic example is ``FaceQnet'' \cite{Hernandezortega-FQA-FaceQnetV0-ICB-2019}\cite{Hernandezortega-FQA-FaceQnetV1-2020}.
\end{enumerate}

\item \attrDfri{} - FR-based inference:
Approaches that directly utilize FR models during FIQA model training or inference,
without FIQA model training on ground truth QSs.
This obviates a distinction between FR-derived and human-defined ground truth QSs,
although \eg{} the subject identities of the FR training data may still be specified by humans.
The used FR models themselves are not modified with respect to their FR feature inference.
All surveyed approaches in this category are monolithic.
Recent examples are ``SER-FIQ'' \cite{Terhorst-FQA-SERFIQ-CVPR-2020} and ``ProbFace'' \cite{Chen-FRwithFQA-ProbFace-arXiv-2021}.

\item \attrDint{} - FR-integration:
Hybrid FR/FIQA approaches that simultaneously trained FR and FIQA as part of a single integrated system/model,
generating both FR features and quality assessment output during inference.
The only surveyed approaches that fall into this category are the recent monolithic
``data uncertainty learning'' \cite{Chang-FRwithFQA-UncertaintyLearning-CVPR-2020} and ``MagFace'' \cite{Meng-FRwithFQA-MagFace-arXiv-2021}.
Most recently,
the latter has also been included in pure evaluation literature \cite{Fu-FQA-FaceMask-FGR-2021}\cite{Fu-FQA-DeepInsightMeasuring-WACV-2022}.

\end{enumerate}
Many surveyed works considered multiple clearly separable approaches.
Thus, to minimize clutter in the overview tables, each work is marked only with the highest applicable category as per the list order above, \ie{} from \attrDhc{} to \attrDint{}.

\subsection{Aspect: Fusion}
\label{sec:aspect-fusion}

Various works fused multiple separable FIQAAs.
Note that only pure FIQAA fusion methods are marked,
since some surveyed works included approaches that also incorporated non-FIQAA-derived information into the fusion,
such as FR scores \cite{Kryszczuk-FQA-OnFaceImageQualityMeasures-MMUA-2006}
or EXIF data \cite{Phillips-FQA-ExistenceOfFaceQuality-BTAS-2013}.
While the output of fusion methods may be similarly opaque to the output of monolithic FIQAAs,
their input FIQAAs can be (and often were) factor-specific.
\begin{itemize}
\item \attrFe{} - Explicit:
	These approaches derived a single QS from the output of the separable FIQAAs
	by computing weighted sums with manually determined weights
	\cite{Fourney-FQA-VideoFaceImageLogs-CRV-2007}%
	\cite{Nasrollahi-FQA-InVideoSequences-BioID-2008}\cite{Nasrollahi-FQA-LowResolutionVideoSequence-TCSVT-2011},
	or via other hand-crafted fusion functions
	\cite{Rizorodriguez-FQA-IlluminationQualityMeasure-ICPR-2010}%
	\cite{Abaza-FQA-QualityMetricsPractical-ICPR-2012}\cite{Abaza-FQA-PhotometricIQA-IET-2014}%
	\cite{Fu-FQA-RelativeContributionsOfFacialParts-BIOSIG-2021}.
\item \attrFt{} - Trained:
	Trained fusion approaches did likewise include weighted sum computation,
	except with automatically derived weights
	\cite{Nikitin-FQA-InVideo-GraphiCon-2014}%
	\cite{Chen-FQA-LearningToRank-SPL-2015}%
	\cite{Bestrowden-FQA-FromHumanAssessments-arXiv-2017},
	but more often relied on various types of machine learning models such as
	ANNs (Artificial Neural Networks, including deep learning)
	\cite{Luo-FQA-TrainingbasedNoreferenceIQAA-ICIP-2004}%
	\cite{Hsu-FQA-QualityAssessmentISO197945-BCC-2006}%
	\cite{Rizorodriguez-FQA-IlluminationQualityMeasure-ICPR-2010}%
	\cite{Yu-FQA-LightCNNwithMFM-PRLE-2018}%
	\cite{Lijun-FQA-MultibranchCNN-ICCT-2019},
	GMMs (Gaussian Mixture Models)
	\cite{Luo-FQA-TrainingbasedNoreferenceIQAA-ICIP-2004}%
	\cite{Abdelmottaleb-FQA-BlurLightPoseExpression-CIM-2007}%
	\cite{Raghavendra-FQA-ABCVideoPoseGLCM-ICPR-2014},
	AdaBoost
	\cite{Kim-FQA-FaceImageAssessment-ICIP-2015},
	or random forests
	\cite{Wasnik-FQA-SmartphoneISO297945-IWBF-2017}%
	\cite{Wang-FQA-SubjectiveRandomForestHybrid-ICCC-2017}%
	\cite{Henniger-FQA-HandcraftedFeatures-BIOSIG-2020}.
\item \attrFc{} - Cascade:
	Cascaded approaches
	\cite{Subasic-FQA-ValidationICAO-ISPA-2005}%
	\cite{Liao-FQA-GaborCascadeSVM-ICBEB-2012}%
	\cite{Raghavendra-FQA-ABCVideoPoseGLCM-ICPR-2014}%
	\cite{Kim-FQA-CascadedVideoFrame-ISM-2014}%
	\cite{Wang-FQA-SubjectiveRandomForestHybrid-ICCC-2017}
	combined FIQAAs in multiple stages.
	Since the cascade algorithm itself was hand-crafted in all surveyed cases,
	these approaches can be considered as a special kind of explicit fusion.
	The difference to the other explicit fusion methods is that
	these approaches can exit the cascade early in each stage if the quality is deemed to be too low.
	This design can help to reduce the computational workload of the entire quality assessment subsystem
	when many of the input images are of low quality,
	\eg{} in a video frame selection scenario.
	While the FIQAAs within the stages are clearly separable,
	approaches may reuse common data to further improve computational efficiency,
	as done in \cite{Subasic-FQA-ValidationICAO-ISPA-2005}.
	Also, while the per-stage FIQAAs are clearly separable in the sense that they could technically be used as individual FIQAAs,
	the cascaded SVM (Support Vector Machine) approach in
	\cite{Liao-FQA-GaborCascadeSVM-ICBEB-2012}%
	\cite{Wang-FQA-SubjectiveRandomForestHybrid-ICCC-2017}
	trained binary SVM classifiers specifically for the cascaded combination,
	which used the early exits to determine a discrete quality level per stage (1 to 5).
\end{itemize}

\subsection{Aspect: Deep Learning}
\label{sec:aspect-dl}

The surveyed FIQA literature can be broadly categorized into works that do not make use of deep learning for FIQA (``non-DL'') and works that do (``DL'').
Most of the surveyed works overall are non-DL literature,
but the majority of the more recent works are DL literature.
The trend towards DL-based FIQA research is illustrated by the timeline in \autoref{fig:fqa-timeline}.
In the taxonomy
most of the non-DL works belong to the factor-specific branch,
while most DL works can be found under the monolithic category.
Note that non-DL literature does encompass FIQA approaches based on other kinds of machine learning (including shallow artificial neural networks),
as well as purely hand-crafted methods.
The DL literature is marked with ``\attrDl{}'' in the tables.

\subsection{Aspect: Video}
\label{sec:aspect-video}

While face video quality assessment that used temporal inter-frame information
is outside the scope of this face (single-)image quality assessment survey,
we do include video-centric literature that used single-image methods to assess isolated video-frames.
These works are marked with ``\attrV{}'' in the tables to distinguish them from the ``pure'' FIQA works,
but be aware that this does not indicate a technical difference of the FIQAAs themselves.

\section{Face Image Quality Assessment Algorithms}
\label{sec:fiqaa}

\begin{table*}
	\newcommand{\datasetUsageBoth}{\textcolor[HTML]{35a2e5}{B}}
	\newcommand{\datasetUsageEvaluation}{\textcolor[HTML]{E53935}{E}}
	\newcommand{\datasetUsageConstruction}{\textcolor[HTML]{d335e5}{C}}
	\caption{\label{tab:datasets} Datasets that were used in the literature to create
	or evaluate FIQA approaches. In-house datasets or datasets used only for other purposes (such as pure FR model training) are not listed. The left table lists datasets that were used once, and the right table lists datasets used in multiple works.
	The FIQA literature references in the rightmost columns are preceded by markers that denote the usage type:
	\datasetUsageConstruction{} - Dataset used only for FIQAA \textcolor[HTML]{d335e5}{creation} (model training or manual configuration);
\ifdefined\ARXIVVERSION
	\newline
\else
\fi
	\datasetUsageEvaluation{} - Only for FIQAA \textcolor[HTML]{E53935}{evaluation};
\ifdefined\ARXIVVERSION
\else
	\newline
\fi
	\datasetUsageBoth{} - \textcolor[HTML]{35a2e5}{Both} creation \& evaluation.
	}
	\centering
	\tableFontSize
	\setlength{\tabcolsep}{2.18pt}
\begin{tabular}{rcl}
  \hline
  \textbf{Dataset} & \textbf{Year} & \textbf{Used in}\\
  \hline

\mbox{UTKFace \cite{Zhang-Face-AgeProgressionRegression-CVPR-2017}} & 2021 &  \datasetUsageEvaluation{} \cite{Ou-FQA-SimilarityDistributionDistance-arXiv-2021} \\
\mbox{CyberExtruder \cite{CyberExtruderUltimateFaceDataset}} & 2020 &  \datasetUsageEvaluation{} \cite{Hernandezortega-FQA-FaceQnetV1-2020} \\
\mbox{MEDS-I \cite{Curry-MultipleEncounterDatasetI-NIST-2009}} & 2020 &  \datasetUsageBoth{} \cite{Henniger-FQA-HandcraftedFeatures-BIOSIG-2020} \\
\mbox{IJB-S \cite{Kalka-Face-IARPAJanusBenchmarkS-BTAS-2018}} & 2019 &  \datasetUsageEvaluation{} \cite{Shi-FRwithFQA-ProbabilisticFaceEmbeddings-ICCV-2019} \\
\mbox{ImageNet \cite{dengImageNetLargeScaleHierarchical2009}} & 2019 &  \datasetUsageConstruction{} \cite{Yang-FQA-DFQA-ICIG-2019} \\
\mbox{CMU-FIA \cite{Goh-CMUFIAFaceInActionDatabase-AMFG-2005}} & 2018 &  \datasetUsageBoth{} \cite{Qi-FQA-VideoFrameCNN-ICB-2018} \\
\mbox{NCKU face \cite{NCKU-face-database}} & 2018 &  \datasetUsageConstruction{} \cite{Wasnik-FQA-EvaluationSmartphoneCNN-BTAS-2018} \\
\mbox{FIIQD \cite{Zhang-FQA-SubjectiveIlluminationResNet50-ICONIP-2017}} & 2017 &  \datasetUsageBoth{} \cite{Zhang-FQA-SubjectiveIlluminationResNet50-ICONIP-2017} \\
\mbox{Honda/UCSD \cite{Lee-Face-TrackingRecognitionProbabilisticAppearanceManifolds-CVIU-2005}} & 2017 &  \datasetUsageEvaluation{} \cite{Wang-FQA-SubjectiveRandomForestHybrid-ICCC-2017} \\
\mbox{FEI \cite{Thomaz-PCANewRankingMethodFaceImageAnalysis-IMAVIS-2010}} & 2016 &  \datasetUsageBoth{} \cite{Hu-FQA-IlluminationKPLSR-PIC-2016} \\
\mbox{MIT \cite{TurkPentland-EigenfacesRecognition-1991}} & 2016 &  \datasetUsageBoth{} \cite{Hu-FQA-IlluminationKPLSR-PIC-2016} \\
\mbox{AFLW \cite{Kostinger-AnnotatedFacialLandmarksWild-ICCVW-2011}} & 2015 &  \datasetUsageBoth{} \cite{Chen-FQA-LearningToRank-SPL-2015} \\
\mbox{BioLab-ICAO \cite{Ferrara-FQA-BioLabICAO-TIFS-2012}} & 2012 &  \datasetUsageBoth{} \cite{Ferrara-FQA-BioLabICAO-TIFS-2012} \\
\mbox{IIT-NRC \cite{Gorodnichy-FaceRecognition-VideoFrameworkAndDatabaseIITNRC-CRV-2005}} & 2011 &  \datasetUsageEvaluation{} \cite{Nasrollahi-FQA-LowResolutionVideoSequence-TCSVT-2011} \\
\mbox{Pointing'04 \cite{Pointing04Datasets}} & 2011 &  \datasetUsageConstruction{} \cite{Nasrollahi-FQA-LowResolutionVideoSequence-TCSVT-2011} \\
\mbox{XM2VTS \cite{Messer-XM2VTSDBExtendedM2VTS-AVBPA-1999}} & 2010 &  \datasetUsageBoth{} \cite{Rizorodriguez-FQA-IlluminationQualityMeasure-ICPR-2010} \\
\mbox{FRI-CVL \cite{Solina-ColorbasedFaceDetection-Mirage-2003}} & 2008 &  \datasetUsageEvaluation{} \cite{Nasrollahi-FQA-InVideoSequences-BioID-2008} \\
\mbox{HERMES project \cite{HermesProjectFP6IST-027110}} & 2008 &  \datasetUsageEvaluation{} \cite{Nasrollahi-FQA-InVideoSequences-BioID-2008} \\
\mbox{Cohn-Kanade \cite{Kanade-Face-ComprehensiveDatabaseExpression-FGR-2000}} & 2007 &  \datasetUsageBoth{} \cite{Abdelmottaleb-FQA-BlurLightPoseExpression-CIM-2007} \\
\mbox{WVU \cite{Mandala-EffectOfLightingDirectionOnFaceRecognitionPerformance-WVU-2005}} & 2007 &  \datasetUsageBoth{} \cite{Abdelmottaleb-FQA-BlurLightPoseExpression-CIM-2007} \\

  \hline
\end{tabular}
	\setlength{\tabcolsep}{2.18pt}
\ifdefined\ARXIVVERSION
\begin{tabular}{rc>{\raggedright\arraybackslash}p{0.4\linewidth}}
\else
\begin{tabular}{rc>{\raggedright\arraybackslash}p{0.425\linewidth}}
\fi
  \hline
  \textbf{Dataset} & \textbf{Usage timespan} & \textbf{Used in}\\
  \hline

\mbox{LFW \cite{LFWTech}} & 2011 to 2021 & 17:  \datasetUsageBoth{} \cite{Wang-FQA-SubjectiveRandomForestHybrid-ICCC-2017}\allowbreak{}\cite{Bestrowden-FQA-FromHumanAssessments-arXiv-2017}\allowbreak{}\cite{Chen-FQA-LearningToRank-SPL-2015}\allowbreak{} \datasetUsageEvaluation{} \cite{Lijun-FQA-MultibranchCNN-ICCT-2019}\allowbreak{}\cite{Yu-FQA-LightCNNwithMFM-PRLE-2018}\allowbreak{}\cite{Demarsico-FQA-LandmarkPoseLightSymmetry-MiFor-2011}\allowbreak{}\cite{Fu-FQA-DeepInsightMeasuring-WACV-2022}\allowbreak{}\cite{Fu-FQA-RelativeContributionsOfFacialParts-BIOSIG-2021}\allowbreak{}\cite{Chen-FQA-LightQNet-SPL-2021}\allowbreak{}\cite{Meng-FRwithFQA-MagFace-arXiv-2021}\allowbreak{}\cite{Ou-FQA-SimilarityDistributionDistance-arXiv-2021}\allowbreak{}\cite{Chen-FRwithFQA-ProbFace-arXiv-2021}\allowbreak{}\cite{Hernandezortega-FQA-FaceQnetV1-2020}\allowbreak{}\cite{Chang-FRwithFQA-UncertaintyLearning-CVPR-2020}\allowbreak{}\cite{Terhorst-FQA-SERFIQ-CVPR-2020}\allowbreak{}\cite{Shi-FRwithFQA-ProbabilisticFaceEmbeddings-ICCV-2019}\allowbreak{}\cite{Yang-FQA-DFQA-ICIG-2019} \\
\mbox{FERET \cite{Phillips-FERETEvaluationMethodologyFaceRecognition-PAMI-2000}} & 2007 to 2020 & 9:  \datasetUsageBoth{} \cite{Abdelmottaleb-FQA-BlurLightPoseExpression-CIM-2007}\allowbreak{}\cite{Hu-FQA-IlluminationKPLSR-PIC-2016}\allowbreak{}\cite{Chen-FQA-LearningToRank-SPL-2015}\allowbreak{}\cite{Wong-FQA-PatchbasedProbabilistic-CVPRW-2011}\allowbreak{} \datasetUsageConstruction{} \cite{Terhorst-FQA-SERFIQ-CVPR-2020}\allowbreak{} \datasetUsageEvaluation{} \cite{Abaza-FQA-PhotometricIQA-IET-2014}\allowbreak{}\cite{Abaza-FQA-QualityMetricsPractical-ICPR-2012}\allowbreak{}\cite{Demarsico-FQA-LandmarkPoseLightSymmetry-MiFor-2011}\allowbreak{}\cite{Sang-FQA-StandardGaborIDCT-ICB-2009} \\
\mbox{VGGFace2 \cite{Cao-VGGFace2Dataset-FGR-2018}} & 2019 to 2021 & 7:  \datasetUsageBoth{} \cite{Hernandezortega-FQA-FaceQnetV1-2020}\allowbreak{}\cite{Hernandezortega-FQA-FaceQnetV0-ICB-2019}\allowbreak{} \datasetUsageConstruction{} \cite{Xie-FQA-PredictiveUncertaintyEstimation-BMVC-2020}\allowbreak{} \datasetUsageEvaluation{} \cite{Fu-FQA-DeepInsightMeasuring-WACV-2022}\allowbreak{}\cite{Fu-FQA-RelativeContributionsOfFacialParts-BIOSIG-2021}\allowbreak{}\cite{Chen-FRwithFQA-ProbFace-arXiv-2021}\allowbreak{}\cite{Yang-FQA-DFQA-ICIG-2019} \\
\mbox{CASIA-WebFace \cite{Yi-LearningFaceRepresentationFromScratchCASIAWebFace-arXiv-2014}} & 2017 to 2021 & 7:  \datasetUsageBoth{} \cite{Yu-FQA-LightCNNwithMFM-PRLE-2018}\allowbreak{}\cite{Zhao-FQA-SemiSupervisedCNN-ICCPR-2019}\allowbreak{}\cite{Shi-FRwithFQA-ProbabilisticFaceEmbeddings-ICCV-2019}\allowbreak{} \datasetUsageConstruction{} \cite{Ou-FQA-SimilarityDistributionDistance-arXiv-2021}\allowbreak{}\cite{Bestrowden-FQA-FromHumanAssessments-arXiv-2017}\allowbreak{} \datasetUsageEvaluation{} \cite{Lijun-FQA-MultibranchCNN-ICCT-2019}\allowbreak{}\cite{Yang-FQA-DFQA-ICIG-2019} \\
\mbox{CAS-PEAL \cite{Gao-CASPEALLargeScaleChineseFaceDatabase-TSMCA-2008}} & 2009 to 2018 & 6:  \datasetUsageBoth{} \cite{Abaza-FQA-PhotometricIQA-IET-2014}\allowbreak{}\cite{Hu-FQA-IlluminationKPLSR-PIC-2016}\allowbreak{}\cite{Bharadwaj-FQA-HolisticRepresentations-ICIP-2013}\allowbreak{} \datasetUsageConstruction{} \cite{Abaza-FQA-QualityMetricsPractical-ICPR-2012}\allowbreak{}\cite{Wasnik-FQA-EvaluationSmartphoneCNN-BTAS-2018}\allowbreak{} \datasetUsageEvaluation{} \cite{Zhang-FQA-AsymmetrySIFT-ISVC-2009} \\
\mbox{FRGC \cite{Phillips-OverviewFaceRecognitionGrandChallengeFRGC-CVPR-2005}} & 2006 to 2018 & 6:  \datasetUsageBoth{} \cite{Kim-FQA-FaceImageAssessment-ICIP-2015}\allowbreak{}\cite{Kim-FQA-CascadedVideoFrame-ISM-2014}\allowbreak{}\cite{Hsu-FQA-QualityAssessmentISO197945-BCC-2006}\allowbreak{}\cite{Chen-FQA-LearningToRank-SPL-2015}\allowbreak{} \datasetUsageConstruction{} \cite{Raghavendra-FQA-ABCVideoPoseGLCM-ICPR-2014}\allowbreak{}\cite{Wasnik-FQA-EvaluationSmartphoneCNN-BTAS-2018} \\
\mbox{MS-Celeb-1M \cite{Guo-Face-MSCeleb1M-ECCV-2016}} & 2019 to 2020 & 5:  \datasetUsageBoth{} \cite{Shi-FRwithFQA-ProbabilisticFaceEmbeddings-ICCV-2019}\allowbreak{} \datasetUsageConstruction{} \cite{Chang-FRwithFQA-UncertaintyLearning-CVPR-2020}\allowbreak{}\cite{Terhorst-FQA-SERFIQ-CVPR-2020}\allowbreak{}\cite{Yang-FQA-DFQA-ICIG-2019}\allowbreak{} \datasetUsageEvaluation{} \cite{Lijun-FQA-MultibranchCNN-ICCT-2019} \\
\mbox{CFP \cite{Sengupta-Face-CelebritiesFrontalProfile-WACV-2016}} & 2019 to 2021 & 5:  \datasetUsageEvaluation{} \cite{Chen-FQA-LightQNet-SPL-2021}\allowbreak{}\cite{Meng-FRwithFQA-MagFace-arXiv-2021}\allowbreak{}\cite{Chen-FRwithFQA-ProbFace-arXiv-2021}\allowbreak{}\cite{Chang-FRwithFQA-UncertaintyLearning-CVPR-2020}\allowbreak{}\cite{Shi-FRwithFQA-ProbabilisticFaceEmbeddings-ICCV-2019} \\
\mbox{IJB-C \cite{Maze-Face-IARPAJanusBenchmarkC-ICB-2018}} & 2019 to 2021 & 5:  \datasetUsageEvaluation{} \cite{Meng-FRwithFQA-MagFace-arXiv-2021}\allowbreak{}\cite{Ou-FQA-SimilarityDistributionDistance-arXiv-2021}\allowbreak{}\cite{Xie-FQA-PredictiveUncertaintyEstimation-BMVC-2020}\allowbreak{}\cite{Chang-FRwithFQA-UncertaintyLearning-CVPR-2020}\allowbreak{}\cite{Shi-FRwithFQA-ProbabilisticFaceEmbeddings-ICCV-2019} \\
\mbox{YTF \cite{Wolf-YouTubeFacesRecognitionUnconstrained-CVPR-2011}} & 2014 to 2020 & 5:  \datasetUsageBoth{} \cite{Nikitin-FQA-InVideo-GraphiCon-2014}\allowbreak{} \datasetUsageEvaluation{} \cite{Yu-FQA-LightCNNwithMFM-PRLE-2018}\allowbreak{}\cite{Damer-FRwithFQA-PersonalizedFaceReferenceVideo-FFER-2015}\allowbreak{}\cite{Chang-FRwithFQA-UncertaintyLearning-CVPR-2020}\allowbreak{}\cite{Shi-FRwithFQA-ProbabilisticFaceEmbeddings-ICCV-2019} \\
\mbox{MS1MV2 \cite{Deng-ArcFace-IEEE-CVPR-2019}} & 2021 & 4:  \datasetUsageConstruction{} \cite{Chen-FQA-LightQNet-SPL-2021}\allowbreak{}\cite{Meng-FRwithFQA-MagFace-arXiv-2021}\allowbreak{}\cite{Ou-FQA-SimilarityDistributionDistance-arXiv-2021}\allowbreak{}\cite{Chen-FRwithFQA-ProbFace-arXiv-2021} \\
\mbox{IJB-A \cite{Klare-Face-IARPAJanusBenchmarkA-CVPR-2015}} & 2017 to 2019 & 4:  \datasetUsageBoth{} \cite{Lijun-FQA-MultibranchCNN-ICCT-2019}\allowbreak{} \datasetUsageConstruction{} \cite{Yang-FQA-DFQA-ICIG-2019}\allowbreak{} \datasetUsageEvaluation{} \cite{Shi-FRwithFQA-ProbabilisticFaceEmbeddings-ICCV-2019}\allowbreak{}\cite{Bestrowden-FQA-FromHumanAssessments-arXiv-2017} \\
\mbox{ChokePoint \cite{Wong-FQA-PatchbasedProbabilistic-CVPRW-2011}} & 2011 to 2018 & 4:  \datasetUsageBoth{} \cite{Qi-FQA-VideoFrameCNN-ICB-2018}\allowbreak{}\cite{Vignesh-FQA-VideoCNN-GlobalSIP-2015}\allowbreak{} \datasetUsageEvaluation{} \cite{Wasnik-FQA-EvaluationSmartphoneCNN-BTAS-2018}\allowbreak{}\cite{Wong-FQA-PatchbasedProbabilistic-CVPRW-2011} \\
\mbox{SCface \cite{Grgic-SCfaceSurveillanceCamerasFaceDatabase-MTA-2011}} & 2011 to 2018 & 4:  \datasetUsageBoth{} \cite{Bharadwaj-FQA-HolisticRepresentations-ICIP-2013}\allowbreak{} \datasetUsageEvaluation{} \cite{Demarsico-FQA-LandmarkPoseLightSymmetry-MiFor-2011}\allowbreak{}\cite{Wasnik-FQA-EvaluationSmartphoneCNN-BTAS-2018}\allowbreak{}\cite{Chen-FQA-LearningToRank-SPL-2015} \\
\mbox{Extended Yale \cite{Lee-AcquiringLinearSubspacesFaceRecognitionVariableLighting-PAMI-2005}} & 2010 to 2018 & 4:  \datasetUsageBoth{} \cite{Rizorodriguez-FQA-IlluminationQualityMeasure-ICPR-2010}\allowbreak{}\cite{Qu-FQA-GaussianLowPassIllumination-CCIS-2012}\allowbreak{}\cite{Sellahewa-FQA-LuminanceDistortion-TIM-2010}\allowbreak{} \datasetUsageConstruction{} \cite{Wasnik-FQA-EvaluationSmartphoneCNN-BTAS-2018} \\
\mbox{CPLFW \cite{Zheng-Face-CrossPoseLFW-BUPT-2018}} & 2021 & 3:  \datasetUsageEvaluation{} \cite{Chen-FQA-LightQNet-SPL-2021}\allowbreak{}\cite{Meng-FRwithFQA-MagFace-arXiv-2021}\allowbreak{}\cite{Chen-FRwithFQA-ProbFace-arXiv-2021} \\
\mbox{IJB-B \cite{Whitelam-Face-IARPAJanusBenchmarkB-CVPRW-2017}} & 2021 & 3:  \datasetUsageEvaluation{} \cite{Chen-FQA-LightQNet-SPL-2021}\allowbreak{}\cite{Meng-FRwithFQA-MagFace-arXiv-2021}\allowbreak{}\cite{Chen-FRwithFQA-ProbFace-arXiv-2021} \\
\mbox{Adience \cite{Eidinger-AgeGenderEstimationUnfilteredFaces-TIFS-2014}} & 2020 to 2021 & 3:  \datasetUsageEvaluation{} \cite{Chen-FQA-LightQNet-SPL-2021}\allowbreak{}\cite{Ou-FQA-SimilarityDistributionDistance-arXiv-2021}\allowbreak{}\cite{Terhorst-FQA-SERFIQ-CVPR-2020} \\
\mbox{BioSecure \cite{OrtegaGarcia-BioSecureDatabase-IEEE-PAMI-2010}} & 2019 to 2021 & 3:  \datasetUsageEvaluation{} \cite{Fu-FQA-DeepInsightMeasuring-WACV-2022}\allowbreak{}\cite{Hernandezortega-FQA-FaceQnetV1-2020}\allowbreak{}\cite{Hernandezortega-FQA-FaceQnetV0-ICB-2019} \\
\mbox{GBU \cite{Phillips-IntroductionGoodBadUglyFaceRecognitionChallenge-FGR-2011}} & 2012 to 2014 & 3:  \datasetUsageBoth{} \cite{Abaza-FQA-PhotometricIQA-IET-2014}\allowbreak{}\cite{Abaza-FQA-QualityMetricsPractical-ICPR-2012}\allowbreak{} \datasetUsageEvaluation{} \cite{Phillips-FQA-ExistenceOfFaceQuality-BTAS-2013} \\
\mbox{AT\&T \cite{Samaria-Face-ParameterisationStochasticModel-ACV-1994}} & 2010 to 2016 & 3:  \datasetUsageBoth{} \cite{Hu-FQA-IlluminationKPLSR-PIC-2016}\allowbreak{}\cite{Sellahewa-FQA-LuminanceDistortion-TIM-2010}\allowbreak{} \datasetUsageConstruction{} \cite{Raghavendra-FQA-ABCVideoPoseGLCM-ICPR-2014} \\
\mbox{CMU-PIE \cite{Sim-Face-CMUPoseIlluminationExpressionDatabase-PAMI-2003}} & 2009 to 2011 & 3:  \datasetUsageConstruction{} \cite{Beveridge-FQA-LightingAndFocus-CVPRW-2010}\allowbreak{} \datasetUsageEvaluation{} \cite{Sang-FQA-StandardGaborIDCT-ICB-2009}\allowbreak{}\cite{Wong-FQA-PatchbasedProbabilistic-CVPRW-2011} \\
\mbox{FRVT 2006 \cite{Phillips-FaceIris-FRVT2006ICE2006LargeScaleResults-NIST-2007}} & 2008 to 2010 & 3:  \datasetUsageEvaluation{} \cite{Beveridge-FQA-LightingAndFocus-CVPRW-2010}\allowbreak{}\cite{Beveridge-FQA-QuoVadisFaceQualityFRVT-IMAVIS-2010}\allowbreak{}\cite{Beveridge-FQA-PredictingFRVTPerformance-FG-2008} \\
\mbox{Yale \cite{Georghiades-Face-IlluminationConeModelsVariableLightingPose-PAMI-2001}} & 2007 to 2014 & 3:  \datasetUsageBoth{} \cite{Abaza-FQA-PhotometricIQA-IET-2014}\allowbreak{}\cite{Abaza-FQA-QualityMetricsPractical-ICPR-2012}\allowbreak{} \datasetUsageEvaluation{} \cite{Gao-FQA-StandardizationSampleQualityISO297945-ICB-2007} \\
\mbox{BANCA \cite{Baillybailliere-BANCADatabaseAndEvaluationProtocol-AVBPA-2003}} & 2006 to 2008 & 3:  \datasetUsageBoth{} \cite{Kryszczuk-FQA-ScoreAndSignalLevelGMM-EUSIPCO-2006}\allowbreak{}\cite{Kryszczuk-FQA-OnFaceImageQualityMeasures-MMUA-2006}\allowbreak{} \datasetUsageEvaluation{} \cite{Rua-FQAwithFR-VideoFrameSelectionAndScoreNormalization-BioID-2008} \\
\mbox{AgeDB \cite{Moschoglou-Face-AgeDB-CVPRW-2017}} & 2021 & 2:  \datasetUsageEvaluation{} \cite{Meng-FRwithFQA-MagFace-arXiv-2021}\allowbreak{}\cite{Chen-FRwithFQA-ProbFace-arXiv-2021} \\
\mbox{CALFW \cite{Zheng-Face-CrossAgeLFW-arXiv-2017}} & 2021 & 2:  \datasetUsageEvaluation{} \cite{Meng-FRwithFQA-MagFace-arXiv-2021}\allowbreak{}\cite{Chen-FRwithFQA-ProbFace-arXiv-2021} \\
\mbox{MEDS-II \cite{Founds-MultipleEncounterDatasetII-NISTIR7807-2011}} & 2019 to 2020 & 2:  \datasetUsageBoth{} \cite{Rose-FQA-FacialAttributes-Springer-2020}\allowbreak{}\cite{Rose-FQA-FacialAttributesDeepLearning-ASONAM-2019} \\
\mbox{MegaFace \cite{Kemelmachershlizerman-Face-MegaFaceBenchmark-CVPR-2016}} & 2019 to 2020 & 2:  \datasetUsageEvaluation{} \cite{Chang-FRwithFQA-UncertaintyLearning-CVPR-2020}\allowbreak{}\cite{Shi-FRwithFQA-ProbabilisticFaceEmbeddings-ICCV-2019} \\
\mbox{AR \cite{Martinez-ARFaceDatabase-CVC-1998}} & 2014 to 2018 & 2:  \datasetUsageConstruction{} \cite{Raghavendra-FQA-ABCVideoPoseGLCM-ICPR-2014}\allowbreak{}\cite{Wasnik-FQA-EvaluationSmartphoneCNN-BTAS-2018} \\
\mbox{PaSC \cite{Beveridge-FaceRecognitionDigitalPointAndShootCameras-BTAS-2013}} & 2013 to 2018 & 2:  \datasetUsageBoth{} \cite{Qi-FQA-VideoFrameCNN-ICB-2018}\allowbreak{} \datasetUsageEvaluation{} \cite{Phillips-FQA-ExistenceOfFaceQuality-BTAS-2013} \\
\mbox{MBGC \cite{Phillips-OverviewMultipleBiometricsGrandChallengeMBGC-ICB-2009}} & 2012 to 2014 & 2:  \datasetUsageEvaluation{} \cite{Abaza-FQA-PhotometricIQA-IET-2014}\allowbreak{}\cite{Abaza-FQA-QualityMetricsPractical-ICPR-2012} \\
\mbox{Q-FIRE \cite{Johnson-QualityFaceIrisResearchEnsembleQFIRE-BTAS-2010}} & 2012 to 2014 & 2:  \datasetUsageEvaluation{} \cite{Abaza-FQA-PhotometricIQA-IET-2014}\allowbreak{}\cite{Hua-FQA-BlurMTF-ICB-2012} \\

  \hline
\end{tabular}
\end{table*}

The following subsections and tables are divided into the factor-specific and monolithic categories introduced in \autoref{sec:categorization}.
For each there is one subsection that highlights the overarching commonalities/differences (factor-specific \autoref{sec:fiqaa-factor}, monolithic \autoref{sec:fiqaa-monolithic}),
followed by a corresponding subsection with introductions for all of the surveyed works in chronological order (factor-specific \autoref{sec:fiqaa-literature-factor}, monolithic \autoref{sec:fiqaa-literature-monolithic}).
\autoref{tab:fiqaa-factor} (factor-specific) and \autoref{tab:fiqaa-monolithic} (monolithic) provide a condensed overview of the literature,
and show the categorization of the works for every aspect listed in \autoref{fig:taxonomy}.
\autoref{tab:datasets} additionally lists the datasets used to develop and evaluate the FIQA approaches of the surveyed literature.
The implications of the dataset variety are discussed in \autoref{sec:comparability}.

The surveyed FIQA works have been developed by a large variety of research groups.
Independently of author relationships,
various FIQA works are clearly based on prior work,
which is noted both in the introductory literature text and the overview tables.

\newcommand{\commonalityPart}[1]{\item #1:}

\subsection{Factor-specific - Commonalities}
\label{sec:fiqaa-factor}

\begin{table*}
	\caption{\label{tab:fiqaa-factor} Factor-specific FIQA literature in reverse chronological order.}
	\centering
	\tableFontSize
	\setlength{\tabcolsep}{1.9pt}\begin{tabular}{>{\raggedright\arraybackslash}p{0.075\linewidth}>{\raggedright\arraybackslash}p{0.085\linewidth}>{\raggedright\arraybackslash}p{0.58\linewidth}>{\raggedright\arraybackslash}p{0.224\linewidth}}
\hline
\textbf{Reference} & \textbf{Aspects} & \textbf{Method(s)} & \textbf{Datasets} \\
\hline

2020 \cite{Henniger-FQA-HandcraftedFeatures-BIOSIG-2020} & \attrNondl{}\attrDfrt{}\attrFt{} & 17 hand-crafted ISO/IEC TR 29794-5:2010 \cite{ISO-IEC-29794-5-TR-FaceQuality-100312} related measures: Left-right symmetry $\times 7$, capture-related $\times 10$; 11 fused as random forests. \textit{2 black-box COTS systems for FR.} & MEDS-I \\\hline
2020 \cite{Rose-FQA-FacialAttributes-Springer-2020} & \attrDl{}\attrDuat{} & 3 binary attributes (Eyes open, glasses, frontal); Non-DL: 23 models, \ia{} SVMs; DL: Pretrained AlexNet \cite{krizhevskyImageNetClassificationDeep2017}, GoogLeNet \cite{szegedyGoingDeeperConvolutions2014}. \textit{Best results via SVM+DL score-level fusion.} & In-house, MEDS-II \\\hline
2019 \cite{Lijun-FQA-MultibranchCNN-ICCT-2019} & \attrDl{}\attrDhgt{}\attrFt{} & Multi-branch CNN trained for 4 factors: Alignment, Occlusion, Pose, Blur (+ fused overall QS); QS ground truths manually annotated for 3000 images. \textit{} & IJB-A, MS-Celeb-1M, CASIA-WebFace, LFW \\\hline
2019 \cite{Rose-FQA-FacialAttributesDeepLearning-ASONAM-2019} & \attrDl{}\attrDuat{} & Same as \cite{Rose-FQA-FacialAttributes-Springer-2020}, but \ia{} with smartphone images. \textit{Continuation of \cite{Rose-FQA-FacialAttributes-Springer-2020}.} & In-house, MEDS-II \\\hline
2019 \cite{Khodabakhsh-FQA-SubjectiveVsObjectiveISO297945Quality-ICBEA-2019} & \attrNondl{}\attrDhc{} & 8 factors compared to mean scores from 26 humans. \textit{Continuation of \cite{Wasnik-FQA-SmartphoneISO297945-IWBF-2017}.} & In-house (Smartphone) \\\hline
2018 \cite{Yu-FQA-LightCNNwithMFM-PRLE-2018} & \attrDl{}\attrDfrt{}\attrFt{} & CNN with MFM\cite{wuLightCNNDeep2018} \& NIN\cite{linNetworkNetwork2014} layers, trained using 15 synthetic degradation classes (5 types $\times$ 3 settings). \textit{} & CASIA-WebFace, LFW, YTF \\\hline
2017 \cite{Wang-FQA-SubjectiveRandomForestHybrid-ICCC-2017} & \attrNondl{}\attrDhgt{}\attrFc{}\attrFt{} & Subjective QS random forest, 7 hand-crafted features. \textit{} & LFW, Honda/UCSD \\\hline
2017 \cite{Wasnik-FQA-SmartphoneISO297945-IWBF-2017} & \attrNondl{}\attrDfrt{}\attrFt{} & 9 factors plus random forest: Lighting symmetry, Pose symmetry, Brightness, Image contrast, Global Contrast Factor, Exposure, Blur, Sharpness, Vertical edge density. \textit{} & In-house (Smartphone) \\\hline
2017 \cite{Zhang-FQA-SubjectiveIlluminationResNet50-ICONIP-2017} & \attrDl{}\attrDhgt{} & ResNet-50 trained on subjective illumination QSs. \textit{Open source.} & FIIQD \\\hline
2015 \cite{Kim-FQA-FaceImageAssessment-ICIP-2015} & \attrNondl{}\attrDfrt{}\attrFt{} & AdaBoost on 3 ``objective'' measures \cite{Kim-FQA-CascadedVideoFrame-ISM-2014} + optional training-set-``relative'' measures. \textit{Continuation of \cite{Kim-FQA-CascadedVideoFrame-ISM-2014}.} & FRGC \\\hline
2015 \cite{Damer-FRwithFQA-PersonalizedFaceReferenceVideo-FFER-2015} & \attrNondl{}\attrDuat{}\attrV{} & Entropy, Viola-Jones \cite{Viola-RapidObjectDetection-CVPR-2001} face detection confidence. \textit{} & YTF \\\hline
2014 \cite{Abaza-FQA-PhotometricIQA-IET-2014} & \attrNondl{}\attrDfrt{}\attrFe{} & ANN on 5 factors\slash 7 measures equivalent to \cite{Abaza-FQA-QualityMetricsPractical-ICPR-2012} \vs{} logistic regression, SVR, and 10 normalization\slash fusion combinations. \textit{Continuation of \cite{Abaza-FQA-QualityMetricsPractical-ICPR-2012}.} & CAS-PEAL, Yale, GBU, FERET, MBGC, Q-FIRE \\\hline
2014 \cite{Kim-FQA-CascadedVideoFrame-ISM-2014} & \attrNondl{}\attrDuat{}\attrFc{}\attrV{} & Pose\slash Alignment (Reconstruction), Blur, Illumination. \textit{} & FRGC \\\hline
2014 \cite{Raghavendra-FQA-ABCVideoPoseGLCM-ICPR-2014} & \attrNondl{}\attrDuat{}\attrFc{}\attrFt{}\attrV{} & Two stages: 1. Pose (yaw\slash roll), 2. 12 GLCM features \cite{haralickTexturalFeaturesImage1973} fed into a GMM. \textit{} & In-house (ABC), FRGC, AR, AT\&T \\\hline
2014 \cite{Nikitin-FQA-InVideo-GraphiCon-2014} & \attrNondl{}\attrDfrt{}\attrFt{}\attrV{} & Facial symmetry, Illumination, Blur, Resolution. \textit{} & YTF \\\hline
2013 \cite{Phillips-FQA-ExistenceOfFaceQuality-BTAS-2013} & \attrNondl{}\attrDhc{} & 9 FIQAA, \ia{} Illumination (Direction), SEMC focus \cite{Beveridge-FQA-LightingAndFocus-CVPRW-2010}, Edge density \cite{Beveridge-FQA-QuoVadisFaceQualityFRVT-IMAVIS-2010}, \dots{}, and SVM vs. GPO oracle. \textit{Continuation of \cite{Beveridge-FQA-LightingAndFocus-CVPRW-2010}.} & Unknown, GBU, PaSC \\\hline
2012 \cite{Ferrara-FQA-BioLabICAO-TIFS-2012} & \attrNondl{}\attrDuat{} & 30 factors, \ia{} Hair Across Eyes, Looking Away, Varied Background. \textit{} & BioLab-ICAO \\\hline
2012 \cite{Hua-FQA-BlurMTF-ICB-2012} & \attrNondl{}\attrDhc{} & Blur (MTF vs\@.: ED \cite{weberQualityMeasuresFace2006}, LoG, SG, DCT). \textit{} & Q-FIRE \\\hline
2012 \cite{Abaza-FQA-QualityMetricsPractical-ICPR-2012} & \attrNondl{}\attrDuat{}\attrFe{} & 12 measures: Sharpness $\times 4$, Contrast $\times 2$, Illumination $\times 2$, Focus $\times 2$, Brightness $\times 2$; Combined FIQAA with 7 factors. \textit{} & CAS-PEAL, Yale, GBU, FERET, MBGC \\\hline
2012 \cite{Liao-FQA-GaborCascadeSVM-ICBEB-2012} & \attrNondl{}\attrDhgt{}\attrFc{} & 5-class cascade SVM with Gabor magnitude features. \textit{} & In-house \\\hline
2011 \cite{Nasrollahi-FQA-LowResolutionVideoSequence-TCSVT-2011} & \attrNondl{}\attrDuat{}\attrFe{}\attrV{} & Pose (Linear Auto-associative Neural Networks), Illumination, Blur, Resolution. \textit{QS relative to face image sequence. Continuation of \cite{Nasrollahi-FQA-InVideoSequences-BioID-2008}.} & In-house (100 videos), Pointing'04, IIT-NRC \\\hline
2011 \cite{Demarsico-FQA-LandmarkPoseLightSymmetry-MiFor-2011} & \attrNondl{}\attrDuat{} & Landmark-based: Pose (Yaw\slash pitch\slash roll), Illumination (Histogram mass center variance), Symmetry (Lines). \textit{} & FERET, LFW, SCface \\\hline
2010 \cite{Rizorodriguez-FQA-IlluminationQualityMeasure-ICPR-2010} & \attrNondl{}\attrDhgt{}\attrFe{}\attrFt{} & Illumination of triangle mesh regions (Mean, ANN-weighted, Combined). \textit{} & Extended Yale, XM2VTS \\\hline
2010 \cite{Beveridge-FQA-LightingAndFocus-CVPRW-2010} & \attrNondl{}\attrDhc{} & Illumination (Direction), SEMC focus, Edge density. \textit{Continuation of \cite{Beveridge-FQA-QuoVadisFaceQualityFRVT-IMAVIS-2010}.} & FRVT 2006, CMU-PIE \\\hline
2010 \cite{Beveridge-FQA-QuoVadisFaceQualityFRVT-IMAVIS-2010} & \attrNondl{}\attrDhc{} & Region density, Edge density, Eye distance. \textit{Continuation of \cite{Beveridge-FQA-PredictingFRVTPerformance-FG-2008}.} & FRVT 2006 \\\hline
2009 \cite{Sang-FQA-StandardGaborIDCT-ICB-2009} & \attrNondl{}\attrDhc{} & 2 factors: Asymmetry (Imaginary Gabor filters), Sharpness ((I)DCT). \textit{} & FERET, CMU-PIE \\\hline
2009 \cite{Zhang-FQA-AsymmetrySIFT-ISVC-2009} & \attrNondl{}\attrDhc{} & Symmetry (3 variations based on SIFT \cite{Lowe-SIFT-DistinctiveImageFeatures-IJCV-2004}). \textit{} & CAS-PEAL \\\hline
2008 \cite{Beveridge-FQA-PredictingFRVTPerformance-FG-2008} & \attrNondl{}\attrDhc{} & Edge density, Eye distance. \textit{} & FRVT 2006 \\\hline
2008 \cite{Rua-FQAwithFR-VideoFrameSelectionAndScoreNormalization-BioID-2008} & \attrNondl{}\attrDhc{}\attrV{} & Blur (Sobel \& Laplacian), Symmetry (Per-pixel). \textit{} & BANCA \\\hline
2008 \cite{Nasrollahi-FQA-InVideoSequences-BioID-2008} & \attrNondl{}\attrDuat{}\attrFe{}\attrV{} & Pose (Center of mass distance), Illumination, Blur, Resolution. \textit{} & FRI-CVL, HERMES project \\\hline
2007 \cite{Fourney-FQA-VideoFaceImageLogs-CRV-2007} & \attrNondl{}\attrDuat{}\attrFe{}\attrV{} & Pose (Eye positions via gradient image), Illumination range \& symmetry, Blur, Resolution, Skin content. \textit{} & Unspecified (7 videos) \\\hline
2007 \cite{Gao-FQA-StandardizationSampleQualityISO297945-ICB-2007} & \attrNondl{}\attrDuat{} & 6 factors: Lighting + Pose symmetry (LBP), Inter-eye distance, Illumination strength (Histogram), Contrast (Standard deviation), Blur (Gradient). \textit{} & Yale \\\hline
2007 \cite{Abdelmottaleb-FQA-BlurLightPoseExpression-CIM-2007} & \attrNondl{}\attrDfrt{}\attrFt{} & 4 measures: Blur (Frequency kurtosis), Illumination (Weighted sum), Pose (Yaw), Expression (GMM). \textit{} & FERET, WVU, Cohn-Kanade \\\hline
2006 \cite{Hsu-FQA-QualityAssessmentISO197945-BCC-2006} & \attrNondl{}\attrDhgt{}\attrFt{} & 27 factors listed, but few metric details; classification-error-based QS normalization; $3\times$ QS fusion, \ia{} ANN-based. \textit{} & In-house (Passport database), FRGC \\\hline
2006 \cite{Kryszczuk-FQA-ScoreAndSignalLevelGMM-EUSIPCO-2006} & \attrNondl{}\attrDuat{} & Same as \cite{Kryszczuk-FQA-OnFaceImageQualityMeasures-MMUA-2006}, plus another score-level measure. \textit{Continuation of \cite{Kryszczuk-FQA-OnFaceImageQualityMeasures-MMUA-2006}.} & BANCA \\\hline
2006 \cite{Kryszczuk-FQA-OnFaceImageQualityMeasures-MMUA-2006} & \attrNondl{}\attrDuat{} & Average face image correlation, Blur, Classification score sum of log-likelihoods. \textit{} & BANCA \\\hline
2005 \cite{Subasic-FQA-ValidationICAO-ISPA-2005} & \attrNondl{}\attrDhc{}\attrFc{} & 17 factors: Image resolution\slash AR, Blur, Illumination, Color balance, Background uniformity\slash tone, Shadows, Hot spots, Eyes tilt\slash position\slash red\slash looking away, Head width\slash height\slash rotation. \textit{} & Unspecified (189 images) \\\hline
2004 \cite{Yang-FQA-PoseVideoFrame-ICPR-2004} & \attrNondl{}\attrDuat{}\attrV{} & Pose (Haar features learned via SquareLev\@.R). \textit{} & Unspecified (300 faces) \\\hline
2004 \cite{Luo-FQA-TrainingbasedNoreferenceIQAA-ICIP-2004} & \attrNondl{}\attrDhgt{}\attrFt{} & RBF-ANN on: Brightness, Spectrum $\times 7$, Noise $\times 2$. \textit{} & Unspecified (850 images) \\\hline

\end{tabular}

\end{table*}

The factor-specific approach commonalities can be described by the factor subcategories depicted in \autoref{fig:taxonomy}:
\begin{itemize}
\commonalityPart{Size}
Testing the
inter-eye distance
\cite{Hsu-FQA-QualityAssessmentISO197945-BCC-2006}\allowbreak{}%
\cite{Gao-FQA-StandardizationSampleQualityISO297945-ICB-2007}\allowbreak{}%
\cite{Beveridge-FQA-PredictingFRVTPerformance-FG-2008}\cite{Beveridge-FQA-QuoVadisFaceQualityFRVT-IMAVIS-2010}\allowbreak{}%
\cite{Ferrara-FQA-BioLabICAO-TIFS-2012}\allowbreak{}%
\cite{Phillips-FQA-ExistenceOfFaceQuality-BTAS-2013}\allowbreak{}%
\cite{Henniger-FQA-HandcraftedFeatures-BIOSIG-2020}
or the image resolution
\cite{Subasic-FQA-ValidationICAO-ISPA-2005}\allowbreak{}%
\cite{Fourney-FQA-VideoFaceImageLogs-CRV-2007}\allowbreak{}%
\cite{Nasrollahi-FQA-InVideoSequences-BioID-2008}\allowbreak{}%
\cite{Nasrollahi-FQA-LowResolutionVideoSequence-TCSVT-2011}\allowbreak{}%
\cite{Nikitin-FQA-InVideo-GraphiCon-2014}
against some threshold is a comparatively simple approach to FIQA.
It is present in various mostly older works alongside other FIQA methods.
The referenced image resolution approaches mostly considered images cropped to the face and focused on video-frame selection.
Besides the face detection step, using the image resolution is trivial,
while inter-eye distance requires eye landmark detection.
\commonalityPart{Illumination}
Many of the surveyed works included mostly simple illumination measures, comprising
the brightness moments (mean, variance, skewness, or kurtosis)
\cite{Luo-FQA-TrainingbasedNoreferenceIQAA-ICIP-2004}%
\cite{Abdelmottaleb-FQA-BlurLightPoseExpression-CIM-2007}%
\cite{Nasrollahi-FQA-InVideoSequences-BioID-2008}\cite{Nasrollahi-FQA-LowResolutionVideoSequence-TCSVT-2011}%
\cite{Abaza-FQA-QualityMetricsPractical-ICPR-2012}\cite{Abaza-FQA-PhotometricIQA-IET-2014}%
\cite{Phillips-FQA-ExistenceOfFaceQuality-BTAS-2013}%
\cite{Wang-FQA-SubjectiveRandomForestHybrid-ICCC-2017}%
\cite{Khodabakhsh-FQA-SubjectiveVsObjectiveISO297945Quality-ICBEA-2019}%
\cite{Henniger-FQA-HandcraftedFeatures-BIOSIG-2020},
contrast measures
\cite{Gao-FQA-StandardizationSampleQualityISO297945-ICB-2007}%
\cite{Abaza-FQA-QualityMetricsPractical-ICPR-2012}\cite{Abaza-FQA-PhotometricIQA-IET-2014}%
\cite{Phillips-FQA-ExistenceOfFaceQuality-BTAS-2013}%
\cite{Khodabakhsh-FQA-SubjectiveVsObjectiveISO297945Quality-ICBEA-2019}%
\cite{Henniger-FQA-HandcraftedFeatures-BIOSIG-2020},
dynamic range measures
\cite{Hsu-FQA-QualityAssessmentISO197945-BCC-2006}%
\cite{Fourney-FQA-VideoFaceImageLogs-CRV-2007}%
\cite{Wang-FQA-SubjectiveRandomForestHybrid-ICCC-2017},
entropy measures
\cite{Kim-FQA-CascadedVideoFrame-ISM-2014}\cite{Kim-FQA-FaceImageAssessment-ICIP-2015}%
\cite{Damer-FRwithFQA-PersonalizedFaceReferenceVideo-FFER-2015},
or uniformity measures
\cite{Hsu-FQA-QualityAssessmentISO197945-BCC-2006}%
\cite{Rizorodriguez-FQA-IlluminationQualityMeasure-ICPR-2010}%
\cite{Demarsico-FQA-LandmarkPoseLightSymmetry-MiFor-2011}%
\cite{Beveridge-FQA-LightingAndFocus-CVPRW-2010}%
\cite{Zhang-FQA-SubjectiveIlluminationResNet50-ICONIP-2017}.
Note that ``illumination'' is of course also directly or indirectly measured by FIQAAs categorized under other parts of the taxonomy,
which are consequently not listed here to avoid many duplicate listings of the same FIQAAs.
\commonalityPart{Blur}
Blur measures, or conversely sharpness measures, are also known as (de)focus measures.
The measures can be subdivided into edge analysis approaches
\cite{Kryszczuk-FQA-OnFaceImageQualityMeasures-MMUA-2006}\allowbreak{}%
\cite{Kryszczuk-FQA-ScoreAndSignalLevelGMM-EUSIPCO-2006}\allowbreak{}%
\cite{Hsu-FQA-QualityAssessmentISO197945-BCC-2006}\allowbreak{}%
\cite{Gao-FQA-StandardizationSampleQualityISO297945-ICB-2007}\allowbreak{}%
\cite{Rua-FQAwithFR-VideoFrameSelectionAndScoreNormalization-BioID-2008}\allowbreak{}%
\cite{Beveridge-FQA-PredictingFRVTPerformance-FG-2008}\allowbreak{}%
\cite{Beveridge-FQA-QuoVadisFaceQualityFRVT-IMAVIS-2010}\allowbreak{}%
\cite{Beveridge-FQA-LightingAndFocus-CVPRW-2010}\allowbreak{}%
\cite{Hua-FQA-BlurMTF-ICB-2012}\allowbreak{}%
\cite{Ferrara-FQA-BioLabICAO-TIFS-2012}\allowbreak{}%
\cite{Abaza-FQA-QualityMetricsPractical-ICPR-2012}\allowbreak{}%
\cite{Abaza-FQA-PhotometricIQA-IET-2014}\allowbreak{}%
\cite{Phillips-FQA-ExistenceOfFaceQuality-BTAS-2013}\allowbreak{}%
\cite{Nikitin-FQA-InVideo-GraphiCon-2014}\allowbreak{}%
\cite{Wasnik-FQA-SmartphoneISO297945-IWBF-2017}\allowbreak{}%
\cite{Khodabakhsh-FQA-SubjectiveVsObjectiveISO297945Quality-ICBEA-2019}\allowbreak{}%
\cite{Henniger-FQA-HandcraftedFeatures-BIOSIG-2020},
frequency analysis approaches
\cite{Abdelmottaleb-FQA-BlurLightPoseExpression-CIM-2007}\allowbreak{}%
\cite{Fourney-FQA-VideoFaceImageLogs-CRV-2007}\allowbreak{}%
\cite{Sang-FQA-StandardGaborIDCT-ICB-2009}\allowbreak{}%
\cite{Hua-FQA-BlurMTF-ICB-2012}\allowbreak{}%
\cite{Kim-FQA-CascadedVideoFrame-ISM-2014}\allowbreak{}%
\cite{Kim-FQA-FaceImageAssessment-ICIP-2015},
and low-pass filter approaches
\cite{Nasrollahi-FQA-InVideoSequences-BioID-2008}\allowbreak{}%
\cite{Nasrollahi-FQA-LowResolutionVideoSequence-TCSVT-2011}\allowbreak{}%
\cite{Sang-FQA-StandardGaborIDCT-ICB-2009}\allowbreak{}%
\cite{Hua-FQA-BlurMTF-ICB-2012}.
Edge analysis involves image gradient computation,
frequency analysis inspects the image transformed into the frequency domain,
and low-pass filter approaches compare an artificially blurred version of the image with the original.
Besides these more common subcategories,
there were some comparatively opaque (\ie{} less easily explainable) deep learning approach among the FIQA literature that measured blur
\cite{Yu-FQA-LightCNNwithMFM-PRLE-2018}%
\cite{Lijun-FQA-MultibranchCNN-ICCT-2019}.
\commonalityPart{Symmetry}
Holistic symmetry measures compare the entire left and right half of the face.
The halves are defined either as fixed left/right splits of the whole input image
\cite{Rua-FQAwithFR-VideoFrameSelectionAndScoreNormalization-BioID-2008}%
\cite{Sang-FQA-StandardGaborIDCT-ICB-2009}%
\cite{Phillips-FQA-ExistenceOfFaceQuality-BTAS-2013}%
\cite{Wasnik-FQA-SmartphoneISO297945-IWBF-2017}%
\cite{Khodabakhsh-FQA-SubjectiveVsObjectiveISO297945Quality-ICBEA-2019},
or are fitted to the face within the image
\cite{Gao-FQA-StandardizationSampleQualityISO297945-ICB-2007}%
\cite{Fourney-FQA-VideoFaceImageLogs-CRV-2007}%
\cite{Wasnik-FQA-SmartphoneISO297945-IWBF-2017}%
\cite{Khodabakhsh-FQA-SubjectiveVsObjectiveISO297945Quality-ICBEA-2019}%
\cite{Henniger-FQA-HandcraftedFeatures-BIOSIG-2020}.
Fixed left/right halves assume that the face is frontal without rotation,
while fitted halves can account for some degree of head rotation.
Besides these holistic methods
there are localized symmetry measures which compare a number of paired local features within the left/right face halves,
either based on general key points \eg{} via SIFT (Scale Invariant Feature Transform) \cite{Zhang-FQA-AsymmetrySIFT-ISVC-2009},
or based on facial landmarks
\cite{Demarsico-FQA-LandmarkPoseLightSymmetry-MiFor-2011}%
\cite{Nikitin-FQA-InVideo-GraphiCon-2014}.
Of the surveyed methods,
only the localized landmark-based measures inherently avoided the inclusion of image background information,
although any of the methods could be extended to exclusively consider the facial area.
\commonalityPart{Pose}
Most pose FIQAAs were based on facial landmarks
\cite{Subasic-FQA-ValidationICAO-ISPA-2005}%
\cite{Abdelmottaleb-FQA-BlurLightPoseExpression-CIM-2007}%
\cite{Fourney-FQA-VideoFaceImageLogs-CRV-2007}%
\cite{Demarsico-FQA-LandmarkPoseLightSymmetry-MiFor-2011}%
\cite{Ferrara-FQA-BioLabICAO-TIFS-2012}%
\cite{Raghavendra-FQA-ABCVideoPoseGLCM-ICPR-2014}%
\cite{Wang-FQA-SubjectiveRandomForestHybrid-ICCC-2017}.
Others operated in a holistic manner
by using appearance templates \cite{Murphy-HeadPoseEstimationSurvey-TPAMI-2009} to estimate pose angle ranges
\cite{Yang-FQA-PoseVideoFrame-ICPR-2004}%
\cite{Nasrollahi-FQA-LowResolutionVideoSequence-TCSVT-2011},
by assessing the frontal face reconstruction error
\cite{Kim-FQA-CascadedVideoFrame-ISM-2014}\cite{Kim-FQA-FaceImageAssessment-ICIP-2015},
by assessing the pose without angles in terms of the pixels' center of mass deviation
\cite{Nasrollahi-FQA-InVideoSequences-BioID-2008},
or via comparatively opaque machine learning approaches that assessed whether the pose is frontal or not,
either with scalar \cite{Lijun-FQA-MultibranchCNN-ICCT-2019}
or binary \cite{Rose-FQA-FacialAttributesDeepLearning-ASONAM-2019}\cite{Rose-FQA-FacialAttributes-Springer-2020} output.
Among the methods that did correspond to specific pitch/yaw/roll angles (see \autoref{fig:face-pose-angle-pitch-yaw-roll})
most did consider the yaw angle in addition to either the roll or pitch angle,
while the rest considered either only the yaw angle or all three angles
\cite{Subasic-FQA-ValidationICAO-ISPA-2005}%
\cite{Abdelmottaleb-FQA-BlurLightPoseExpression-CIM-2007}%
\cite{Demarsico-FQA-LandmarkPoseLightSymmetry-MiFor-2011}.
The one landmark-based method that did not correspond to any specific angle \cite{Fourney-FQA-VideoFaceImageLogs-CRV-2007}
computed the deviation of a landmark-derived point from the horizontal face center,
which is closer to the holistic center of mass deviation approach of \cite{Nasrollahi-FQA-InVideoSequences-BioID-2008}.
\commonalityPart{Other}
There are some comparatively rare factor-specific FIQA approaches in the literature which were collected under the ``Other'' taxonomy category,
namely
binary attributes such as with/without glasses in \cite{Rose-FQA-FacialAttributesDeepLearning-ASONAM-2019}\cite{Rose-FQA-FacialAttributes-Springer-2020},
noise measures in \cite{Luo-FQA-TrainingbasedNoreferenceIQAA-ICIP-2004}\cite{Yu-FQA-LightCNNwithMFM-PRLE-2018},
skin tone measures in
\cite{Fourney-FQA-VideoFaceImageLogs-CRV-2007}%
\cite{Hsu-FQA-QualityAssessmentISO197945-BCC-2006}%
\cite{Subasic-FQA-ValidationICAO-ISPA-2005},
deep learning ``alignment'' and ``occlusion'' measures based on human ground truth in \cite{Lijun-FQA-MultibranchCNN-ICCT-2019},
and miscellaneous standard requirement check methods such as ink mark \& crease detection in
\cite{Ferrara-FQA-BioLabICAO-TIFS-2012}%
\cite{Hsu-FQA-QualityAssessmentISO197945-BCC-2006}%
\cite{Subasic-FQA-ValidationICAO-ISPA-2005}.
\end{itemize}

\subsection{Factor-specific - Literature introductions}
\label{sec:fiqaa-literature-factor}

\markAuthor{Luo} \cite{Luo-FQA-TrainingbasedNoreferenceIQAA-ICIP-2004}
considered general IQA related to brightness, blur, and noise in the context of face images.
Ten features were extracted from a grayscale image
and passed to a RBF (Radial Basis Function) ANN (Artificial Neural Network) to produce the final quality score.
As an alternative to the ANN, a GMM (Gaussian Mixture Model) was used as well,
but reportedly resulted in worse performance.
The IQAA was trained with and compared against the quality estimates of a single human on an unspecified dataset.
The 10 features consisted of
1 measure for average pixel brightness,
7 values derived from the sub-bands of two-level wavelet decomposition,
and 2 different noise measures
(one based on a square window with minimum grayscale pixel value standard deviation,
and one combining the standard deviation of square windows in binarized versions of the high-frequency sub-bands).

The approach of \markAuthor{Yang \etal{}} \cite{Yang-FQA-PoseVideoFrame-ICPR-2004} estimated only
the left-right\slash up-down pose angle, without producing any kind of normalized QS other than the binary decision between frontal and non-frontal pose;
faces being declared ``frontal'' when both pose angles have absolute values not higher than 10\textdegree{}.
While pure pose estimation literature is outside the scope of this survey,
this paper demonstrated that pose estimation can be used in isolation for FIQA.

\markAuthor{Subasic \etal{}} \cite{Subasic-FQA-ValidationICAO-ISPA-2005} used 17 FIQAAs based on ICAO Doc 9303 \cite{ICAO-9303-2015} requirements.
This includes measures that are less common among the literature, such as background uniformity and color balance.
The 17 FIQAAs were integrated as part of a combined FIQA system,
reusing background/skin/eye-segmentation for multiple measures,
and hierarchically executed, \ie{} resolution and sharpness are examined first.
The combined FIQAA was used to determine whether an input image is ICAO-compliant or not, and the evaluation tested this binary prediction against 189 correspondingly labelled images of an unnamed database, correctly classifying 88\%.
Tolerance ranges were established based on a small subset of images where no quantitative ICAO requirements were available, and some existing ICAO tolerance ranges were relaxed.

In the approach of \markAuthor{Kryszczuk and Drygajlo} \cite{Kryszczuk-FQA-OnFaceImageQualityMeasures-MMUA-2006},
2 image-based (``signal-level'') and 1 classification-score-based (``score-level'') measure were used,
and all 3 were
combined by means of 2 GMMs with 12 Gaussian components each for binary assessments
regarding ``correct'' and ``erroneous'' FR classifier decisions.
The authors also added another score-level measure to the approach in \cite{Kryszczuk-FQA-ScoreAndSignalLevelGMM-EUSIPCO-2006}.
But the inclusion of measures based on FR classification scores means that the combined method can only be used after a FR comparison has taken place, so this component would have to be excluded to allow isolated single-image FIQA using the remaining 2 image-based methods.
Of these, one measured sharpness as the mean of horizontal\slash vertical pixel intensity differences (corresponding to high-frequency features),
and the other computed Pearson's cross-correlation coefficient between the face image and an average face image (corresponding to low-frequency features).
The average face image was formed from the average of the first 8 PCA (Principal Component Analysis) eigenfaces for a given training image set.

\markAuthor{Hsu \etal{}} \cite{Hsu-FQA-QualityAssessmentISO197945-BCC-2006} used 27 FIQA factors,
which mostly relate to ISO/IEC 19794-5:2005 \cite{ISO-IEC-19794-5-2005} requirements.
While only very brief descriptions of the underlying FIQA approaches were provided,
the work proposed quality score normalization and fusion with more details.
The normalization per metric was based on the classification error against binary human quality labels (``good''/``poor'').
Raw quality metric values were mapped to $[0, 1]$ via 5 raw value thresholds, interpolated via sigmoid functions.
The 5 raw threshold values were taken from 5 specific points of the false-accept/reject classification error curves, and corresponded to the quality scores ${0,0.4,0.5,0.7,1.0}$.
For FIQAA fusion, 3 models were trained, and the evaluation showed that a non-linear neural network obtained the best results in terms of correlation with FR performance.

\markAuthor{Abdel-Mottaleb and Mahoor} \cite{Abdelmottaleb-FQA-BlurLightPoseExpression-CIM-2007} proposed FIQAAs to assess blur, lighting, pose, and facial expression.
Blur was measured as the kurtosis in the frequency domain.
The lighting QS was formed by a weighted sum of the mean intensity values for 16 weight-defined regions,
to focus more on the center of the image.
Pose was estimated as the yaw angle (see \autoref{fig:face-pose-angle-pitch-yaw-roll}),
derived by comparing the amount of skin tone pixels between the left\slash right-side triangle,
which in turn were defined by the 3 center points of the eyes and the mouth.
Fisher Discriminant Analysis (FDA) was employed to differentiate skin pixels from other regions.
To assess whether the expression is good or bad in terms of quality,
a GMM was trained based on the correct\slash incorrect decisions of an FR algorithm for a labeled facial expression dataset.

\markAuthor{Gao \etal{}} \cite{Gao-FQA-StandardizationSampleQualityISO297945-ICB-2007}
proposed FIQAAs for asymmetry, inter-eye distance, illumination strength, contrast, and blur.
Lighting/pose asymmetry was computed as the sum of the rectilinear distances between the histogram pairs for multiple LBP (Local Binary Pattern) features at designated locations in the face image halves.
Illumination strength was proposed to be computed
as the difference between a histogram for the input image and a fixed standard illumination histogram,
contrast as the pixel value standard deviation,
and sharpness as the gradient magnitude sum.
The asymmetry metric was tested in terms of classification accuracy with a small labeled dataset.
The methods have been incorporated into ISO/IEC TR 29794-5:2010 \cite{ISO-IEC-29794-5-TR-FaceQuality-100312} (but the work is only cited directly for the lighting/pose symmetry part).

\markAuthor{Fourney and Laganiere} \cite{Fourney-FQA-VideoFaceImageLogs-CRV-2007} defined a pose QS
as linearly degrading from 0\textdegree{} to 45\textdegree{}, anything above 45\textdegree{} resulting in a score of 0,
a clear contrast to the binary decision in \cite{Yang-FQA-PoseVideoFrame-ICPR-2004}.
The pose estimation in \cite{Fourney-FQA-VideoFaceImageLogs-CRV-2007} also worked in a different manner, namely by locating the eye positions in a gradient image, which was noted to be ineffective for faces with glasses or non-upright orientation.
Based on this pose estimation data, illumination symmetry FIQA was also conducted by comparing normalized histograms of the left\slash right side of the face,
which was done in addition to an assessment of the overall utilization of the available (\eg{} 8-bit grayscale) illumination range within the face image.
The remaining factors in \cite{Fourney-FQA-VideoFaceImageLogs-CRV-2007} were unrelated to the pose estimation:
A normalized blur\slash sharpness QS was derived from the frequency domain;
the face image resolution\slash pixel count was transformed into a normalized QS,
with anything at or above $60 \times 60$ pixels
corresponding to the maximum (a QS of 1);
and a ``skin content'' measure detected whether human skin appears to be present in the image, which was done by determining the percentage of pixels with a hue of $[-30^{\circ},+30^{\circ}]$ and saturation of $[5\%,95\%]$.
The final combined QS of the 6 factors consisted of the number of satisfied per-factor thresholds, plus a weighted sum of the factor scores to break ties between video frames.

For the works \cite{Nasrollahi-FQA-InVideoSequences-BioID-2008} and \cite{Nasrollahi-FQA-LowResolutionVideoSequence-TCSVT-2011} from \markAuthor{Nasrollahi and Moeslund}, 
it is important to note that both
derived a QS for each of their factors, except resolution, relative to
minimum or maximum values for a sequence of face images
- so the described approaches are not directly usable for single-image FIQA.
We can remedy this obstacle using simple tricks, for example by choosing constant minima\slash maxima,
hence why these works are still included here.
The first of the two papers, \cite{Nasrollahi-FQA-InVideoSequences-BioID-2008}, \ia{} cited \cite{Fourney-FQA-VideoFaceImageLogs-CRV-2007} and directly adapted the face image resolution factor,
but presented different approaches to measure the other shared factors:
The FIQAA started with information gathered as part of the face detection stage,
which determined potential facial regions per-pixel by skin tone,
applying a cascading classifier thereon to obtain the face image(s) for further steps.
Skin tone pixel count percentages were however not used directly for a QS,
in contrast to \cite{Fourney-FQA-VideoFaceImageLogs-CRV-2007} and \cite{Ferrara-FQA-BioLabICAO-TIFS-2012}.
Instead,
\ia{} the facial center of mass was derived from this per-pixel segmentation.
The paper noted that estimating the pose cannot be reliable when using facial features (such as the eyes in \cite{Fourney-FQA-VideoFaceImageLogs-CRV-2007}),
since they may not be visible for sufficiently large angles of rotation,
or can be occluded by \eg{} glasses.
Therefore the difference between the facial center of mass and the center of the face image was used,
a method diverging from previously mentioned approaches that estimated specific angles.
Illumination was measured as the average pixel brightness over the face image (against the maximum value for a face image sequence; but here a simple normalization could be applied instead for single-image FIQA).
Sharpness\slash blur was assessed using the approach presented in \cite{weberQualityMeasuresFace2006}, \ie{} by first subtracting a low-pass ($3 \times 3$ mean filtered) version of the face image from the original per-pixel, then averaging the absolute values of all these pixel differences.
The FIQAA in \cite{Nasrollahi-FQA-LowResolutionVideoSequence-TCSVT-2011} can be seen as a continuation of \cite{Nasrollahi-FQA-InVideoSequences-BioID-2008},
with the sharpness, brightness, and resolution measures being almost identical.
Brightness had now been more clearly defined as the Y component of the YCbCr color space
and the resolution QS bound was removed (\ie{} it became completely relative to an image sequence).
The pose estimation was changed,
stating that the prior center of mass approach in \cite{Nasrollahi-FQA-InVideoSequences-BioID-2008} tended to be sensitive to environmental conditions.
The new approach estimated actual angles and is adapted from \cite{gourierHeadPoseEstimation2007},
using one auto-associative memory (an ANN without hidden layers) per detectable pose.

\markAuthor{Rúa \etal{}} \cite{Rua-FQAwithFR-VideoFrameSelectionAndScoreNormalization-BioID-2008} proposed three FIQA methods in the context of face video frame selection.
One method measured symmetry by comparing the image against a horizontally flipped version of itself,
calculating the per-pixel difference,
meaning that this measure assumed a centered frontal pose.
The other two FIQA methods assessed blur by computing the average value for either the Sobel or the Laplace operator over the entire input image.

\markAuthor{Beveridge \etal{}} examined the impact of a number of factors on FR verification performance in \cite{Beveridge-FQA-PredictingFRVTPerformance-FG-2008} and \cite{Beveridge-FQA-QuoVadisFaceQualityFRVT-IMAVIS-2010} using GLMMs (Generalized Linear Mixed Models).
Taking the examined preexisting labels such as age or gender out of consideration,
three described measurements were considered for automatic image-only quality assessment,
one of which is the image resolution\slash eye distance.
Two more complex measurements remain,
with \cite{Beveridge-FQA-PredictingFRVTPerformance-FG-2008} introducing an edge density metric consisting of the averaged Sobel filter pixel magnitude,
and \cite{Beveridge-FQA-QuoVadisFaceQualityFRVT-IMAVIS-2010} adding a region density metric that segments the face and counts the distinct regions. 
Both of these metrics were applied on grayscale images,
with the face area being masked by an ellipse
to reduce the metrics' sensitivity to environmental factors in the rest of the image.
The authors continued in \cite{Beveridge-FQA-LightingAndFocus-CVPRW-2010}
by comparing their edge density metric to two newly introduced FIQAAs.
One was the Strong Edge and Motion Compensated focus measure (SEMC focus),
a successor to the edge density metric that was computed based on the strongest edges in the face region (instead of all),
which was intended to correlate more clearly to focus\slash blur in images (instead of also being affected by other factors such as illumination).
The second new FIQAA estimated to which degree a face is lit from the front (positive number output) or the side (negative number).
Experiments in \cite{Beveridge-FQA-LightingAndFocus-CVPRW-2010} used GLMMs and FRVT 2006 test data\slash FR algorithms similar to \cite{Beveridge-FQA-PredictingFRVTPerformance-FG-2008} and \cite{Beveridge-FQA-QuoVadisFaceQualityFRVT-IMAVIS-2010},
and found that the illumination measure subsumed both the edge density and the SEMC focus measure regarding FR performance prediction.
These measures were studied further in \cite{Phillips-FQA-ExistenceOfFaceQuality-BTAS-2013}, as described below.

\markAuthor{Zhang and Wang} \cite{Zhang-FQA-AsymmetrySIFT-ISVC-2009} proposed three symmetry measure variations based on SIFT \cite{Lowe-SIFT-DistinctiveImageFeatures-IJCV-2004} (Scale Invariant Feature Transform).
The first variation counted the number of SIFT points in the left and right half of the image,
and divided the minimum of the two numbers by their maximum to obtain the QS.
Using the fixed left\slash right image halves entails that this measure is intended for frontal face images.
The second QS variation was formed by the amount of SIFT points that have
a mated point in the other half based on their location.
And the third variation further added a Euclidean distance comparison of the SIFT feature vectors to
define corresponding points,
using a horizontally flipped version of the image to establish
target points with directly comparable SIFT features.
As part of the evaluation, the first and simplest variant was shown to have the highest correlation with Eigenface- and LBP-based FR comparison scores.

\markAuthor{Sang \etal{}} \cite{Sang-FQA-StandardGaborIDCT-ICB-2009}
proposed a Gabor-filter-based asymmetry FIQAA to assess the illumination\slash pose,
and a sharpness FIQAA using DCT+IDCT (Discrete Cosine Transform + Inverse DCT).
The asymmetry FIQAA used the left/right halves of the input image,
expecting an aligned frontal image.
It was computed as the sum of the absolute difference between the left/right pixels
for multiple filtered versions of the image halves,
mirroring the right half for the comparison.
The imaginary parts of Gabor filters were used with 5 orientations,
also mirrored on the right half.
To assess sharpness,
DCT followed by IDCT was applied to the input image
to obtain a reconstructed version without high-frequency information,
and the difference between both image variants was used to establish the sharpness value.
The asymmetry FIQAA was examined via score plots for images with different lighting and pose conditions (of the same subjects),
and the sharpness FIQAA similarly for either unmodified or synthetically blurred images,
demonstrating classification potential for both.
The asymmetry FIQAA approach from \cite{Gao-FQA-StandardizationSampleQualityISO297945-ICB-2007} was included in the tests and produced similar output compared to the proposed FIQAA.

\markAuthor{Rizo-Rodriguez \etal{}} \cite{Rizorodriguez-FQA-IlluminationQualityMeasure-ICPR-2010} presented a frontal illumination assessment method.
First, a triangular mesh was fitted to the face in the input image.
Then the mean luminance was computed for each of the triangle regions,
forming a histogram of mean luminance values per face,
which was observed to approximate a normal distribution in face images with homogeneous frontal illumination.
This was used to derive a binary QS using an experimentally obtained threshold.
To additionally account for differences in importance between the regions,
a three layer perceptron was trained for important regions only
- \ie{} input neurons for 24 triangles in the vicinity of the nose.
A binary QS was obtained from this ANN as well,
and both of the QS decisions were optionally combined.

\markAuthor{De Marsico \etal{}} \cite{Demarsico-FQA-LandmarkPoseLightSymmetry-MiFor-2011} proposed landmark-based measures for pose, illumination, and symmetry.
For pose, the yaw\slash pitch\slash roll angles were assessed using landmarks for the eye centers, the tip and root of the nose, and the chin.
A weighted sum of the three $[0,1]$ angle QSs formed the pose QS,
whereby the weights for yaw ($0.6$), pitch ($0.3$), and roll ($0.1$) were derived experimentally.
Illumination was measured by applying a sigmoid function to the variance of the mass centers for 8 gray level histograms,
which were computed for areas around 8 landmarks (3 on the nasal ridge, 2 on each cheek, 1 on the chin).
Symmetry was measured by comparing the grayscale values of point pairs
sampled along 8 lines defined by landmark pairs on each side of the face.
All three measures provided $[0,1]$ scalar QS results.
They were not fused,
but it was noted that the symmetry measure inherently takes both pose and illumination into account.
The evaluations demonstrated \ia{} that the FR performance improvement capabilities of the measures differed depending on the used FR algorithm.

\markAuthor{Liao \etal{}} \cite{Liao-FQA-GaborCascadeSVM-ICBEB-2012} trained an SVM (Support Vector Machine) cascade to predict subjective QS labels using Gabor filter magnitude values as features.
The SVM cascade had four stages, each being a binary classifier, so that the approach predicted integer QS levels from 1 to 5 (\eg{} the first SVM decides whether the QS is 1, or whether it might be higher).
Two of these SVM cascades were used for two different image crop sizes, and their output QSs were fused by taking the mean.
Training and evaluation used partitions of a dataset with 22,720 grayscale images,
all with subjective ground truth QS labels (1 to 5; 1 being the best quality).
The evaluation showed that the fusion approach provided the best predictive performance overall.

Multiple IQA methods were examined for FIQA by \markAuthor{Abaza \etal{}} in \cite{Abaza-FQA-QualityMetricsPractical-ICPR-2012},
and later \cite{Abaza-FQA-PhotometricIQA-IET-2014},
\ia{} incorporating synthetic image degradations regarding contrast, brightness, and blurriness for the evaluations.
Of the 12 tested individual measures in \cite{Abaza-FQA-QualityMetricsPractical-ICPR-2012},
7 were retained to represent 5 input factors for a combined single-image FIQAA,
using Gaussian models for normalization and the geometric mean for fusion.
Contrast was measured as the RMS (Root Mean Square) of image intensity,
brightness as the average HSB (\ie{} HSV, Hue Saturation Value\slash Brightness) color space brightness
(computable as the maximum of the normalized red\slash green\slash blue channel value per pixel \cite{Abaza-FQA-PhotometricIQA-IET-2014}),
focus as the mean of the image gradient's $L_1$-norm and the Laplacian energy \cite{yapImageFocusMeasure2004},
sharpness as the mean of the two average gradient measures \cite{Kryszczuk-FQA-ScoreAndSignalLevelGMM-EUSIPCO-2006} and \cite{Gao-FQA-StandardizationSampleQualityISO297945-ICB-2007},
and illumination using the weighted sum technique proposed by \cite{Abdelmottaleb-FQA-BlurLightPoseExpression-CIM-2007}.
The 5 measures that were not used for the combined FIQAA comprise
the Michelson contrast measure \cite{bexSpatialFrequencyPhase2002},
the brightness measure from \cite{bezryadinBrightnessCalculationDigital2007},
the Tenengrad sharpness measure plus an adaptive variant from \cite{yaoImprovingLongRange2008},
and the luminance distortion \cite{zhouwangUniversalImageQuality2002} measure previously seen in \cite{Sellahewa-FQA-LuminanceDistortion-TIM-2010} (but without the face average as the ``reference'').
Note that according to both \cite{Abaza-FQA-QualityMetricsPractical-ICPR-2012} and \cite{Abaza-FQA-PhotometricIQA-IET-2014} the selected brightness measurement was chosen due to its reduced computational workload in comparison to the other tested method (which achieved better predictive performance).
Continuing with \cite{Abaza-FQA-PhotometricIQA-IET-2014},
the same 5 factors based on the chosen 7 (of 12) measures were presented as in \cite{Abaza-FQA-QualityMetricsPractical-ICPR-2012},
but now an ANN was trained to combine the 5 factors without any prior normalization to produce a binary QS classification.
A single-layer ANN with six neurons was found to provide the best classification results
among 10 different ANNs with either 1 or 2 layers (and 4 to 20 neurons per layer),
logistic regression,
SVR (Support Vector Regression),
as well as 10 combination approaches formed from a normalization ($\times 2$, linear or Gaussian model) and a fusion ($\times 5$) part,
including the previous method from \cite{Abaza-FQA-QualityMetricsPractical-ICPR-2012}.  
However, the tested methods\slash ANNs' 5-factor input vector apparently was the per-element minimum of the vectors for both a probe and a gallery image,
so here the probe image was not used in isolation.

To measure blur in face images,
\markAuthor{Hua \etal{}} \cite{Hua-FQA-BlurMTF-ICB-2012} proposed using the Modulation Transfer Function (MTF),
and evaluated this approach together with various other blur related measures:
A measure based on the radial spatial frequencies of 2D DCT coefficients,
a Squared Gradient (SG in \autoref{tab:fiqaa-factor}) metric that consisted of the gradient image (edge) magnitudes,
and a Laplacian of Gaussian (LoG) method.
There also was an Edge Density (ED) measure,
which was formed by first subtracting the $3 \times 3$ mean filtered image from the original,
then taking the average of the result's absolute pixel values \cite{weberQualityMeasuresFace2006}.
This measure also occurred in \cite{Nasrollahi-FQA-InVideoSequences-BioID-2008},
but is not to be confused with the previously mentioned Sobel filter edge density from \cite{Beveridge-FQA-PredictingFRVTPerformance-FG-2008} and \cite{Beveridge-FQA-QuoVadisFaceQualityFRVT-IMAVIS-2010}.
The correlation of these measures (applied to a face image) to a ground truth MTF applied to an optical chart was assessed,
with the face image MTF showing the highest, and edge density the lowest average correlation,
the other mentioned measures having high correlation closer to the MTF result.

\markAuthor{Ferrara \etal{}} \cite{Ferrara-FQA-BioLabICAO-TIFS-2012} introduced the ``BioLab-ICAO'' framework for ISO/ICAO-compliance assessment,
comprising a database, a testing protocol, and a set of 30 FIQAAs for requirements derived from ISO/IEC 19794-5:2005 \cite{ISO-IEC-19794-5-2005} (plus corrigenda/amendments).
The 30 FIQA measures included factors that were less common in the surveyed literature, such as the detection of ink marks or creases.
The evaluation individually tested 23 of the 30 measures,
together with 2 unnamed COTS (Commercial Off-The-Shelf) FIQAAs,
mostly in terms of compliance prediction accuracy (range $[0,100]$) against ground truth labels.
Most of the proposed BioLab-ICAO methods either outperformed both COTS systems or lacked a testable COTS counterpart,
although the assessment of various requirements was still deemed to be difficult.
BioLab-ICAO methods were later used in the training data preparation for FaceQnet v0 \cite{Hernandezortega-FQA-FaceQnetV0-ICB-2019} and v1 \cite{Hernandezortega-FQA-FaceQnetV1-2020}.

\markAuthor{Phillips \etal{}} \cite{Phillips-FQA-ExistenceOfFaceQuality-BTAS-2013} examined 13 quality measures,
including the edge density metric from \cite{Beveridge-FQA-PredictingFRVTPerformance-FG-2008} and \cite{Beveridge-FQA-QuoVadisFaceQualityFRVT-IMAVIS-2010},
plus the SEMC focus measure from \cite{Beveridge-FQA-LightingAndFocus-CVPRW-2010}
(all four of these papers share authors).
There also was an ``illumination direction'' measure that might correspond to \cite{Beveridge-FQA-LightingAndFocus-CVPRW-2010} as well, but this was not clarified.
Similar to the two prior papers \cite{Beveridge-FQA-PredictingFRVTPerformance-FG-2008} and \cite{Beveridge-FQA-QuoVadisFaceQualityFRVT-IMAVIS-2010},
the 13 quality measures in \cite{Phillips-FQA-ExistenceOfFaceQuality-BTAS-2013} contained preexisting labels from EXIF (Exchangeable Image File) metadata, \eg{} exposure time,
leaving 9 measures that can clearly consist of FIQA approaches which use the actual image (pixel) data:
Edge density \cite{Beveridge-FQA-PredictingFRVTPerformance-FG-2008},
SEMC focus \cite{Beveridge-FQA-LightingAndFocus-CVPRW-2010},
illumination direction (possibly \cite{Beveridge-FQA-LightingAndFocus-CVPRW-2010}),
left-right side illumination histogram comparison,
eye distance,
face saturation (the number of face pixels holding the maximum intensity value),
pixel standard deviation,
mean ratio (mean pixel value of the face region compared to the entire image),
and pose (yaw angle, 0 being frontal).
The 13th quality measure was an SVM that
summarized the other 12 measures.
Pruning based on the 13 measures was compared against a Greedy Pruned Order (GPO) oracle that discarded images in an approximately optimal fashion to improve FR performance,
thus representing an upper bound for FR performance improvements enabled by some FIQAA.
Experimental results indicated a substantial gap between the oracle and the 13 quality measures,
with various measures such as the illumination direction additionally leading to worse
FMR (False Match Rate) results.
Another FIQAA using PCA followed by LDA (Linear Discriminant Analysis) was trained, but it was observed to generalize poorly to the test set.

In the single-image FIQAA of \markAuthor{Nikitin \etal{}} \cite{Nikitin-FQA-InVideo-GraphiCon-2014} the resolution and illumination measurement,
as well as the fusion to combine the factor-QSs,
did not differ much from what has been mentioned previously
(resolution QS relative to constants, illumination dynamic range usage QS, fusion via weighted sum).
However, here facial landmarks were detected to measure symmetry by comparing the left\slash right landmark-local gradient histograms,
and to measure sharpness via averaged Laplace operator values only within the landmark-defined facial area.

A two stage approach was proposed for - and evaluated with - an ABC (Automatic Border Control)
system by \markAuthor{Raghavendra \etal{}} \cite{Raghavendra-FQA-ABCVideoPoseGLCM-ICPR-2014},
with the first stage consisting of a yaw\slash roll angle pose estimation based on the eye and nose position.
The final QS was represented by three bins, poor\slash fair\slash good,
and if the pose was not detected as frontal, the overall FIQAA stopped, assigning the image to the poor QS bin.
If the pose was detected as frontal,
the second stage decided between the fair\slash good QS bin assignment.
It consisted of 12 GLCM (Gray Level Co-occurrence Matrix) features \cite{haralickTexturalFeaturesImage1973},
which were further processed by a GMM (Gaussian Mixture Model) trained on public non-ABC datasets,
and the output thereof was used to obtain the final binary QS bin decision via a threshold.

The approach of \markAuthor{Kim \etal{}} \cite{Kim-FQA-CascadedVideoFrame-ISM-2014}
began by employing (frontal) face reconstruction to assess pose\slash alignment quality as the difference between the original and the reconstructed face image;
then in stage two blur was measured as the kurtosis of the CDF (Cumulative Distribution Function) of the DFT (Discrete Fourier Transform) magnitude;
and the last stage assessed brightness by comparing the histogram for the face image against a given reference histogram,
whereby the latter simply was chosen to be the uniform histogram.
Each of these three stages ended by comparing the error value result against a predefined threshold,
aborting the overall FIQAA if the threshold was exceeded.
This cascaded approach in \cite{Kim-FQA-CascadedVideoFrame-ISM-2014} was primarily meant to reduce the computational complexity for video processing.
In the follow-up paper \cite{Kim-FQA-FaceImageAssessment-ICIP-2015} the same three measures were utilized, but without the cascaded approach.
Instead, the output of the three so called ``objective'' measures formed a QS vector.
An additional ``relative'' quality measurement was conducted to assess the dissimilarity of the input image (\eg{} from the test dataset) to the training dataset images.
This was done via a multivariate Gaussian distribution for a 6-dimensional vector,
consisting of the averaged red\slash green\slash blue color channel values,
and the three aforementioned ``objective'' measure values.
To finally predict a binary QS label,
an unspecified number of weak classifiers were learned via AdaBoost
to form a combined FIQAA,
the input thereof being a 9-dimensional vector made up of the 3-dimensional ``objective'' and 6-dimensional ``relative'' measure output.
Note that the ``relative'' measure is entirely optional,
but it did improve the quality assessment according to the evaluation in \cite{Kim-FQA-FaceImageAssessment-ICIP-2015}.
In these evaluations both variants of the proposed FIQAA appeared superior to the also tested RQS \cite{Chen-FQA-LearningToRank-SPL-2015},
which seemed to actually degrade FR performance.

The work \cite{Damer-FRwithFQA-PersonalizedFaceReferenceVideo-FFER-2015} by \markAuthor{Damer \etal{}} included three face frame selection methods,
of which two could be considered for single-image FIQA.
One method measured the entropy of the color channels,
higher entropy being preferred.
The other method calculated the confidence for a Viola-Jones \cite{Viola-RapidObjectDetection-CVPR-2001} face detector as the sub-image classifier detection count,
which can correspond \ia{} to pose and illumination.

\markAuthor{Zhang \etal{}} \cite{Zhang-FQA-SubjectiveIlluminationResNet50-ICONIP-2017} created FIIQD, a ``Face Image Illumination Quality Database'' with subjective illumination quality scores for $224,733$ images with 200 different illumination patterns (established patterns were transferred to images from various other databases, together with their associated ground truth QS labels).
Then a model based on ResNet50 \cite{heDeepResidualLearning2015} was trained with that data to estimate the illumination quality.
A strong correlation was shown between the predicted illumination QSs and the labels,
but the impact on FR performance was not evaluated.

\markAuthor{Wasnik \etal{}} \cite{Wasnik-FQA-SmartphoneISO297945-IWBF-2017}
examined FIQA in the context of smartphone-based FR,
evaluating 8 FIQAAs based on ISO/IEC TR 29794-5:2010 \cite{ISO-IEC-29794-5-TR-FaceQuality-100312} specifications,
and proposing a vertical edge density FIQAA for pose/lighting symmetry,
plus a combined random forest FIQAA.
The vertical edge density FIQAA computed the input image gradient,
only keeping the magnitude for (vertical edge) pixels in a certain gradient phase range,
and used the mean of all magnitude pixel values to form a scalar result.
The random forest FIQAA combined the 8 ISO metrics,
and a second variant replaced the ISO symmetry assessment part with the proposed vertical edge density FIQAA.
To train the random forest algorithm, a database was first separated into good and bad quality images using a COTS system (VeriLook 5.4 \cite{neurotechnology-face}) plus subsequent manual checks by three trained experts.
All 9 individual FIQAAs, the 2 random forest FIQAAs,
and the COTS FIQAA
were evaluated by computing ERCs
using a FR implementation from the same COTS suite \cite{neurotechnology-face}.
The COTS FIQAA and the random forest algorithm incorporating the vertical edge metric
provided the best results in terms of partial (20\%) ERC AUC.
The work by \markAuthor{Khodabakhsh \etal{}} \cite{Khodabakhsh-FQA-SubjectiveVsObjectiveISO297945Quality-ICBEA-2019}
can be considered as a continuation of \cite{Wasnik-FQA-SmartphoneISO297945-IWBF-2017} which examined the 8 ISO FIQAAs in comparison to subjective quality assessments made by 26 human participants for smartphone images.
It concluded \ia{} that the human FIQA highly correlated with FR performance, but not with the tested FIQAAs,
indicating that the tested FIQAAs show limitations.
Correlation between the metrics were also shown.

\markAuthor{Wang} \cite{Wang-FQA-SubjectiveRandomForestHybrid-ICCC-2017} presented a hybrid approach to estimate subjective QSs using features consisting of 7 factor-specific scores.
The factors comprised brightness, dynamic range, illuminance uniformity, sharpness, pose (yaw\slash pitch angles), as well as the landmark-based similarity to a ``typical'' face formed from the average of various training images.
A random forest regressor was trained using these factors to estimate subjective ground truth QSs from 1 to 5.
The single-image part of the evaluation compared the predictive performance of this approach against the cascaded SVM method of \cite{Liao-FQA-GaborCascadeSVM-ICBEB-2012},
with the results favoring the proposed approach for QSs 2 to 3.

\markAuthor{Yu \etal{}} \cite{Yu-FQA-LightCNNwithMFM-PRLE-2018} proposed using a CNN architecture with MFM \cite{wuLightCNNDeep2018} (Max-Feature-Map) and NIN \cite{linNetworkNetwork2014} (Network In Network) layers for FIQA.
Training used 16 classes:
One represented the original unmodified training images,
while the other 15 represented 5 types of synthetic degradation thereof,
with 3 configurations of increasing severity each.
These 5 degradation types comprised nearest-neighbor downscaling, Gaussian blur, AWGN (Additive White Gaussian Noise), salt-and-pepper noise, and Poisson noise.
This was sufficient to train a network to classify these degradations.
To also estimate a scalar QS,
a FR accuracy score was precomputed for each of the 16 classes,
and the sum of the multiplication of those scores with the 16 classification probabilities formed the combined QS.
The proposed CNN architecture was also used for the FR part (as a separately trained model), using the cosine distance as the similarity measure.
Three variants of the network were evaluated for FIQA:
One trained from scratch for FIQA,
one first trained for FR before training for FIQA,
and one that used ReLU instead of MFM layers.
The evaluation \ia{} compared the variants regarding their degradation classification performance,
showing superior accuracy for the two MFM variants in contrast to the ReLU architecture,
whereby the best overall results stemmed from the FR transfer learning variant.
Regarding the 5 degradation types,
the FR performance appeared to be predominantly affected by AWGN as well as salt-and-pepper noise,
while the other types were less impactful even for their more severe configurations.

\markAuthor{Rose and Bourlai} evaluated DL and non-DL methods to determine three binary facial attributes in \cite{Rose-FQA-FacialAttributes-Springer-2020} and \cite{Rose-FQA-FacialAttributesDeepLearning-ASONAM-2019} (which was a continuation of \cite{Rose-FQA-FacialAttributes-Springer-2020} despite the publication date order):
Whether the eyes are open or closed, whether there are glasses or not, and whether the face pose is mostly frontal or not.
The two DL methods in both papers consisted of AlexNet \cite{krizhevskyImageNetClassificationDeep2017} and GoogLeNet \cite{szegedyGoingDeeperConvolutions2014} (an incarnation of the Inception architecture),
pretrained on ImageNet \cite{dengImageNetLargeScaleHierarchical2009} data.
Their architectures were modified to classify 2 labels per attribute (\ie{} 6 classes).
And there were 23 non-DL models tested in \cite{Rose-FQA-FacialAttributes-Springer-2020},
including SVMs, K-Nearest Neighbors, Decision Trees, and Ensemble classifiers.
LBP and HOG features were evaluated for these non-DL methods, and HOG was found to consistently outperform LBP.
A score-level fusion of a SVM and either AlexNet or GoogLeNet led to the best results in \cite{Rose-FQA-FacialAttributes-Springer-2020}.
The evaluations in \cite{Rose-FQA-FacialAttributesDeepLearning-ASONAM-2019} employed a smartphone (iPhone 5S) dataset in addition to the non-smartphone data used in \cite{Rose-FQA-FacialAttributes-Springer-2020},
the latter of which was only used for training.
Of the non-DL methods, result values in \cite{Rose-FQA-FacialAttributesDeepLearning-ASONAM-2019} were only shown for the cubic kernel SVM approach,
because the other methods performed worse.
Whether the performance of the SVM or one of the two DL methods was better varied between the experiments of \cite{Rose-FQA-FacialAttributesDeepLearning-ASONAM-2019},
which proposed to use the SVM trained on a combination of all used datasets (score-level fusion of the SVM and one of the DL networks was not tested).

\markAuthor{Lijun \etal{}} \cite{Lijun-FQA-MultibranchCNN-ICCT-2019} proposed a multi-branch FIQA network, called MFQA, consisting of a feature extraction and a quality score part.
The former was a CNN to derive image features.
The latter fed these features into four fully connected branches for different quality properties,
and fused the output thereof into a final QS via another fully connected layer.
These four branches corresponded to scores for alignment, visibility (\ie{} occlusion), frontal pose, and clarity (\ie{} blur).
For training, 3,000 images were manually annotated with ground truth labels for the four factor scores and the overall QS.

\markAuthor{Henniger \etal{}} \cite{Henniger-FQA-HandcraftedFeatures-BIOSIG-2020}
examined 17 hand-crafted FIQA measures drawn from ISO/IEC TR 29794-5:2010 \cite{ISO-IEC-29794-5-TR-FaceQuality-100312}.
Of these, 7 measured symmetry of the left-right face image halves,
whereof 1 measure summed the normalized pixel luminance differences,
and the other 6 calculated the
cross-entropy,
Kullback-Leibler divergence,
or histogram intersection
for either the normalized or LBP-filtered pixel luminance values.
The remaining 10 measures were capture-related methods found in ISO/IEC TR 29794-5:2010 \cite{ISO-IEC-29794-5-TR-FaceQuality-100312},
namely general image contrast, global contrast factor,
mean/variance/skewness/kurtosis of pixel luminance,
exposure, sharpness, inter-eye distance, and blur.
In the evaluation, the face image utility was first derived via FR comparisons using 2 unnamed black-box COTS FR systems,
labeling images per system either
as high-quality if their minimum mated comparison score was greater than a threshold for 60\% FNMR \cite{ISO-IEC-2382-37-170206},
as low-quality if their maximum mated score was below a threshold for 30\% FNMR,
or leaving them unlabelled otherwise.
The 17 measures were then examined in terms of their correlation with the FR-system-derived utility,
and by means of FNMR ERC plots.
Based thereon, 11 measures were selected to create random forest models,
namely the 3 histogram symmetry measures and all capture-related measures except variance and skewness.
Random forest training used the utility labels for the 2 individual systems, in addition to the union and intersection thereof.

\ifdefined\ARXIVVERSION
\else
\clearpage
\fi
\subsection{Monolithic - Commonalities}
\label{sec:fiqaa-monolithic}

\begin{table*}
	\caption{\label{tab:fiqaa-monolithic} Monolithic FIQA literature in reverse chronological order.}
	\centering
	\tableFontSize
	\setlength{\tabcolsep}{1.55pt}\begin{tabular}{>{\raggedright\arraybackslash}p{0.075\linewidth}>{\raggedright\arraybackslash}p{0.057\linewidth}>{\raggedright\arraybackslash}p{0.585\linewidth}>{\raggedright\arraybackslash}p{0.254\linewidth}}
\hline
\textbf{Reference} & \textbf{Aspects} & \textbf{Method(s)} & \textbf{Datasets} \\
\hline

2021 \cite{Fu-FQA-DeepInsightMeasuring-WACV-2022} & \attrDl{}\attrDint{} & Evaluation of 6 monolithic FIQAAs, 10 IQAAs, and 9 factor-specific hand-crafted methods. \textit{} & LFW, VGGFace2, BioSecure \\\hline
2021 \cite{Fu-FQA-FaceMask-FGR-2021} & \attrDl{}\attrDint{} & Evaluation of monolithic FIQAAs \cite{Chen-FQA-LearningToRank-SPL-2015}\cite{Terhorst-FQA-SERFIQ-CVPR-2020}\cite{Hernandezortega-FQA-FaceQnetV1-2020}\cite{Meng-FRwithFQA-MagFace-arXiv-2021} on face images without masks, with real face masks, and images with synthesized masks. \textit{} & In-house \cite{Damer-FR-FaceMaskExtended-IET-2021} \\\hline
2021 \cite{Fu-FQA-RelativeContributionsOfFacialParts-BIOSIG-2021} & \attrDl{}\attrDuat{}\newline\attrFe{} & CNN IQA \cite{Kang-IQA-NoReferenceCNN-CVPR-2014} on various facial areas (eyes, nose, mouth, averaged fusion) and on cropped or aligned images. Compared against monolithic FIQAAs \cite{Chen-FQA-LearningToRank-SPL-2015}\cite{Hernandezortega-FQA-FaceQnetV0-ICB-2019}\cite{Terhorst-FQA-SERFIQ-CVPR-2020}\cite{Meng-FRwithFQA-MagFace-arXiv-2021}. \textit{} & LFW, VGGFace2 \\\hline
2021 \cite{Chen-FQA-LightQNet-SPL-2021} & \attrDl{}\attrDfrt{} & Identification quality (IDQ) loss focused on the FR comparison threshold, used to train a FIQA branch in a frozen FR network, in turn used for knowledge distillation to train a separate lightweight FIQA network. \textit{Open source.} & MS1MV2, LFW, CFP, CPLFW, IJB-B, Adience \\\hline
2021 \cite{Meng-FRwithFQA-MagFace-arXiv-2021} & \attrDl{}\attrDint{} & Extends ArcFace \cite{Deng-ArcFace-IEEE-CVPR-2019} training loss with FR feature embedding magnitude-aware angular margin and regularization, so that magnitude corresponds to quality. \textit{Open source.} & MS1MV2, LFW, CFP, AgeDB, CALFW, CPLFW, IJB-B, IJB-C \\\hline
2021 \cite{Ou-FQA-SimilarityDistributionDistance-arXiv-2021} & \attrDl{}\attrDfrt{} & The Wasserstein distance between FR comparison score sets for randomly selected mated and non-mated pairs is used to form ground truth QSs for FIQA network training with Huber loss. \textit{Open source.} & MS1MV2, CASIA-WebFace, LFW, Adience, UTKFace, IJB-C \\\hline
2021 \cite{Chen-FRwithFQA-ProbFace-arXiv-2021} & \attrDl{}\attrDfri{} & Reduces the \cite{Shi-FRwithFQA-ProbabilisticFaceEmbeddings-ICCV-2019} concept to a single uncertainty scalar, adding regularization relative to mini-batch uncertainty average, plus two uncertainty-aware identification loss variants, and the network uses multi-layer fusion. \textit{Open source.} & MS1MV2, LFW, CFP, CALFW, CPLFW, AgeDB, IJB-B, VGGFace2 \\\hline
2020 \cite{Xie-FQA-PredictiveUncertaintyEstimation-BMVC-2020} & \attrDl{}\attrDfrt{} & ResNet18 \cite{heDeepResidualLearning2015} FIQA model trained on ResNet34 \cite{heDeepResidualLearning2015} FR model ground truth scores, computing loss for the predicted QS minimum of each image pair. \textit{} & VGGFace2, IJB-C \\\hline
2020 \cite{Hernandezortega-FQA-FaceQnetV1-2020} & \attrDl{}\attrDfrt{} & Same as \cite{Hernandezortega-FQA-FaceQnetV0-ICB-2019}, but with dropout before the first fully connected layer, and multiple FR feature extractors to obtain the ground truth QSs. \textit{Continuation of \cite{Hernandezortega-FQA-FaceQnetV0-ICB-2019}. Open source. Benchmarked in NIST FRVT QA \cite{Grother-FQA-4thDraftOngoingFRVT-2021}.} & VGGFace2, LFW, CyberExtruder, BioSecure \\\hline
2020 \cite{Chang-FRwithFQA-UncertaintyLearning-CVPR-2020} & \attrDl{}\attrDint{} & Learns both uncertainty and FR features: Either 1. KL divergence loss to train an entire network, or 2. fixed FR network extension with loss relative to subject feature centers. Builds upon the uncertainty vector concept of \cite{Shi-FRwithFQA-ProbabilisticFaceEmbeddings-ICCV-2019}. \textit{} & MS-Celeb-1M, LFW, MegaFace, CFP, YTF, IJB-C \\\hline
2020 \cite{Terhorst-FQA-SERFIQ-CVPR-2020} & \attrDl{}\attrDfri{} & QS based on comparing embeddings from 100 random subnetworks; Works on FR networks trained with dropout, or by adding a network on top. \textit{Open source.} & MS-Celeb-1M, FERET, Adience, LFW \\\hline
2019 \cite{Zhao-FQA-SemiSupervisedCNN-ICCPR-2019} & \attrDl{}\attrDhgt{} & CNN trained on binary labels derived automatically based on fewer manual labels and non-DL methods. \textit{Predicts scalar QSs after binary training.} & In-house, CASIA-WebFace \\\hline
2019 \cite{Shi-FRwithFQA-ProbabilisticFaceEmbeddings-ICCV-2019} & \attrDl{}\attrDfri{} & Based on a pretrained FR network, trains separate two-layer perceptron network to measure per-feature-dimension uncertainty, compares via MLS (Mutual Likelihood Score). \textit{Open source.} & CASIA-WebFace, MS-Celeb-1M, LFW, YTF, MegaFace, CFP, IJB-A, IJB-C, IJB-S \\\hline
2019 \cite{Hernandezortega-FQA-FaceQnetV0-ICB-2019} & \attrDl{}\attrDfrt{} & Frozen FR-pretrained ResNet-50 \cite{heDeepResidualLearning2015}, training two new final layer replacements on QSs derived from FR features \vs{} BioLab-ICAO\cite{Ferrara-FQA-BioLabICAO-TIFS-2012}-selected references. \textit{Open source. Benchmarked in NIST FRVT QA \cite{Grother-FQA-4thDraftOngoingFRVT-2021}.} & VGGFace2, BioSecure \\\hline
2019 \cite{Yang-FQA-DFQA-ICIG-2019} & \attrDl{}\attrDhgt{} & ``DFQA'', a SqueezeNet\cite{iandolaSqueezeNetAlexNetlevelAccuracy2016}-based two-branch CNN; training with SVR loss; ground truth QSs generated by another CNN, in turn trained using 3000 rule-guided human QS labels. \textit{Direct SqueezeNet successor: SqueezeNext \cite{gholamiSqueezeNextHardwareAwareNeural2018}.} & ImageNet, IJB-A, MS-Celeb-1M, CASIA-WebFace, VGGFace2, LFW \\\hline
2018 \cite{Wasnik-FQA-EvaluationSmartphoneCNN-BTAS-2018} & \attrDl{}\attrDhgt{} & 14 methods: 2 FIQA CNNs, 5 non-FIQA CNNs, 3 non-FIQA mobile CNNs, 3 Hand-crafted, 1 COTS; Binary training labels (good\slash bad). \textit{Considers FIQA in the context of smartphone FR in addition to non-smartphone data.} & In-house, CAS-PEAL, Extended Yale, AR, FRGC, NCKU face, ChokePoint, SCface \\\hline
2018 \cite{Qi-FQA-VideoFrameCNN-ICB-2018} & \attrDl{}\attrDfrt{}\attrV{} & CNN with inception module, trained using gallery FR comparison score minima for detected faces. \textit{} & In-house, PaSC, ChokePoint, CMU-FIA \\\hline
2017 \cite{Bestrowden-FQA-FromHumanAssessments-arXiv-2017} & \attrDl{}\attrDfrt{}\newline\attrFt{} & 5 methods, using either human or FR-based labels, DL or L2R \cite{Chen-FQA-LearningToRank-SPL-2015} features, and SVR or L2R models. \textit{Another paper version is \cite{Bestrowden-FQA-FromHumanAssessments-TIFS-2018}.} & LFW, IJB-A, CASIA-WebFace \\\hline
2016 \cite{Hu-FQA-IlluminationKPLSR-PIC-2016} & \attrNondl{}\attrDfrt{} & Kernel Partial Least Squares Regression using mean luminance and Laplacian of $\medmuskip=0mu10\times{}10$ image sub-blocks. \textit{} & CAS-PEAL, FERET, MIT, FEI, AT\&T \\\hline
2015 \cite{Vignesh-FQA-VideoCNN-GlobalSIP-2015} & \attrDl{}\attrDfrt{}\attrV{} & CNN, PCA whitened input, QS labels via MSM. \textit{} & ChokePoint \\\hline
2015 \cite{Chen-FQA-LearningToRank-SPL-2015} & \attrDl{}\attrDhgt{}\newline\attrFt{} & 2-stage learning to rank for five feature extractors: CNN (Landmarks), HOG, Gist \cite{olivaModelingShapeScene2001}, Gabor, LBP (per-feature-vector QS formed by weighted sum). \textit{Can be considered non-DL by removing the CNN extractor. Open source.} & In-house\slash{}Unknown, FERET, FRGC, LFW, AFLW, SCface \\\hline
2013 \cite{Bharadwaj-FQA-HolisticRepresentations-ICIP-2013} & \attrNondl{}\attrDfrt{} & 4-class SVM on Gist\cite{olivaModelingShapeScene2001} or HOG. \textit{} & SCface, CAS-PEAL \\\hline
2012 \cite{Qu-FQA-GaussianLowPassIllumination-CCIS-2012} & \attrNondl{}\attrDuat{} & Gaussian low-pass filter \vs{} fixed 38-image-average reference. \textit{} & Extended Yale \\\hline
2012 \cite{Klare-FQA-ImpostorbasedUniqueness-BTAS-2012} & \attrNondl{}\attrDfri{} & ``Impostor-based Uniqueness Measure'', \ie{} FR \vs{} fixed image set as FIQA. \textit{} & In-house (Police) \\\hline
2011 \cite{Wong-FQA-PatchbasedProbabilistic-CVPRW-2011} & \attrNondl{}\attrDuat{} & Per block low-frequency 2D DCT components compared to ``ideal'' frontal face. \textit{} & FERET, CMU-PIE, ChokePoint \\\hline
2010 \cite{Sellahewa-FQA-LuminanceDistortion-TIM-2010} & \attrNondl{}\attrDuat{} & Luminance distortion from \cite{zhouwangUniversalImageQuality2002} against an averaged frontal face image. \textit{} & Extended Yale, AT\&T \\\hline

\end{tabular}

\end{table*}

The monolithic approaches do not have factor-specific subcategories by definition,
but the dominant commonalities and differences can be highlighted via the data aspect instead:
\begin{itemize}
\commonalityPart{Utility-agnostic training (\attrDuat{})}
The most recent approach \cite{Fu-FQA-RelativeContributionsOfFacialParts-BIOSIG-2021} evaluated a general IQA CNN from \cite{Kang-IQA-NoReferenceCNN-CVPR-2014}
for the purposes of FIQA primarily on different facial areas.
Besides \cite{Fu-FQA-RelativeContributionsOfFacialParts-BIOSIG-2021},
all of the works that exclusively proposed monolithic \attrDuat{} approaches
\cite{Sellahewa-FQA-LuminanceDistortion-TIM-2010}%
\cite{Wong-FQA-PatchbasedProbabilistic-CVPRW-2011}%
\cite{Qu-FQA-GaussianLowPassIllumination-CCIS-2012}
happened to rely on model data derived from a fixed set of training images,
and were non-DL.
Both \cite{Qu-FQA-GaussianLowPassIllumination-CCIS-2012} and \cite{Sellahewa-FQA-LuminanceDistortion-TIM-2010} directly compared the input against an averaged image,
while \cite{Wong-FQA-PatchbasedProbabilistic-CVPRW-2011} compared against Gaussian distributions derived from the training images.
\\
A few other works proposed both factor-specific and monolithic (\attrDuat{}) FIQAAs,
namely \cite{Kryszczuk-FQA-OnFaceImageQualityMeasures-MMUA-2006}\cite{Kryszczuk-FQA-ScoreAndSignalLevelGMM-EUSIPCO-2006}, which contained a monolithic average face image correlation measure,
and \cite{Damer-FRwithFQA-PersonalizedFaceReferenceVideo-FFER-2015}, which proposed to use the Viola-Jones face detection confidence as a monolithic FIQAA.
To avoid duplicates, this literature is only listed in the factor-specific \autoref{tab:fiqaa-factor} and introduced in \autoref{sec:fiqaa-literature-factor}.
\commonalityPart{Human ground truth training (\attrDhgt{})}
These FIQA approaches were trained to estimate ground truth QS labels that stemmed from human assessments
\cite{Chen-FQA-LearningToRank-SPL-2015}%
\cite{Wasnik-FQA-EvaluationSmartphoneCNN-BTAS-2018}%
\cite{Yang-FQA-DFQA-ICIG-2019}%
\cite{Zhao-FQA-SemiSupervisedCNN-ICCPR-2019}.
Some of these works automatically transferred the human QSs to additional unlabeled images to extend the available training data
\cite{Yang-FQA-DFQA-ICIG-2019}%
\cite{Zhao-FQA-SemiSupervisedCNN-ICCPR-2019}.
\commonalityPart{FR-based ground truth training (\attrDfrt{})}
These approaches obtained training data from FR models
\cite{Bharadwaj-FQA-HolisticRepresentations-ICIP-2013}%
\cite{Vignesh-FQA-VideoCNN-GlobalSIP-2015}%
\cite{Hu-FQA-IlluminationKPLSR-PIC-2016}%
\ifdefined\ARXIVVERSION
\cite{Bestrowden-FQA-FromHumanAssessments-arXiv-2017}%
\else
\cite{Bestrowden-FQA-FromHumanAssessments-arXiv-2017}
\fi
\cite{Qi-FQA-VideoFrameCNN-ICB-2018}%
\cite{Hernandezortega-FQA-FaceQnetV0-ICB-2019}%
\cite{Hernandezortega-FQA-FaceQnetV1-2020}%
\cite{Xie-FQA-PredictiveUncertaintyEstimation-BMVC-2020}%
\cite{Ou-FQA-SimilarityDistributionDistance-arXiv-2021}%
\cite{Chen-FQA-LightQNet-SPL-2021}.
The majority of the monolithic approaches belong to this category.
\commonalityPart{FR-based inference (\attrDfri{})}
FIQAAs with FR-based inference utilize FR models as part of the quality assessment process even outside the training stage,
but do not alter the FR model training
\cite{Klare-FQA-ImpostorbasedUniqueness-BTAS-2012}%
\cite{Shi-FRwithFQA-ProbabilisticFaceEmbeddings-ICCV-2019}%
\cite{Terhorst-FQA-SERFIQ-CVPR-2020}%
\cite{Chen-FRwithFQA-ProbFace-arXiv-2021}.
A notably early non-DL variant in this category is \cite{Klare-FQA-ImpostorbasedUniqueness-BTAS-2012},
which directly compared the input image against a comparatively large fixed set of 1000 images from different subjects with a FR system to assess the quality.
The later approaches are DL-centric and estimate uncertainty for a FR model.
\commonalityPart{FR-integration (\attrDint{})}
FR-integrated FIQA methods not only use the FR model during inference,
but are fully integrated into the FR model,
meaning that FR and FIQA training are intertwined
\cite{Chang-FRwithFQA-UncertaintyLearning-CVPR-2020}%
\cite{Meng-FRwithFQA-MagFace-arXiv-2021}.
This concept has emerged more recently than the others.
\end{itemize}

\subsection{Monolithic - Literature introductions}
\label{sec:fiqaa-literature-monolithic}

\markAuthor{Sellahewa and Jassim} \cite{Sellahewa-FQA-LuminanceDistortion-TIM-2010}
used the luminance distortion component from the ``universal image quality index'' \cite{zhouwangUniversalImageQuality2002}
to compare a face input image against a fixed average reference image generated from a training set
(not to be confused with full-reference IQA, where a high-quality variant of the input image itself is known).
This method worked by sliding a $8 \times 8$ window
simultaneously over the input and reference image,
computing $2L_\mathrm{input}L_\mathrm{reference}/(L_\mathrm{input}^2 + L_\mathrm{reference}^2)$
therein with $L$ being the mean luminance,
and using the mean of all window results as the final $[0,1]$ QS.

\markAuthor{Wong \etal{}} \cite{Wong-FQA-PatchbasedProbabilistic-CVPRW-2011} presented a FIQAA for frontal face images.
Low-frequency 2D DCT (Discrete Cosine Transform) components were extracted for overlapping blocks of a normalized grayscale face image.
Per block, these were compared against
Gaussian distributions derived from a set of training images with frontal illumination,
and a final QS was formed by fusing the resulting probabilities.

\markAuthor{Klare and Jain} \cite{Klare-FQA-ImpostorbasedUniqueness-BTAS-2012} presented the impostor-based uniqueness measure (IUM),
an approach inherently adaptive to any used FR system.
It was computed for a face image by comparing it against a given set of ``impostor'' face images\slash feature vectors via the FR system itself.
Based on experiments, \cite{Klare-FQA-ImpostorbasedUniqueness-BTAS-2012} proposed to use 1,000 feature vectors from different subjects to form this set.
Note that the paper appeared to only utilize frontal face images
(from an operational police dataset).

\markAuthor{Qu \etal{}} \cite{Qu-FQA-GaussianLowPassIllumination-CCIS-2012} proposed a FIQAA based on Gaussian blur face model similarity.
The Gaussian blur was applied to the input image,
which was then compared, in terms of the normalized correlation,
against a fixed reference image formed by the average of 38 training images.
The paper evaluated a range of sizes for the Gaussian blur.
FR performance was not evaluated,
but an evaluation can be found as part of the illumination methods considered in \cite{Hu-FQA-IlluminationKPLSR-PIC-2016}.

\markAuthor{Bharadwaj \etal{}} \cite{Bharadwaj-FQA-HolisticRepresentations-ICIP-2013} trained a one-vs-all SVM for 4 quality bins using either sparsely pooled Histogram of Oriented Gradient (HOG)
or Gist \cite{olivaModelingShapeScene2001}
input features.
The quality bin training labels were obtained using 2 COTS FR systems on training images that had a single designated good\slash studio quality image in addition to several probe images per subject.

\markAuthor{Chen \etal{}} \cite{Chen-FQA-LearningToRank-SPL-2015} proposed the learning to rank approach with two stages.
In stage one a number of preexisting feature extractors were used on the input image,
and for each feature output vector thereof a RQS (Rank based Quality Score) was derived
as the features' weighted sum.
Stage two applied a polynomial kernel to the RQS output vector of stage one,
and again used the weighted sum of the resulting vector elements to obtain the final scalar RQS (normalized to $[0,100]$).
``Learning to rank'' refers to learning the various weights for the aforementioned weighted sums
so that each RQS differentiates between images from a number of training datasets
with a given assumed quality ordering (\eg{} some training dataset A is defined to be of higher quality than dataset B, which in turn is defined to be of higher quality than dataset C).
Conceptually, this approach does not have to use any deep learning,
but the evaluated FIQAA implementation incorporated a CNN for facial landmark detection as one of five feature extractors.
The other four (non-DL) feature extractors comprised
Gist \cite{olivaModelingShapeScene2001}, HOG, Gabor, and LBP.

In \cite{Vignesh-FQA-VideoCNN-GlobalSIP-2015} by \markAuthor{Vignesh \etal{}} a CNN was utilized to directly output a final FR-performance-focused QS for a $64 \times 64$ face image input.
The network had 4 convolutional layers and the face image input was preprocessed using PCA whitening.
Training this approach required a ground truth QS corresponding to each training image,
which the paper notably computed by comparing each given probe frame against a sequence of gallery frames via the MSM (Mutual Subspace Method) based on either LBP or HOG features.
Since the CNN itself only uses single-image input,
this ground truth QS generation could naturally be replaced by some single-image approach as well.

\markAuthor{Hu \etal{}} \cite{Hu-FQA-IlluminationKPLSR-PIC-2016} proposed to train a KPLSR (Kernel Partial Least Squares Regression) model for FIQA.
Two features were derived for $10 \times 10$ sub-blocks of an image,
forming a 200-dimensional feature vector as input for the KPLSR model.
These features were the mean luminance and Laplacian gradient per sub-block.
The training ground truth QSs were LBP-based FR comparison scores,
whereby each image pair consisted of one image with ``standard'' (\ie{} presumably good and unaltered) illumination,
and one image variant with reduced luminance\slash contrast.
A strong correlation between the FIQAA and the FR performance was demonstrated in the evaluation.

In \cite{Bestrowden-FQA-FromHumanAssessments-arXiv-2017} and \cite{Bestrowden-FQA-FromHumanAssessments-TIFS-2018},
\markAuthor{Best-Rowden and Jain} presented multiple FIQAA variants partially based on DL.
Five FIQAAs were evaluated, including the RQS approach of \cite{Chen-FQA-LearningToRank-SPL-2015}.
Of the four newly proposed FIQAAs,
three used training ground truth QSs derived from pairwise relative human assessments,
and one derived the ground truth QSs from FR-method-dependent comparison scores with manually selected gallery images.
Two of the methods used the 320-dimensional feature vector of a FR CNN \cite{wangFaceSearchScale2015} to train a SVR model for the QS prediction,
one method targeting the FR scores (Matcher Quality Values, ``MQV''),
the other targeting the human assessment ground truth (Human Quality Values, ``HQV-0'').
The CNN features were also used in another variant of the human ground truth methods,
which replaced the SVR with the L2R (learning to rank) approach of \cite{Chen-FQA-LearningToRank-SPL-2015} (``HQV-1'').
The fourth method trained the L2R approach of \cite{Chen-FQA-LearningToRank-SPL-2015} with the features described therein,
but for the human ground truth instead of the RQS dataset constraints \cite{Chen-FQA-LearningToRank-SPL-2015} (``HQV-2'').
In the evaluation, the CNN of \cite{wangFaceSearchScale2015} was also used as one of the FR algorithms, in addition to two unnamed COTS systems.
The methods HQV-2 and MQV showed the lowest improvements regarding FR performance.
The best FR improvements were achieved
using HQV-1 for the CNN \cite{wangFaceSearchScale2015},
and RQS \cite{Chen-FQA-LearningToRank-SPL-2015} for one of the COTS systems.

\markAuthor{Qi \etal{}} \cite{Qi-FQA-VideoFrameCNN-ICB-2018} used a CNN architecture with an inception module for FIQA.
Ground truth QS labels were established in form of gallery DL FR comparison dissimilarity score (\ie{} cosine distance) minima for detected faces in training video data.
In other words, each training probe image was compared to all training gallery images, and the best score was selected as the ground truth QS to train the FIQA network.
A pretrained VGG-16 \cite{simonyanVeryDeepConvolutional2015} and 
Inception-v3 \cite{szegedyRethinkingInceptionArchitecture2016}
network was used for the FR part.
The video frame FR performance improvement evaluation \ia{} compared against
the CNN approach of \markAuthor{Vignesh \etal{}} \cite{Vignesh-FQA-VideoCNN-GlobalSIP-2015}
and the learning to rank approach of \markAuthor{Chen \etal{}} \cite{Chen-FQA-LearningToRank-SPL-2015},
with the proposed CNN showing the best results.

\markAuthor{Wasnik \etal{}} \cite{Wasnik-FQA-EvaluationSmartphoneCNN-BTAS-2018} compared 14 methods for FIQA using 7 publicly available datasets (plus in-house datasets) in the context of smartphone FR.
Of the 14 methods,
10 were CNNs,
3 were hand-crafted,
and 1 was a COTS system (VeriLook 5.4 \cite{neurotechnology-face}).
Among the 3 hand-crafted methods,
2 were general IQAAs (BLIINDS-II \cite{saadBlindImageQuality2012}, BRISQUE \cite{mittalNoReferenceImageQuality2012}),
and 1 was \markAuthor{Wasnik \etal{}} \cite{Wasnik-FQA-SmartphoneISO297945-IWBF-2017}.
Among the 10 pretrained CNNs,
2 were meant specifically for FIQA (the illumination-focused FIQA \cite{Zhang-FQA-SubjectiveIlluminationResNet50-ICONIP-2017}, and the general FIQA \cite{Qi-FQA-VideoFrameCNN-ICB-2018}),
3 were mobile networks (MobileNetV2 \cite{sandlerMobileNetV2InvertedResiduals2019}, DenseNet-169 \cite{huangDenselyConnectedConvolutional2018}, NASNet \cite{zophLearningTransferableArchitectures2018}),
and the other 5 were AlexNet \cite{krizhevskyImageNetClassificationDeep2017}, VGG-16\slash VGG-19 \cite{simonyanVeryDeepConvolutional2015}, Inception \cite{szegedyGoingDeeperConvolutions2014}, and Xception \cite{cholletXceptionDeepLearning2017}.
Of the 2 FIQA-specific CNNs,
for \cite{Zhang-FQA-SubjectiveIlluminationResNet50-ICONIP-2017} a pretrained network provided by the authors was used,
and for \cite{Qi-FQA-VideoFrameCNN-ICB-2018} the network described therein was recreated
while using the training dataset of \cite{Wasnik-FQA-EvaluationSmartphoneCNN-BTAS-2018}.
To adapt the non-FIQA CNNs for the FIQA task,
the last three layers were replaced by fully connected layers of size 1024, 512 and 2,
2 being the number of training data classes.
So training images were either labeled good or bad regarding quality,
with the latter referring to presumed flaws for \eg{} illumination or pose.
Note that this means that the training did not directly target some ground truth QS produced via \eg{} an FR system.
Nevertheless, the best FR performance improvements in the evaluation
were achieved by the two larger FIQA-adapted CNNs AlexNet and Inception.
This evaluation used 5 separate datasets,
and the VeriLook SDK 5.4 \cite{neurotechnology-face} for FR comparisons.

\markAuthor{Yang \etal{}} \cite{Yang-FQA-DFQA-ICIG-2019} presented ``DFQA'',
a FIQA CNN based on SqueezeNet \cite{iandolaSqueezeNetAlexNetlevelAccuracy2016},
which itself was notably meant to provide performance comparable to AlexNet with 50× fewer parameters
(also note that by this point in time a direct successor exists, namely SqueezeNext \cite{gholamiSqueezeNextHardwareAwareNeural2018}).
However, it was not proven whether this performance equivalence is true for the biometric FIQA task here,
since \cite{Yang-FQA-DFQA-ICIG-2019} did not compare against any AlexNet-based FIQA variant,
\eg{} one analogous to their SqueezeNet-based approach, or the one used in \cite{Wasnik-FQA-EvaluationSmartphoneCNN-BTAS-2018}.
Most of the SqueezeNet architecture parts in the DFQA \cite{Yang-FQA-DFQA-ICIG-2019} network were represented in two functionally identical weight-sharing branches (also called ``streams'' in \cite{Yang-FQA-DFQA-ICIG-2019}),
each of which was followed by a (no longer weight-sharing) $1 \times 1$ kernel convolutional layer with $9 \times 9$ output.
Then the mean of the two outputs was fed to an average pooling layer,
resulting in the output feature vector.
The paper compared both Euclidean and SVR loss,
showing better results for the latter.
Different branch counts, 1 to 4, were evaluated as well.
For training,
3,000 images were first manually annotated with ground truth QS values,
using a defined set of rules to increase the QS objectivity\slash subject-independence.
These images were used to train another CNN, based on a pretrained SqueezeNet,
to predict ground truth QSs for the MS-Celeb-1M \cite{Guo-Face-MSCeleb1M-ECCV-2016} dataset,
which were then used to train the actual DFQA.

\markAuthor{Hernandez-Ortega \etal{}} created the open source FIQAA ``FaceQnet'' v0 \cite{Hernandezortega-FQA-FaceQnetV0-ICB-2019} and v1 \cite{Hernandezortega-FQA-FaceQnetV1-2020}.
As part of the training data preparation for both FaceQnet versions,
the BioLab-ICAO framework from \cite{Ferrara-FQA-BioLabICAO-TIFS-2012} was employed to select suitable high-quality images per subject,
which were used to compute the ground truth QSs for the subjects' remaining training images.
This ground truth QS computation consisted of the normalized Euclidean distances of embeddings produced by a number of FR feature extractors (three for v1; and only one, FaceNet \cite{schroffFaceNetUnifiedEmbedding2015}, for v0).
Both FaceQnet versions were based on a ResNet50 \cite{heDeepResidualLearning2015} model pretrained for FR using the VGGFace2 \cite{Cao-VGGFace2Dataset-FGR-2018} dataset,
replacing the final output layer with two fully connected layers.
Only these two new layers were trained, the rest of the network weights were frozen.
FaceQnet v1 extended the training architecture by adding dropout before the first fully connected layer.
\Ie{} the architecture of FaceQnet v1 and v0 after training are identical,
but FaceQnet v1 was trained with dropout and using ground truth QSs derived from multiple feature extractors.
Both versions used a 300-subject subset of the VGGFace2 \cite{Cao-VGGFace2Dataset-FGR-2018} for training.
At the time of writing, FaceQnet v0 and v1 are the only surveyed approaches that have been included in the report of the new NIST FRVT Quality Assessment campaign \cite{Grother-FQA-4thDraftOngoingFRVT-2021}.

\markAuthor{Shi and Jain} \cite{Shi-FRwithFQA-ProbabilisticFaceEmbeddings-ICCV-2019}
proposed PFE (Probabilistic Face Embeddings),
an approach to compute an uncertainty vector that directly corresponds to the FR feature vector for a single face image.
In other words, the two output vectors represent the Gaussian variance and mean, respectively.
The work focused on using the uncertainty to improve the FR comparisons,
so producing a single scalar QS was not the primary goal.
It was nevertheless noted that the uncertainty could be used for FIQA purposes,
and a part of the evaluations showed that filtering images by the inverse harmonic mean of the uncertainty vector elements can be more effective to improve FR performance than filtering using face detection scores.
So the uncertainty can certainly be considered as a kind of QS,
and a scalar QS can be derived from such a vector.
The implementation of \cite{Shi-FRwithFQA-ProbabilisticFaceEmbeddings-ICCV-2019} used a fixed pretrained FR network as basis to compute the FR feature vector (\ie{} Gaussian mean),
and trained an additional module for the uncertainty vector (\ie{} variance),
on the same training dataset used for the FR network.
The uncertainty module was a two-layer perceptron network,
using the same input as the FR layer that outputs the original feature vector.
To incorporate the uncertainty vector in the FR comparison,
a MLS (Mutual Likelihood Score) was proposed by \cite{Shi-FRwithFQA-ProbabilisticFaceEmbeddings-ICCV-2019},
which weighed and penalized feature dimensions depending on the uncertainty.
The uncertainty module training attempted to maximize this MLS for all genuine image pairs.
In addition, \cite{Shi-FRwithFQA-ProbabilisticFaceEmbeddings-ICCV-2019} explained how the uncertainty can be used to fuse embeddings for multiple images.

\markAuthor{Zhao \etal{}} \cite{Zhao-FQA-SemiSupervisedCNN-ICCPR-2019} trained a CNN for FIQA in a semi-supervised fashion.
First, binary labels (good\slash bad) were manually assigned to a number of images to train a preliminary version of the DL model.
This preliminary network then predicted labels for a different (larger) dataset in the second stage.
The third stage updated these labels utilizing various additional binary constraints
derived from the inter-eye distance, the pitch and yaw rotation, the contrast,
and further factors not listed in \cite{Zhao-FQA-SemiSupervisedCNN-ICCPR-2019} due to paper length limitations.
For all ``good'' labels predicted by the preliminary network,
the label were be changed to ``bad'' if any of these binary constraints were ``bad'',
but existing ``bad'' label predictions were not altered.
This newly labeled dataset was then used in the fourth and final stage to fine-tune the model.
Hinge loss was used during training for the binary classification task,
but after training the network was modified to output a [0, 1] scalar QS prediction instead.
It was noted that the CNN had better computational performance than the CNN proposed by \cite{Yu-FQA-LightCNNwithMFM-PRLE-2018}.

\markAuthor{Terhörst \etal{}} \cite{Terhorst-FQA-SERFIQ-CVPR-2020} proposed the open source ``SER-FIQ'' method in two variants,
measuring FR-model-specific quality
by comparing the output embeddings of a number of randomly chosen subnetworks,
\ie{} without requiring any ground truth QS training labels.
A QS was computed as the sigmoid of the negative mean of the Euclidean distances between all random subnetwork embeddings,
meaning that the computational complexity grows quadratically with respect to the number of subnetworks (100 were used in \cite{Terhorst-FQA-SERFIQ-CVPR-2020}).
The ``same model'' variant of SER-FIQ can be used on FR networks trained using dropout,
without additional training.
For this variant's implementation in \cite{Terhorst-FQA-SERFIQ-CVPR-2020},
the random subnetwork passes used the last two FR layers.
The other variant was the ``on-top model'',
meaning that a small additional network was trained with dropout on top of the FR model to transform its FR embeddings.
Five layers with dropout were used in the implementation,
which included the identity classification layer for training.
Removing that, the first and last layer of the network had the same dimensions as the FR embedding.
Evaluations used FaceNet \cite{schroffFaceNetUnifiedEmbedding2015} and ArcFace \cite{Deng-ArcFace-IEEE-CVPR-2019} for FR,
and selected images using QSs from both SER-FIQ variants,
FaceQnet v0 \cite{Hernandezortega-FQA-FaceQnetV0-ICB-2019},
an approach proposed by Best-Rowden in \cite{Bestrowden-FQA-FromHumanAssessments-TIFS-2018},
three general IQAAs (BRISQUE \cite{mittalNoReferenceImageQuality2012}, NIQE \cite{mittalMakingCompletelyBlind2013}, PIQE \cite{venkatanathnBlindImageQuality2015}),
as well as a COTS system (Neurotec Biometric SDK 11.1 \cite{neurotechnology-face}).
The SER-FIQ ``on-top model'' was noted to mostly outperform all baseline approaches,
and to always deliver close to top performance.
The ``same model'' approach mostly outperformed the baseline methods by a larger margin,
showing especially strong FNMR (False Non-Match Rate) performance improvements for a fixed FMR (False Match Rate) of 0.001.

Extending the PFE concept of \markAuthor{Shi and Jain} \cite{Shi-FRwithFQA-ProbabilisticFaceEmbeddings-ICCV-2019},
\markAuthor{Chang \etal{}} \cite{Chang-FRwithFQA-UncertaintyLearning-CVPR-2020} proposed two methods to learn both uncertainty (variance) and feature (mean) at the same time,
without a separate module.
This means that the uncertainty can improve the overall training by reducing the influence of low quality images,
which implies that the FR performance may improve even if the uncertainty is not used after training,
although it is noted that this kind of quality attention can reduce performance when only low quality cases are considered after training.
By omitting a separate uncertainty vector for comparisons,
the MLS of \cite{Shi-FRwithFQA-ProbabilisticFaceEmbeddings-ICCV-2019} does not have to be used,
thus avoiding increased computational complexity as evaluated in \cite{Chang-FRwithFQA-UncertaintyLearning-CVPR-2020}.
One of the two methods in \cite{Chang-FRwithFQA-UncertaintyLearning-CVPR-2020} was ``classification-based''
and learned an entire FR network with both regular feature and uncertainty output,
together forming a sampling representation for training,
using the reparameterization trick \cite{kingmaAutoEncodingVariationalBayes2014} to enable backpropagation.
Instead of using the MLS, the cost function consisted of a softmax classification loss,
plus a regularization term to control the uncertainty aspect.
The latter was the Kullback-Leibler divergence scaled by a scalar hyper-parameter,
comparing the mean and variance output relative to a normal distribution.
The other learning method of \cite{Chang-FRwithFQA-UncertaintyLearning-CVPR-2020} was ``regression-based'' and more akin to the separate uncertainty module training concept of \cite{Shi-FRwithFQA-ProbabilisticFaceEmbeddings-ICCV-2019}:
Similar to \cite{Shi-FRwithFQA-ProbabilisticFaceEmbeddings-ICCV-2019} it began by using a FR feature network trained in isolation,
then the weights were frozen and uncertainty output was added.
But in contrast to \cite{Shi-FRwithFQA-ProbabilisticFaceEmbeddings-ICCV-2019}
the FR features (mean) were not frozen with the rest of the pretrained layers,
and the method continued training them simultaneously with the uncertainty,
using loss based on the per-subject feature vector center derived from the isolated FR network stage.
As part of the evaluations on multiple FR base models in \cite{Chang-FRwithFQA-UncertaintyLearning-CVPR-2020},
the two methods of \cite{Chang-FRwithFQA-UncertaintyLearning-CVPR-2020} (using cosine similarity for comparisons) and the method of \cite{Shi-FRwithFQA-ProbabilisticFaceEmbeddings-ICCV-2019} (using MLS for comparisons, including fusion where applicable)
were compared.
The ``classification-based'' method \cite{Chang-FRwithFQA-UncertaintyLearning-CVPR-2020} was found to mostly result in better performance increases than the PFE method from \cite{Shi-FRwithFQA-ProbabilisticFaceEmbeddings-ICCV-2019},
while the ``regression-based'' method \cite{Chang-FRwithFQA-UncertaintyLearning-CVPR-2020} appeared either worse or better depending on the scenario (and further examination in the future was considered due to some observed performance regression with respect to the FR baseline).

\markAuthor{Xie \etal{}} \cite{Xie-FQA-PredictiveUncertaintyEstimation-BMVC-2020}
proposed the ``PCNet'' (Predictive Confidence Network) FIQAA,
and evaluated it \ia{} against the conceptually similar FaceQnet v0 \cite{Hernandezortega-FQA-FaceQnetV0-ICB-2019}.
In contrast to FaceQnet, the network was trained from scratch, a more lightweight ResNet18 \cite{heDeepResidualLearning2015} was employed, and a different training scheme was used.
To obtain comparison scores for FIQA training, a separate ResNet34 was first trained for FR, using cosine similarity for the comparisons.
This was done twice, separately for both halves of a dataset,
so that FR comparison scores were not computed on FR training data.
Only mated image pairs were used in the process.
The FIQA model, PCNet, was then trained to predict a QS for each image of a pair,
the loss being the squared difference between the pair's QS prediction minimum and the pair's previously computed FR comparison score.
PCNet (using ResNet18) consistently outperformed
FaceQnet v0 \cite{Hernandezortega-FQA-FaceQnetV0-ICB-2019} and MNet \cite{Xie-FR-MulticolumnNetworks-BMVC-2018} (both using ResNet50)
in the evaluations,
which \ia{} tested image-to-image verification improvements via ERC plots,
and set-to-set verification, with set feature fusion weighted by the per-image quality.
In the tests, three open source FR models and VGGFace2 \cite{Cao-VGGFace2Dataset-FGR-2018} were used.

\markAuthor{Chen \etal{}} \cite{Chen-FRwithFQA-ProbFace-arXiv-2021}
proposed ``ProbFace'' based on the PFE (Probabilistic Face Embeddings) concept from \markAuthor{Shi and Jain} \cite{Shi-FRwithFQA-ProbabilisticFaceEmbeddings-ICCV-2019}.
The FR base model was fixed during training,
similar to \cite{Shi-FRwithFQA-ProbabilisticFaceEmbeddings-ICCV-2019}.
But instead of an uncertainty vector with the same dimension as the FR feature vector,
ProbFace uses a single uncertainty scalar.
As a result the required storage space was reduced
and the MLS (Mutual Likelihood Score) comparison metric from \cite{Shi-FRwithFQA-ProbabilisticFaceEmbeddings-ICCV-2019}
was simplified to an uncertainty-scalar-adjusted cosine FR embedding comparison.
In addition, the uncertainty training was regularized relative to the average uncertainty of each mini-batch,
and two uncertainty-aware identification loss variants were introduced to consider both mated and non-mated pairs during training.
Of the latter, only one was used for the final ProbFace method configuration, namely uncertainty-aware triplet loss.
Furthermore, ProbFace derived the uncertainty from multiple fused FR base network layers,
to more directly incorporate both low-level local texture information and high-level global semantic information.
The uncertainty (\ie{} quality) assessment aspect was studied mainly in terms of FR comparison improvements
against other FR models,
so no comparisons against pure FIQAAs were included.
ProbFace was however also evaluated against the PFE approach from \cite{Shi-FRwithFQA-ProbabilisticFaceEmbeddings-ICCV-2019}
in terms of ``risk-controlled face recognition'',
including an evaluation method akin to ERC,
showing that both ProbFace and PFE can be effective in a more general FIQA context.

\markAuthor{Ou \etal{}} \cite{Ou-FQA-SimilarityDistributionDistance-arXiv-2021}
proposed SDD-FIQA (Similarity Distribution Distance for FIQA),
an approach to generate ground truth QS training data
by computing the Wasserstein distance between FR comparison score sets
that include both mated and non-mated pairs.
For this purpose 
an equal number of mated and non-mated comparison pairs were selected randomly,
and the average of multiple computation rounds was used to obtain the final ground truth QS for each image.
A FIQA network was then trained with Huber loss using such QS ground truth data.
Similar to FaceQnet \cite{Hernandezortega-FQA-FaceQnetV0-ICB-2019}\cite{Hernandezortega-FQA-FaceQnetV1-2020},
a pretrained FR network was taken to form the base of the FIQA network,
replacing the embedding and classification layer with a fully connected layer for the quality score output,
and applying 50\% dropout,
except here the base network part was not frozen during training.
The SDD-FIQA network was evaluated on various datasets against
FaceQnet v0 \cite{Hernandezortega-FQA-FaceQnetV0-ICB-2019} and v1 \cite{Hernandezortega-FQA-FaceQnetV1-2020},
PFE \cite{Shi-FRwithFQA-ProbabilisticFaceEmbeddings-ICCV-2019},
SER-FIQ \cite{Terhorst-FQA-SERFIQ-CVPR-2020},
PCNet \cite{Xie-FQA-PredictiveUncertaintyEstimation-BMVC-2020},
as well as three IQAAs (BLIINDS-II \cite{saadBlindImageQuality2012}, BRISQUE \cite{mittalNoReferenceImageQuality2012}, PQR \cite{Zeng-IQA-ProbabilisticQualityRepresentation-ICIP-2018}).
The SDD-FIQA model showed superior performance in most cases.
An ERC ``Area Over Curve'' (AOC) measure was also introduced as part of the evaluation,
and the influence of the incorporation of non-mated pairs was demonstrated in an ablation study.

The ``MagFace'' approach from \markAuthor{Meng \etal{}} \cite{Meng-FRwithFQA-MagFace-arXiv-2021}
expanded on the idea of FR with integrated FIQA.
In contrast to previous approaches such as ProbFace \cite{Chen-FRwithFQA-ProbFace-arXiv-2021}, the data uncertainty learning approach from \cite{Chang-FRwithFQA-UncertaintyLearning-CVPR-2020}, or PFE \cite{Shi-FRwithFQA-ProbabilisticFaceEmbeddings-ICCV-2019},
MagFace does not have separate quality or uncertainty output at all.
Instead the quality is directly indicated by the magnitude of the FR feature vector.
The approach works by extending the ArcFace \cite{Deng-ArcFace-IEEE-CVPR-2019} training loss,
changing the angular margin to a magnitude-aware variant,
and adding magnitude regularization.
On one hand the magnitude-aware angular margin increases the margin for larger magnitudes,
penalizing higher magnitudes for lower quality samples,
and on the other hand the regularization rewards higher magnitudes scaled by a hyperparameter.
As a result FR feature vectors for higher quality images are pulled closer to the class center with larger magnitudes,
and vice versa for lower quality samples.
The magnitude is bounded during training, so deriving a normalized quality score only requires linear scaling.
Furthermore, the design also implies that the FR comparison function after training can be left unchanged from ArcFace \cite{Deng-ArcFace-IEEE-CVPR-2019},
while other approaches like ProbFace \cite{Chen-FRwithFQA-ProbFace-arXiv-2021} and PFE \cite{Shi-FRwithFQA-ProbabilisticFaceEmbeddings-ICCV-2019} have to specifically include quality in the comparison function to introduce an effect.
The magnitude quality aspect itself can be separately used \eg{} to facilitate weighted feature fusion,
or for FIQA.
MagFace was evaluated both in a FR context and in a FIQA context.
The FIQA evaluation included ERC results on multiple datasets against
FaceQnet v0 \cite{Hernandezortega-FQA-FaceQnetV0-ICB-2019},
SER-FIQ \cite{Terhorst-FQA-SERFIQ-CVPR-2020},
the data uncertainty learning approach from \cite{Chang-FRwithFQA-UncertaintyLearning-CVPR-2020},
and the three general IQAAs that were also used in the SER-FIQ \cite{Terhorst-FQA-SERFIQ-CVPR-2020} evaluation (BRISQUE \cite{mittalNoReferenceImageQuality2012}, NIQE \cite{mittalMakingCompletelyBlind2013}, PIQE \cite{venkatanathnBlindImageQuality2015}),
showing that MagFace can achieve superior or similar FIQA results compared the other methods.

\markAuthor{Chen \etal{}} \cite{Chen-FQA-LightQNet-SPL-2021} proposed
the identification quality (IDQ) training loss
and the use of knowledge distillation to train a lightweight FIQA network called ``LightQNet''.
The core idea of the IDQ training loss was to concentrate on the FR comparison threshold boundary.
Thus, for a mini-batch with comparison pairs of the same identity,
the IDQ loss incorporated a FR threshold hyperparameter to compute pairwise ground truth labels,
and the pairwise predicted QS was the minimum of each pair's predicted image QSs (compare with PCNet \cite{Xie-FQA-PredictiveUncertaintyEstimation-BMVC-2020}).
The pairwise ground truth labels could be ``hard'' binary labels,
\ie{} either above or below the FR threshold hyperparameter,
but better performance was achieved with a ``soft'' exponential-based label variant that used the threshold as an offset,
in addition to a scaling hyperparameter.
A FIQA branch in a frozen FR network (similar to FaceQnet \cite{Hernandezortega-FQA-FaceQnetV0-ICB-2019}\cite{Hernandezortega-FQA-FaceQnetV1-2020})
and a separate lightweight FIQA network (LightQNet)
were trained using IDQ loss.
Additionally using the FIQA branch as a teacher for the lightweight network
was shown to improve the lightweight network's predictive performance over pure IDQ loss training.
The proposed approach was evaluated against
FaceQnet v1 \cite{Hernandezortega-FQA-FaceQnetV1-2020}, SER-FIQ \cite{Terhorst-FQA-SERFIQ-CVPR-2020}, PFE \cite{Shi-FRwithFQA-ProbabilisticFaceEmbeddings-ICCV-2019}, and PCNet \cite{Xie-FQA-PredictiveUncertaintyEstimation-BMVC-2020}
with better or competitive results on various datasets,
and substantial (approximately threefold at the lowest) computational performance improvements for the lightweight network were observed.

\markAuthor{Fu \etal{}} \cite{Fu-FQA-RelativeContributionsOfFacialParts-BIOSIG-2021} evaluated 
a no-reference general IQA CNN \cite{Kang-IQA-NoReferenceCNN-CVPR-2014}
for the purposes of FIQA in terms of FR utility.
The IQA CNN was applied to various rectangular facial areas in particular,
namely the eyes, nose, and mouth,
to examine the areas' individual usefulness for FIQA.
These area-specific QSs were also fused by averaging them.
In addition to the facial area assessments, the IQA CNN was tested with
tightly cropped image variants
and image variants aligned for FR input.
A clear correlation of the IQA CNN output with FR utility for multiple FR models
was demonstrated especially for the eyes area on the VGGFace2 \cite{Cao-VGGFace2Dataset-FGR-2018} dataset,
although results could not compete with the tested specialized monolithic FIQAAs
(learning to rank \cite{Chen-FQA-LearningToRank-SPL-2015},
FaceQnet v0 \cite{Hernandezortega-FQA-FaceQnetV0-ICB-2019},
SER-FIQ \cite{Terhorst-FQA-SERFIQ-CVPR-2020}, and
MagFace \cite{Meng-FRwithFQA-MagFace-arXiv-2021}).

\markAuthor{Fu \etal{}} \cite{Fu-FQA-FaceMask-FGR-2021} investigated
the effect of face masks on a number of monolithic FIQAAs,
namely 
the learning to rank approach \cite{Chen-FQA-LearningToRank-SPL-2015},
FaceQnet v1 \cite{Hernandezortega-FQA-FaceQnetV1-2020},
SER-FIQ \cite{Terhorst-FQA-SERFIQ-CVPR-2020}, and
MagFace \cite{Meng-FRwithFQA-MagFace-arXiv-2021}.
The FIQAA performance was tested on regular face images without masks,
on images with real masks of varying types,
and on images with synthesized masks.
Synthetic masks were automatically drawn based on detected facial landmarks in a variety of solid colors,
\ie{} without realistic shading,
on top of images without masks.
Results showed a drop in predicted QSs for images with masks for all tested FIQAAs,
corresponding to reduced FR performance of both automatic systems and human experts,
and the QS distributions for images with/without masks were especially distinct for MagFace \cite{Meng-FRwithFQA-MagFace-arXiv-2021} and the learning to rank approach \cite{Chen-FQA-LearningToRank-SPL-2015}.
Differences between results for the real and the synthetic masks were observed as well,
indicating that improved synthesis realism may be desirable for this kind of evaluation in the future.
Additionally, network attention visualizations were examined for FaceQnet v1 \cite{Hernandezortega-FQA-FaceQnetV1-2020} and MagFace \cite{Meng-FRwithFQA-MagFace-arXiv-2021}.
Note that this work was purely about the evaluation of existing FIQAAs,
thus its categorization is based on the included FIQAAs instead of proposed FIQAAs.

\markAuthor{Fu \etal{}} \cite{Fu-FQA-DeepInsightMeasuring-WACV-2022} further evaluated the FR utility prediction performance of
6 monolithic FIQAAs \cite{Chen-FQA-LearningToRank-SPL-2015}\cite{Shi-FRwithFQA-ProbabilisticFaceEmbeddings-ICCV-2019}\cite{Terhorst-FQA-SERFIQ-CVPR-2020}\cite{Hernandezortega-FQA-FaceQnetV1-2020}\cite{Ou-FQA-SimilarityDistributionDistance-arXiv-2021}\cite{Meng-FRwithFQA-MagFace-arXiv-2021},
10 general IQAAs (\ia{} the IQA CNN from \cite{Kang-IQA-NoReferenceCNN-CVPR-2014}, BRISQUE \cite{mittalNoReferenceImageQuality2012}, NIQE \cite{mittalMakingCompletelyBlind2013}, PIQE \cite{venkatanathnBlindImageQuality2015}),
and 9 factor-specific hand-crafted measures (\ia{} for blur, symmetry, inter-eye distance).
Most of the general IQAAs did improve FR performance in ERC tests,
but overall they were outperformed by the best monolithic FIQAAs.
The factor-specific hand-crafted measure results were inconsistent across datasets,
indicating that these individual measures do not generalize sufficiently.
Assessments from the hand-crafted measures also did not correlate strongly with the other IQAA/FIQAA assessments,
while various IQAAs and FIQAAs did exhibit higher assessment overlaps.
Network attention visualizations for some of the DL IQAAs and FIQAAs
illustrated that the tested IQAAs incorporated more image background information than the FIQAAs,
which concentrated more on the face region.
Similarly to \cite{Fu-FQA-FaceMask-FGR-2021},
note that this work focused on the evaluation of existing FIQAAs,
thus its categorization is based on the included FIQAAs instead of proposed FIQAAs.

\section{Evaluation}
\label{sec:evaluation}

The first subsection hereunder introduces a common methodology to evaluate FIQAAs (or other biometric quality assessment algorithms) with respect to their ability to assess the biometric utility of samples for a given FR system and dataset.
In the second subsection we present a concrete evaluation for 14 FIQAAs and discuss the evaluation configuration, results and limitations.

\subsection{Error-versus-Reject-Characteristic}
\label{sec:evaluation-erc}

An Error-versus-Reject-Characteristic (ERC) can be plotted
to evaluate the predictive performance of quality assessment algorithms,
as proposed by \markAuthor{Grother and Tabassi} \cite{Grother-SampleQualityMetricERC-PAMI-2007}.
In the FIQA literature the ``C'' in ERC is occasionally also referred to as ``Curve''.
It is currently intended to standardize the ERC concept in the next (third) edition of ISO/IEC 29794-1 \cite{ISO-IEC-29794-1-WD}
under a different name that replaces the ``reject'' term to avoid confusion with the meaning of ``reject'' in ISO/IEC 2382-37 \cite{ISO-IEC-2382-37-170206}.

In the context of FIQA,
a FR system and a face dataset with subject identity labels is required in addition to the FIQAA to compute the ERC.
The FR system compares face image pairs with a fixed comparison threshold \cite{ISO-IEC-2382-37-170206}
to decide between match \cite{ISO-IEC-2382-37-170206} or non-match \cite{ISO-IEC-2382-37-170206} (depending on the ERC error type) for each pair.
QSs produced by the FIQAA per image are combined for the image pairs (\eg{} by taking the minimum).
A progressively increasing quality threshold is applied to these image pair QSs,
and a FR error measure is calculated for the resulting QS subsets.
In \cite{Grother-SampleQualityMetricERC-PAMI-2007}, it is suggested that the FNMR (False Non-Match Rate) \cite{ISO-IEC-2382-37-170206} error measure should be used as the primary performance indicator.
If desired, the FR threshold can then be derived for a fixed FMR (False Match Rate) \cite{ISO-IEC-2382-37-170206} on the unfiltered image pairs - or vice versa if the FMR was plotted as the error measure.
The error is typically plotted on the vertical axis.
The rejected fraction, plotted on the horizontal axis,
denotes the relative amount of images ($0$ to $100\%$) rejected based on the QS.
Plotting this fraction instead of the increasing QS threshold normalizes the axis independently of the given FIQAA.
This also means that QSs do not have to be constrained to a certain range, only their order is important.

Note that ERCs should usually represent the rejection of samples/images, not individual comparisons,
so that all comparisons with quality below the currently considered quality threshold have to be discarded simultaneously.
This means that the horizontal axis actually denotes the maximum of the fraction of images rejected via the quality threshold,
not the precise fraction of rejected images.
This in turn means that ERC plots should prefer stepwise interpolation by continuing the error value from the last real ERC data point at which a batch of comparisons was rejected.
Linear interpolation, as used by some works, can be misleading for rejection fraction ranges with low quality granularity,
which may occur for realistic evaluation configurations.

\markAuthor{Olsen \etal{}} \cite{Olsen-FingerImageQuality-IETBiometrics-2016} further proposed to compute the scalar Area-Under-Curve (AUC) for some rejection fraction range of an ERC:
\begin{equation*}\label{eq:erc-auc}
\int_{a}^{b} ERC - \text{area under theoretical best}
\end{equation*}
More concretely, \cite{Olsen-FingerImageQuality-IETBiometrics-2016} proposed to compute
the AUC for the full $[0,100\%]$ range,
and a partial AUC (pAUC) to focus only on the $[0,20\%]$ range.
The ``area under theoretical best'' term refers to the (unrealistic) best case where the error value decrease equals the rejected fraction percentage.
Also note that the ``area under theoretical best'' is a constant value for a specific AUC range,
so subtracting it from the FIQAAs' ERC curve areas will not alter their relative performance within that specific AUC range.
Consequently, the subtraction can be omitted for AUC computations when only the per-AUC-range FIQAA ranking is analyzed
(which is the case for \autoref{sec:evaluation-concrete}).

A more realistic approximation of an optimal FIQAA may be achieved by means of an oracle,
the concept of which was described by \markAuthor{Phillips \etal{}} \cite{Phillips-FQA-ExistenceOfFaceQuality-BTAS-2013}.
Since more recent FIQA literature did not continue to explore this,
future work could do so in an attempt to improve ERC evaluations.
Conversely,
the error at 0\% rejection can be considered as the practical worst-case,
because the average of many/infinite ERC curves for random QSs will approximately result in no error change for FNMR or FMR,
and no real FIQAA should be worse than random QS assignment.

The FIQAA literature listed in this survey did not always provide ERC or AUC evaluation results.
For example, some works evaluated the FIQAA in terms of quality label prediction performance,
and did not evaluate the FIQAA in terms of FR performance improvements.
Even if all of the literature had utilized a common evaluation result format,
\eg{} ERC plots with the same error measures,
there would still be differences in the used FR systems and datasets.
This issue makes a precise performance comparison based solely on reported results impossible.
Refer to \autoref{sec:challenges} for further discussions regarding this and other issues.

The ongoing NIST Face Recognition Vendor Test (FRVT)
for face image quality assessment \cite{Grother-FQA-4thDraftOngoingFRVT-2021}
evaluates FIQAAs combined with a number of FR algorithms and dataset types,
showing results \ia{} in the form of ERC plots.
Some noteworthy modifications to the usual ERC methodology were applied according to the current draft report \cite{Grother-FQA-4thDraftOngoingFRVT-2021}:
To compute the FNMR at some rejection fraction,
the evaluation divided by the count of comparisons at that rejection fraction (\ie{} comparisons not removed by the quality threshold),
instead of dividing by the total comparison count constant independently of the rejection fraction.
QSs were perturbed with random uniformly distributed noise as a result tie breaker,
and a logarithmic rejection axis was plotted to emphasize the results for smaller rejection fractions.
The report furthermore introduced the ``Incorrect Sample Rejection Rate'' (ISRR) and ``Incorrect Sample Acceptance Rate'' (ISAR),
which are defined to incorporate both FR comparisons and QS rejections.
A future goal of the project is to investigate (non-linear) calibration methods to map QSs of different FIQAAs to a common [0, 100] range with approximately equalized distribution.

\subsection{Selective Evaluation}
\label{sec:evaluation-concrete}

We conducted a FNMR ERC evaluation with 14 FIQA approaches, including both recent methods and general IQAAs,
and at least one method for each data aspect category (described in \autoref{sec:aspect-data}),
except for human quality ground truth training:
\begin{itemize}
\item Hand-crafted (\attrDhc{}):
\begin{itemize}
\item Pose symmetry, Light symmetry, Blur, Sharpness, Exposure, GCF (Global Contrast Factor):
As described by \markAuthor{Wasnik \etal{}} \cite{Wasnik-FQA-SmartphoneISO297945-IWBF-2017} and ISO/IEC TR 29794-5:2010 \cite{ISO-IEC-29794-5-TR-FaceQuality-100312}.
\item PIQE \cite{venkatanathnBlindImageQuality2015}: Publicly available Python implementation.
\end{itemize}
\item Utility-agnostic training (\attrDuat{}):
\begin{itemize}
\item BRISQUE \cite{mittalNoReferenceImageQuality2012}: Publicly available model (pybrisque implementation).
\item NIQE \cite{mittalMakingCompletelyBlind2013}: Publicly available model (scikit-video implementation).
\end{itemize}
\item FR-based ground truth training (\attrDfrt{}):
\begin{itemize}
\item FaceQnet v0 \cite{Hernandezortega-FQA-FaceQnetV0-ICB-2019} \& v1 \cite{Hernandezortega-FQA-FaceQnetV1-2020}: Publicly available models.
\item PCNet \cite{Xie-FQA-PredictiveUncertaintyEstimation-BMVC-2020}: Model provided by the authors.
\end{itemize}
\item FR-based inference (\attrDfri{}):
\begin{itemize}
\item SER-FIQ \cite{Terhorst-FQA-SERFIQ-CVPR-2020}: Publicly available model (``same model'' variant on ArcFace, which is also used for FR).
\end{itemize}
\item FR-integration (\attrDint{}):
\begin{itemize}
\item MagFace \cite{Meng-FRwithFQA-MagFace-arXiv-2021}: Publicly available model (``iResNet100'' backbone, trained on MS1MV2 \cite{Deng-ArcFace-IEEE-CVPR-2019}).
\end{itemize}
\end{itemize}

\begin{figure*}
    \centering
    \includegraphics[width=0.8\textwidth]{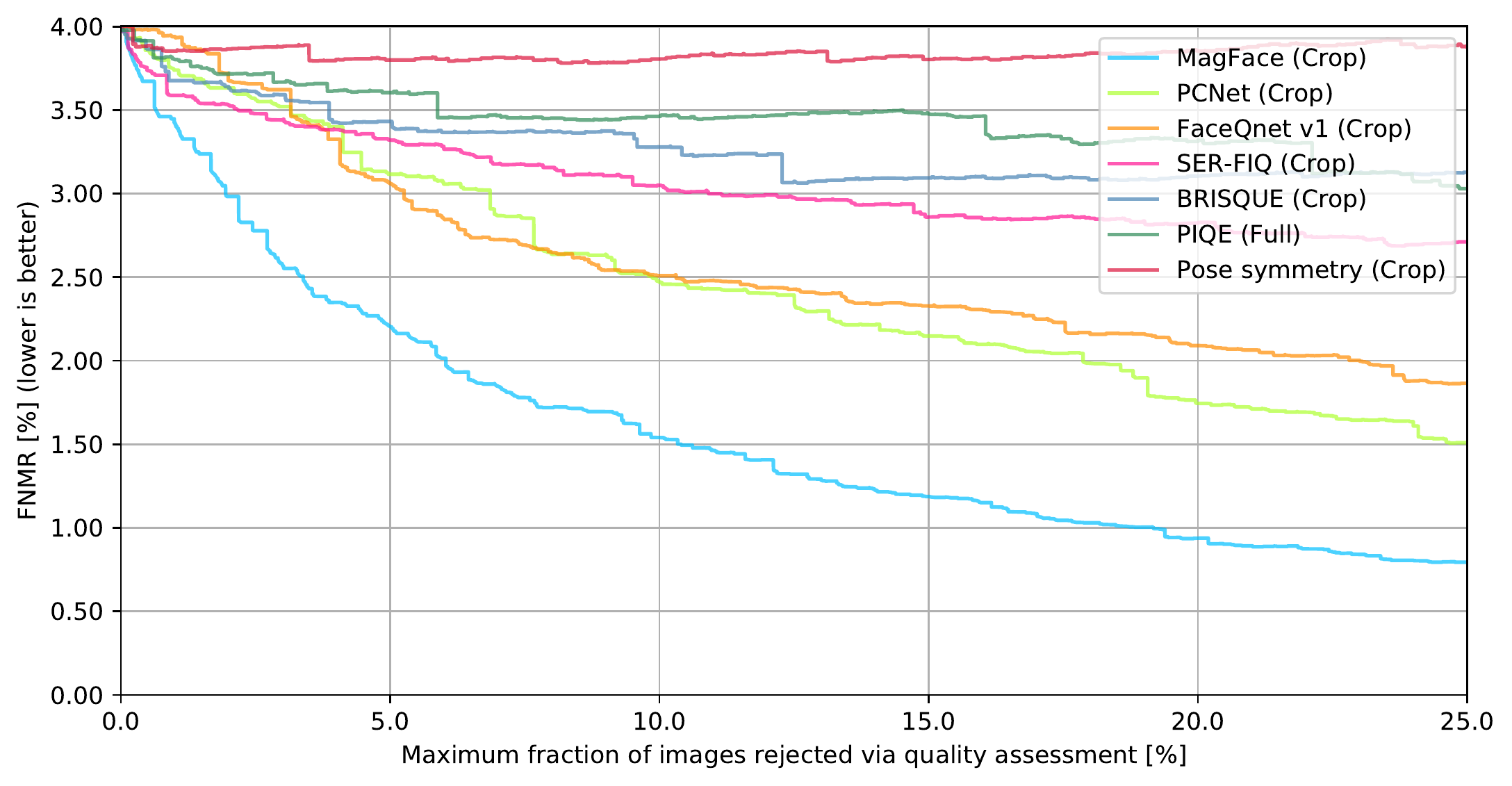}
    \caption{\label{fig:erc-plot} ERC plot for a subset of the evaluated FIQAAs.
    EDC pAUC results for all evaluated FIQAAs are provided in \autoref{tab:erc-auc}.
    }
\end{figure*}

\begin{table*}
	\caption{\label{tab:erc-auc} ERC evaluation results in terms of the partial Area-Under-Curve (pAUC) values for reject fraction ranges from 0\% to 1\%/5\%/20\%.
	For each entry, \eg{} ``1: 0.00\% (0.037\%)'',
	the first number denotes the ranking of the FIQAA (``1:'' being the best, ``14:'' the worst),
	the second number shows the relative performance (``0.00\%'' being the best, ``100.00\%'' the worst),
	and the bracketed third number is the actual absolute pAUC value (higher being worse).
	ERC results are also plotted for a subset of the FIQAAs in \autoref{fig:erc-plot}.}
	\centering
	\tableFontSize
	\begin{tabular}{rrrrrrr}
\hline
\textbf{{FIQAA}} &  & \textbf{1\% pAUC} &  & \textbf{5\% pAUC} &  & \textbf{20\% pAUC} \\
\hline
MagFace (Crop) & \textit{1:} & 0.00\% (0.037\%) & \textit{1:} & 0.00\% (0.144\%) & \textit{1:} & 0.00\% (0.356\%) \\
SER-FIQ (Crop) & \textit{2:} & 30.69\% (0.038\%) & \textit{3:} & 56.54\% (0.176\%) & \textit{5:} & 61.00\% (0.625\%) \\
FaceQnet v0 (Crop) & \textit{3:} & 51.52\% (0.038\%) & \textit{2:} & 54.85\% (0.175\%) & \textit{4:} & 57.92\% (0.612\%) \\
BRISQUE (Crop) & \textit{4:} & 59.35\% (0.039\%) & \textit{6:} & 66.27\% (0.181\%) & \textit{6:} & 69.18\% (0.662\%) \\
PCNet (Crop) & \textit{5:} & 61.10\% (0.039\%) & \textit{4:} & 60.35\% (0.178\%) & \textit{2:} & 40.46\% (0.535\%) \\
PIQE (Full) & \textit{6:} & 69.29\% (0.039\%) & \textit{8:} & 75.74\% (0.186\%) & \textit{8:} & 78.23\% (0.702\%) \\
Pose symmetry (Crop) & \textit{7:} & 71.78\% (0.039\%) & \textit{10:} & 87.60\% (0.193\%) & \textit{11:} & 92.63\% (0.765\%) \\
Blur (Crop) & \textit{8:} & 74.32\% (0.039\%) & \textit{7:} & 70.36\% (0.183\%) & \textit{7:} & 75.66\% (0.690\%) \\
Sharpness (Crop) & \textit{9:} & 82.52\% (0.039\%) & \textit{9:} & 78.77\% (0.188\%) & \textit{9:} & 86.17\% (0.737\%) \\
FaceQnet v1 (Crop) & \textit{10:} & 94.37\% (0.040\%) & \textit{5:} & 65.15\% (0.180\%) & \textit{3:} & 43.40\% (0.548\%) \\
Exposure (Crop) & \textit{11:} & 95.55\% (0.040\%) & \textit{13:} & 99.81\% (0.200\%) & \textit{14:} & 100.00\% (0.798\%) \\
NIQE (Full) & \textit{12:} & 95.74\% (0.040\%) & \textit{11:} & 96.42\% (0.198\%) & \textit{12:} & 96.40\% (0.782\%) \\
Light symmetry (Crop) & \textit{13:} & 97.94\% (0.040\%) & \textit{12:} & 97.78\% (0.198\%) & \textit{13:} & 97.94\% (0.789\%) \\
GCF (Full) & \textit{14:} & 100.00\% (0.040\%) & \textit{14:} & 100.00\% (0.200\%) & \textit{10:} & 91.82\% (0.762\%) \\
\hline
\end{tabular}

\end{table*}

The error at 0\% rejection is set to 4\% FNMR.
ArcFace \cite{Deng-ArcFace-IEEE-CVPR-2019} was used for FR (cosine similarity),
and RetinaFace \cite{Deng-FaceDetection-RetinaFace-CVPR-2020} was used for face/facial landmark detection,
employing the publicly available models for both (InsightFace's ``LResNet100E-IR,ArcFace@ms1m-refine-v2'' \& ``RetinaFace-R50'').

We used LFW (Labeled Faces in the Wild) \cite{LFWTech} as the evaluation dataset
and consider all possible mated pairs therein.
As shown by \autoref{tab:datasets},
LFW \cite{LFWTech} is the dataset that has been employed by the greatest number of FIQA works,
including recent ones.
The FR performance on LFW \cite{LFWTech} appears to already be almost saturated by the state-of-the-art systems,
and the quality distribution correspondingly seems to be more narrow than in \eg{} IJB-C \cite{Maze-Face-IARPAJanusBenchmarkC-ICB-2018},
as demonstrated most recently for MagFace \cite{Meng-FRwithFQA-MagFace-arXiv-2021}.
This conversely means that LFW \cite{LFWTech} is more challenging for FIQA ERC evaluations,
since FIQAAs have to more effectively rank images in terms of biometric utility to decrease the error rate,
especially for lower rejection fractions.

\autoref{fig:erc-plot} shows the ERC plot,
but only with a subset of the FIQAAs for the sake of legibility,
since multiple curves for the less well performing methods would approximately overlap graphically.
\autoref{tab:erc-auc} however lists ranked ERC pAUC results for all 14 FIQAAs,
which is a more useful representation for the analysis of many methods.

LFW \cite{LFWTech} images depict a substantial amount of background information besides the actual face,
so the type of preprocessing is relevant.
For this evaluation,
the ERC was computed for all FIQAA methods using both the full images (marked as ``Full'')
and preprocessed variants (marked as ``Crop'').
The preprocessing variant used RetinaFace \cite{Deng-FaceDetection-RetinaFace-CVPR-2020}
to crop the images to the face,
and to subsequently align the images to the detected facial landmarks (there are no edge cases without a detected face or landmarks).
Only the best performing variants per FIQAA at 1\% pAUC are shown,
to avoid cluttered results.
All \attrDfrt{}/\attrDfri{}/\attrDint{} approaches performed better with the preprocessed images,
as to be expected due to their incorporation of FR during training and/or inference.
A few of the \attrDhc{}/\attrDuat{} approaches did however yield better results using the full images even for the considered 1\% pAUC.
This applies to more \attrDhc{}/\attrDuat{} approaches for higher pAUC maxima (\eg{} for ``Light symmetry'' at 20\% pAUC),
but the difference is never substantial enough to compete with the top ranking FIQAAs regardless,
so we do not include more detailed results regarding this.

As apparent in \autoref{tab:erc-auc} and \autoref{fig:erc-plot},
MagFace \cite{Meng-FRwithFQA-MagFace-arXiv-2021} has distinctly achieved the best results throughout this particular evaluation.
The ranking of the other FIQAAs depends on the considered pAUC range:
For 5\% and 20\% pAUC, the five \attrDfrt{}/\attrDfri{}/\attrDint{} approaches outperformed all \attrDhc{}/\attrDuat{} methods,
but for 1\% pAUC only three \attrDfrt{}/\attrDfri{}/\attrDint{} held the top rankings, BRISQUE \cite{mittalNoReferenceImageQuality2012} being able to compete most closely with them.
Note that we do not include results for higher ERC rejection fractions,
since these are less interesting from an operational perspective.
In practice one would not want to reject \eg{} every second image (50\% rejection fraction),
if the images mostly have high FR utility - as is the case for LFW \cite{LFWTech},
according to the aforementioned high state-of-the-art FR performance.

While issues and challenges in general are discussed in the subsequent section,
it is also important to highlight limitations of this particular evaluation,
which can be used to show what future work may want to consider:
\begin{itemize}
\item Only a low amount of
FIQA/\allowbreak{}%
FR/\allowbreak{}%
dataset/\allowbreak{}%
preprocessing/\allowbreak{}%
hyperparameter configurations was tested in contrast to the available options.
A more comprehensive literature evaluation will require re-implementation efforts for the many listed works that did not provide open reference implementations,
and automated configuration exploration may have to be employed to overcome a combinatorial explosion.
Static ERC plots can quickly become too cluttered,
as was already the case here with only 14 FIQAAs,
but reduced or interactive plots can still useful,
and derived metrics such as pAUC can be used to analyze arbitrary configuration counts.
\item No non-mated pairs were considered, since only the FNMR was used as ERC error.
It could be interesting to test FMR, or possibly other metrics, as the error - 
especially because recent FIQA approaches have started to incorporate both mated and non-mated pairs during training.
\item While the use of publicly available models is beneficial in terms of reproducibility,
the comparisons are not as fair as they could be due to differing training data.
For \attrDfrt{}/\attrDfri{}/\attrDint{}, the different training data does imply that the results may not fairly reflect the potential of the underlying FIQA concepts or network architectures.
Different preprocessing during training or different training time could likewise affect the performance.
Note that this means that black-box FIQAAs (\eg{} COTS systems) cannot be fairly compared by definition.
Comparisons between \attrDfrt{}/\attrDfri{}/\attrDint{} and \attrDhc{}/\attrDuat{} approaches in this evaluation are however rather unproblematic,
since \attrDuat{} approaches typically require different training data by design (\eg{} general IQA training data instead of face images),
and \attrDhc{} requires no training.
\item The computational performance of FIQAAs could be relevant in practice too, thus evaluations could be helpful.
For anything except \attrDfri{}/\attrDint{} approaches, the computational performance should by definition be independent of the FR system choice,
and dataset/preprocessing configuration may also be unimportant in this context.
However, hardware configurations (\ia{} CPU versus GPU) can matter instead,
and implementations details have to be considered as well (\ie{} concrete implementations may not utilize the given hardware as effectively as the underlying concepts would allow).
\end{itemize}

\section{Open Issues and Challenges}
\label{sec:challenges}

An obvious challenge consists of the further improvement of FIQA methods in terms of predictive and computational performance.
For deep learning FIQA approaches, finding better network architectures and training methods is interwoven with general deep learning research progress, for example in the field of automated machine learning \cite{automl_book}.
Naturally, FIQA with the goal of generating quality scores that predict FR utility \cite{ISO-IEC-29794-1-QualityFramework-160915} also depends on FR research.

The following subsections describe further issues and challenges, as well as potential avenues for future work,
and the summary \autoref{sec:summary} highlights the identified key challenges.

\subsection{Comparability and Reproducibility}
\label{sec:comparability}

As previously noted in \autoref{sec:evaluation},
it would be challenging to comprehensively compare the performance of the surveyed FIQA approaches,
since the evaluations presented in the literature differ in multiple aspects that would need to be aligned to facilitate fair direct comparisons:
\begin{itemize}
\item \textbf{Datasets}:
As shown in \autoref{tab:datasets},
a variety of datasets were used for the evaluations among the literature.
Besides these named datasets,
some of the literature only utilized private or unspecified data for evaluation.
In addition,
some literature used only a subset of a dataset (see \eg{} \cite{Hernandezortega-FQA-FaceQnetV1-2020} or \cite{Yang-FQA-DFQA-ICIG-2019} regarding the VGGFace2 dataset \cite{Cao-VGGFace2Dataset-FGR-2018}),
or modified the data \eg{} by synthetically degrading images via increased blur or contrast (see \eg{} \cite{Abaza-FQA-PhotometricIQA-IET-2014}).
Where training data is required for the FIQAA,
the chosen subdivision of the datasets into training and test data also influences the evaluation results.
Furthermore, various works assigned ground truth quality scores or labels to the dataset for FIQAA training and\slash or for the evaluation.
When FIQA is evaluated in terms of FR performance improvements,
the selection of image pairs that are considered initially for FR comparisons \cite{ISO-IEC-2382-37-170206} (\ie{} before filtering them via FIQA decisions) may alter the results as well.
Another potentially interesting question is the degree of existing overlap between datasets regarding FR,
which could be studied both in a general FR context and in the context of FIQA.

\item \textbf{Evaluation methods}:
Different evaluation methods and ways to report results are used among the literature.
Some FIQA approaches are only tested by comparing predicted quality scores or labels against a given ground truth (\eg{} assigned by humans),
\ie{} not all of the literature evaluates FIQA in terms of FR utility \cite{ISO-IEC-29794-1-QualityFramework-160915}\cite{Alonsofernandez-QualityMeasures-SecPri-2012} in the first place.
Instead of evaluating the FIQA on its own,
some literature that included image enhancement steps in the evaluation.
For FR performance improvement evaluations via an ERC as described in \autoref{sec:evaluation-erc},
the FR comparison score threshold \cite{ISO-IEC-2382-37-170206} and the error type configuration can differ between evaluations,
which also applies to ERC-derived AUC results.
Some of the works evaluated FIQA performance exclusively by means other than the ERC
- for example, FR performance was evaluated for 4 FIQA-derived quality bins in \cite{Bharadwaj-FQA-HolisticRepresentations-ICIP-2013}.

\item \textbf{FR algorithms}:
Evaluating FIQA in terms of FR performance improvement is desirable
to examine how well quality scores of a FIQAA reflect FR utility \cite{ISO-IEC-29794-1-QualityFramework-160915},
but this also introduces the FR algorithm choice for feature extraction \cite{ISO-IEC-2382-37-170206} and comparison \cite{ISO-IEC-2382-37-170206} as another evaluation factor.
Furthermore,
there are FIQA approaches among the literature which are conceptually based on FR models to begin with (see \eg{} \cite{Terhorst-FQA-SERFIQ-CVPR-2020}),
and FR algorithms are used by various works to establish ground truth quality scores\slash labels (see \eg{} \cite{Hernandezortega-FQA-FaceQnetV1-2020} for scores, or \cite{Bharadwaj-FQA-HolisticRepresentations-ICIP-2013} for labels in the form of 4 quality bins).
Lastly, some literature exclusively used anonymous and/or closed-source FR systems, which can limit reproducibility and expandability (see \eg{} \cite{Bharadwaj-FQA-HolisticRepresentations-ICIP-2013}).
\end{itemize}

Due to the amount of existing and possible FIQA evaluation configurations,
the comparison of FIQAAs can be considered as a key challenge.
This open issue could be limited in scope \eg{} by only considering FIQA approaches that can conceptually adapt to deep learning FR systems (instead of relying on hand-crafted algorithms, settings, or ground truth quality scores).
One solution for future work is to submit the presented FIQAAs to an evaluation campaign where all algorithms are assessed under the same benchmark,
such as the previously mentioned
NIST FRVT Quality Assessment evaluation \cite{Grother-FQA-4thDraftOngoingFRVT-2021}.
Open evaluation protocols could be established as well.

Another solution is to publicly provide the FIQAA implementations,
allowing other researchers to integrate them in different evaluation environments
without re-implementation.
Besides being redundant effort,
a re-implementation can diverge from the original algorithm to some degree
even without introducing errors,
since \eg{} deep learning model weight initialization can be random
(which however might only be a minor issue).
Since evaluations of machine learning FIQA in particular depend on the used training data,
publishing source code is preferable to pure black box releases.
So for the sake of both comparability and reproducibility,
future work should provide source code and trained models where applicable.
This may also serve as a basis for new FIQA approaches in later work by other researchers.
Effective reuse of prior work implementations can \ia{} be observed in the surveyed literature by the utilization of pretrained FR models.
Providing source code is not necessarily important for approaches that can easily be described in complete detail within a paper,
e.g. simpler hand-crafted methods without any machine learning and few parameters,
but approaches in the recent literature tend to be more complex.
While most of the older surveyed literature did not appear to publish accompanying source code (irrespective of the implementation complexity),
more recent deep learning FIQA works tend to do so, with code being publicly available for \eg{}
FaceQnet \cite{Hernandezortega-FQA-FaceQnetV0-ICB-2019}\cite{Hernandezortega-FQA-FaceQnetV1-2020},
PFE (Probabilistic Face Embeddings) \cite{Shi-FRwithFQA-ProbabilisticFaceEmbeddings-ICCV-2019},
SER-FIQ \cite{Terhorst-FQA-SERFIQ-CVPR-2020},
and MagFace \cite{Meng-FRwithFQA-MagFace-arXiv-2021}.

Likewise, public datasets should preferably be used,
and precise evaluation configurations could be published alongside the implementation.
It may also be helpful to publish the raw evaluation result as supplementary data,
\eg{} the computed comparison scores and quality scores,
although this may be unnecessary if the results are reproducible already.
This result data could \eg{} be used to directly create new visualizations that combine results from multiple works.

Outside of evaluating the predictive performance of FIQAAs,
evaluating the computational performance may be of relevance as well.
This is rarely considered in the surveyed FIQA literature.
Computational performance tests usually focus on measuring the duration required to process input images with a certain format (\eg{} grayscale) and resolution,
since they are typically not influenced by other factors that are unavoidable in utility prediction performance evaluations.
Other factors do however become relevant,
namely the computational optimization of the FIQAA,
as well as the used hardware and the robustness of the time measurements.
Besides measuring inference time,
a different kind of computational performance tests could assess the efficiency of FIQA model training as well,
although this is less relevant in an operational context as long as no frequent (re-)training is required.

\subsection{Explainability and Interpretability}
While the more recent monolithic deep learning FIQA approaches are trained specifically to output quality scores in terms of FR utility \cite{ISO-IEC-2382-37-170206}\cite{Alonsofernandez-QualityMeasures-SecPri-2012},
they are not as interpretable\slash explainable as \eg{} hand-crafted approaches that estimate specific human-understandable factors such as blur.
This can be considered as another key challenge.
Optimally, FIQA models should be able to predict FR utility \cite{ISO-IEC-2382-37-170206} while also providing useful feedback regarding quality-degrading causes.
Future work could thus attempt to improve upon this area,
perhaps by adding visualizations based on a disentangled latent space that corresponds to different kinds of quality degradations.
In this line of explainable Artificial Intelligence (AI) and, in particular, in fairness and bias control in AI systems \cite{Serna-Face-AlgorithmicDiscriminationDeepLearning-SafeAI-2020}\cite{Drozdowski-BiasSurvey-TTS-2020}, we expect growing interest in analyzing the behavior of FIQA methods for different population groups and the development of FIQA methods more transparent \cite{Barredoarrieta-ExplainableArtificialIntelligenceXAI-INFFUS-2020} and agnostic to selected covariates \cite{Morales-Face-SensitiveNetsLearningAgnostic-PAMI-2021}.

\subsection{Use of Synthetic Data}
For FIQA in general,
preferably large amounts of realistic data including different quality levels with different quality-degrading causes should be used for evaluation (and training where applicable),
such that the robustness can be verified for various cases with a high certainty.
Existing images can also be degraded synthetically
- this was done in a few works (\eg{} \cite{Abaza-FQA-PhotometricIQA-IET-2014}).
That is, both known techniques from prior work, such as Gaussian blurring,
and more sophisticated techniques, such as deep learning style transfer,
could be employed in the future.
It is also possible to generate fully synthetic face images (see \eg{} StyleALAE \cite{pidhorskyiAdversarialLatentAutoencoders2020}),
which is a strategy that has not been used in the surveyed FIQA literature.
While fully synthetic data might be less realistic,
it could allow for larger datasets with better control
(in terms of training\slash evaluation sample bias)
than what \eg{} filtering a real dataset might provide.
As a side effect, using fully synthetic data may potentially also alleviate licensing or privacy concerns (see \eg{} the controversy surrounding MS-Celeb-1M \cite{Guo-Face-MSCeleb1M-ECCV-2016}, which has been used in some of the FIQA literature as well).
This latter point is however not entirely clear, since deep learning face synthesis itself is typically trained on real face images.

\subsection{Interoperability}
Examining and improving interoperability in terms of FIQA FR utility prediction generality could be another goal for future work.
While this may partially stand in conflict with the goal of maximizing FR-system-specific utility prediction performance, interoperability can be relevant to avoid vendor lock-in
and may coincide with increased robustness.
An example in the literature is the FaceQnet approach,
which went from using only one FR system as part of the training process in v0 \cite{Hernandezortega-FQA-FaceQnetV0-ICB-2019} to using three in v1 \cite{Hernandezortega-FQA-FaceQnetV1-2020}.

\subsection{Vulnerabilities}

Specific attacks on FIQA may be investigated in future works. For instance, the surveyed machine learning FIQA literature did not study adversarial attacks,
\ie{} attacks that specifically modify
the input (physical \cite{Galbally-Survey-Face-PAD-IEEEAccess-2014} or digital after being captured and processed \cite{Galbally-Face-VerificationHillclimbingAttacks-PR-2010})
to confuse the FIQA model.

\subsection{Standardization}

ISO/IEC 29794-1:2016 \cite{ISO-IEC-29794-1-QualityFramework-160915} defines the notion of biometric sample quality,
and a new edition is currently under development \cite{ISO-IEC-29794-1-WD}.
At the time of this writing, this new edition will \ia{} standardize ERC (\sectionref{sec:evaluation-erc}) for FIQAA evaluation.
ISO/IEC TR 29794-5:2010 \cite{ISO-IEC-29794-5-TR-FaceQuality-100312} describes various actionable FIQA measures,
and the next edition is under development as an International Standard \cite{ISO-IEC-29794-5-WD}.
Current portrait quality specifications are established in
ISO/IEC 39794-5:2019 \cite{ISO-IEC-39794-5-FaceInterchangeFormats-191220},
which contains content from the ICAO Portrait Quality TR \cite{ICAO-PortraitQuality-TR-2018},
which in turn was based on parts of
ISO/IEC 19794-5:2011 \cite{ISO-IEC-19794-5-G2-FaceImage-110304},
ISO/IEC 19794-5:2005 \cite{ISO-IEC-19794-5-2005}
and ICAO Doc 9303 \cite{ICAO-9303-2015}.
ISO/IEC 24358 (``Face-aware capture subsystem specifications'') \cite{ISO-IEC-24358-WD-TS}
is another relevant standard that is under development at the moment.
An important future goal for FIQA is the standardization of some particular FIQA algorithm/model,
analogous to the normative standardization of the open source NIST Fingerprint Image Quality (NFIQ) 2 as part of ISO/IEC 29794-4:2017 \cite{ISO-IEC-29794-4-FingerQuality-2017}.

\subsection{Further Applications}
\label{sec:unexplored}
As described in \autoref{sec:application-areas},
there are further application areas that were barely or not at all examined in the surveyed literature.
For example, lossy compression control was not considered at all, although compression artifacts are mentioned as a quality degrading factor by various works.
FIQA for other areas besides face recognition can also be explored further,
including FIQA in the context of gender or other soft biometrics recognition \cite{Gonzalezsosa-Face-SoftRecognitionWild-TIFS-2018}, attention level estimation \cite{Daza-mEBAL-MultimodalDatabaseEyeBlinkAttentionLevel-arXiv-2020}, emotion analysis \cite{Pena-LearningEmotionalBlindedFaceRepresentations-ICPR-2021}, \etc{}

Almost all of the found FIQA literature focused the visible spectrum.
The exception is the work by \markAuthor{Long \etal{}} \cite{Long-FQA-VideoFrameNIR-ICIG-2011},
which studied quality assessment for near-infrared face video sequences.
They combined measures for sharpness, brightness, resolution, landmark-based head pose, and expression in terms of eyes/mouth being open/closed,
but the evaluation was limited to comparisons against human rankings.
Future work could thus quickly expand on FIQA for near-infrared images, or for other spectra \cite{Moreno-BeyondTheVisibleSpectrum-BioID-2009}.

Furthermore, FIQA may also be relevant for face ``depth'' or ``range'' images, \ie{} 2D images depicting 3D positions in terms of depth.
But, similar to non-visible-spectrum FIQA, few works appear to exist that consider depth FIQA in a biometric context.
The work by \markAuthor{Lin and Chen} \cite{Lin-3dFaceRecognitionQualityAssessment-MTA-2014} is one instance that included depth FIQA
using a deformable shape model to identify excessive expression variations,
and the FIQA part was used to improve 3D FR performance via sample rejection with a fixed quality threshold.
Future work could further explore depth FIQA, or 3D face quality assessment for other 3D representations.
Combinations of \eg{} visible spectrum and depth images for FIQA,
as well as FIQA for other application areas such as biometric depth image enhancement \cite{Schlett-FaceDepthEnhancement-CVIU-2021},
could be investigated as well.

Another related field that future FIQA research may want to consider is face sketch recognition/synthesis,
where literature published so far appears to be focused on perceptual measures
instead of biometric utility prediction \cite{Bi-FaceSketch-SynthesisSurvey-MTA-2021},
a concrete recent example being the work by \markAuthor{Fan \etal{}} \cite{Fan-FaceSketch-ScootPerceptualQuality-ICCV-2019}.

\section{Summary}
\label{sec:summary}

Face image quality assessment is an active research area,
and can be used for a variety of application scenarios
such as filtering and feedback during the acquisition process,
or for database maintenance and monitoring.
The literature surveyed in this work predominantly focused on evaluating the proposed FIQA approaches either in terms of predictive performance with respect to given ground truth quality score labels,
or in terms of utility \cite{ISO-IEC-2382-37-170206}\cite{Alonsofernandez-QualityMeasures-SecPri-2012}
for the purpose of aiding face recognition by discarding images based on the assessed quality
or some kind of quality-based processing or fusion \cite{Fierrez-Fusion-MultipleClassifiersQualityBased-INFFUS-2018}.
Automatic face quality assessment is especially relevant for FR as part of large-scale systems, \eg{} the
European Schengen Information System (SIS), the VISA Information System (VIS), the Entry Exit System (EES), or the US ESTA (Electronic System for Travel Authorization),
due to the amount of data and the multitude of different acquisition locations\slash devices.

A progression over time towards monolithic deep learning approaches was observed in the FIQA literature.
Older methods were predominantly factor-specific and independent of concrete FR systems,
while more recent methods tended to train on ground truth quality scores derived from FR comparisons.
Some of the most recently emerging monolithic methods expanded on the FR focus,
either by relying on FR systems during inference,
or by integrating FIQA into FR models.

One key challenge is to facilitate comparability of the FIQA evaluations,
since many differing evaluation configurations were employed in the literature.
Thus, future work should preferably provide the implementations of the proposed FIQAAs publicly, especially in the form of source code,
enabling evaluations in later works to more easily include these FIQA approaches.
The more recent works have begun to do so, but re-implementation efforts will be required if many of the older approaches are to be evaluated comprehensively.
There also is the ongoing
NIST FRVT Quality Assessment
evaluation \cite{Grother-FQA-4thDraftOngoingFRVT-2021},
to which FIQAAs can be submitted.
Besides evaluating the predictive capabilities of FIQAAs,
more attention could be paid to computational performance evaluations in the future.

Another key challenge is to improve the interpretability of deep learning based FIQA,
which so far mostly fell into the monolithic category of this survey,
meaning that these modern approaches did not focus on providing extensive feedback for human operators to adjust acquisition conditions for increased biometric utility.

Of course there also is the key challenge of further improving performance in terms of both utility
and computational workload
(\eg{} with new deep learning network architectures),
as well as improving robustness\slash decreasing bias
\cite{Serna-Face-AlgorithmicDiscriminationDeepLearning-SafeAI-2020}\cite{Drozdowski-BiasSurvey-TTS-2020}
(\eg{} via the selection or synthetic extension of datasets for different quality degradation cases),
which naturally is dependent on suitable evaluation methodologies.

In the long term,
an important objective is the standardization of a specific FIQA approach,
analogous to the normative standardization of the open source NIST Fingerprint Image Quality (NFIQ) 2 as part of ISO/IEC 29794-4:2017 \cite{ISO-IEC-29794-4-FingerQuality-2017},
and advances regarding the aforementioned challenges can help to achieve this.
Various other application scenarios can be explored further as well,
\eg{} FIQA-guided image enhancement or compression.

\section*{Acknowledgment}

This research work has been funded by the German Federal Ministry of Education and Research and the Hessen State Ministry for Higher Education, Research and the Arts within their joint support of the National Research Center for Applied Cybersecurity ATHENE,
project BIBECA (RTI2018-101248-B-I00 MINECO/FEDER), and project TRESPASS-ETN (H2020-MSCA-ITN-2019-860813).
This project has received funding from the European Union’s Horizon 2020 research and innovation programme under grant agreement No 883356.
This text reflects only the author’s views and the Commission is not liable for any use that may be made of the information contained therein.

\section*{References}
\AtNextBibliography{\small}
\printbibliography[heading=none]

\end{document}